\documentclass[preprint,10pt]{elsarticle}

\usepackage{booktabs}

\usepackage{comment}

\usepackage{amsthm,amsmath,amsfonts,amsfonts}

\usepackage{graphicx}
\usepackage[section]{placeins}
\usepackage{epstopdf}
\usepackage[tight,normalsize,sf,SF]{subfigure}
\usepackage{color}
\usepackage{footnote}
\usepackage{caption}
\usepackage{graphicx}
\usepackage{algorithmic}
\usepackage{algorithm}
\usepackage{wrapfig}
\usepackage{lipsum}

\newcommand{\MYfooter}{\smash{
\hfil\parbox[t][\height][t]{\textwidth}{
}\hfil\hbox{}}}
\makeatletter
\def\ps@IEEEtitlepagestyle{%
\def\@oddfoot{\MYfooter}%
\def\@evenfoot{\MYfooter}}
\makeatother
\newcounter{letter}

\newcommand {\myvec}[1] {{\mbox{\boldmath $#1$}}}
\newcommand {\mymat}[1]  {{\mbox{\boldmath $#1$}}}

\newcommand {\defeq}{\stackrel{\triangle}{=}}

\newcommand {\mX}   {\mymat{X}}
\newcommand {\mPsi} {\mymat{\Psi}}
\newcommand {\mPhi} {\mymat{\Phi}}

\newcommand {\mI}   {\mymat{I}}

\newcommand {\mD}   {\mymat{D}}
\newcommand {\mK}   {\mymat{K}}
\newcommand {\mZ}   {\mymat{Z}}

\newcommand {\mPi}   {\mymat{\Pi}}

\newcommand {\hmPsi} {\widehat{\mPsi}}
\newcommand {\hmD}   {\widehat{\mD}}
\newcommand {\hmK}   {\widehat{\mK}}

\newcommand {\hmP}   {\widehat{\mP}}

\newcommand {\ue} {\myvec{e}}
\newcommand {\ux} {\myvec{x}}
\newcommand {\uy} {\myvec{y}}
\newcommand {\uz} {\myvec{z}}

\newcommand {\Rset} {\mathbb{R}}

\newtheorem{D1}{Definition}
\newtheorem{L1}{Lemma}

\def\Diag{\mathop{\rm Diag}\nolimits}

\def\var{\mathop{\rm var}\nolimits}

\DeclareMathAlphabet\mathbfcal{OMS}{cmsy}{b}{n}

\newcommand {\noten}[1]  {\mathcal{#1}}

\newcommand {\vpsi} {\myvec{\psi}}
\newcommand {\vv} {\myvec{v}}
\newcommand {\vx} {\myvec{x}}
\newcommand {\vz} {\myvec{z}}

\newcommand {\ds} {{\rm d}\myvec{s}}
\newcommand {\dt} {{\rm d}\myvec{t}}
\newcommand {\dy} {{\rm d}\myvec{y}}
\newcommand {\dz} {{\rm d}\myvec{z}}

\newcommand {\mP} {\mymat{P}}
\newcommand {\mQ} {\mymat{Q}}
\newcommand {\mW} {\mymat{W}}

\newcommand {\mLam} {\mymat{\Lambda}}

\newcommand {\mo} {\mymat{0}}

\newcommand {\nD} {\noten{D}}

\newcommand {\nK} {\noten{K}}

\newcommand {\Tr} {\text{Trace}}

\newcommand {\tps} {\text{T}}

\newtheorem{T1}{Theorem}

\begin{document}

\begin{frontmatter}

\title{Multi-View Diffusion Maps  }

\author{Ofir~Lindenbaum\fnref{label1}}
\ead{ofirlin@gmail.com}
\author{       Arie~Yeredor\fnref{label1}}
\author{      Moshe~Salhov\fnref{label2}}
\author{      Amir~Averbuch\fnref{label2}}
\address[label1]{School of Electrical Engineering, Tel Aviv University, Israel}
\address[label2]{School of Computer Science, Tel Aviv University, Israel}


\begin{abstract}
In this paper, we address the challenging task of achieving multi-view dimensionality reduction. The goal is to effectively use the availability of multiple views for extracting a coherent low-dimensional representation of the data. The proposed method exploits the intrinsic relation within each view, as well as the mutual relations between views. The multi-view dimensionality reduction is achieved by defining a cross-view model in which an implied random walk process is restrained to hop between objects in the different views. The method is robust to scaling and insensitive to small structural changes in the data. We define new diffusion distances and analyze the spectra of the proposed kernel. We show that the proposed framework is useful for various machine learning applications such as clustering, classification, and manifold learning. Finally, by fusing multi-sensor seismic data we present a method for automatic identification of seismic events.

\end{abstract}


\end{frontmatter}
\section{Introduction}
\label{}

high-dimensional big data are becoming ubiquitous in a growing variety of fields, and pose new challenges in their analysis. Extracted features are useful in analyzing
these datasets. However, some prior knowledge or modeling is required in
order to identify the essential features. Unsupervised 
dimensionality reduction methods, on the other hand, aim to
find low-dimensional representations based on the
{\textit{intrinsic geometry}} of the analyzed data. A ``good'' dimensionality reduction methodology reduces the
complexity of the data, while preserving its coherency, thereby facilitating data analysis tasks (such as clustering, classification, manifold
learning and more) in the reduced space. Many methods such as Principal
Component Analysis (PCA) \cite{PCA}, Multidimensional Scaling (MDS)
\cite{MDS}, Local Linear Embedding \cite{LLE}, Laplacian Eigenmaps (LE)
\cite{GL}, Diffusion Maps (DM) \cite{Lafon}, p-Laplacian Regularization \cite{p-laplacian}, T-distributed Stochastic Neighbor Embedding (t-SNE) \cite{tsne}, Uniform manifold approximation \cite{umap} and more have been
proposed to achieve robust
dimensionality reduction. 
low-dimensional representations have been shown to be useful in various
applications such as face recognition based on LE \cite{he2005face}, Non-linear independent component analysis using
DM \cite{Singer2}, Musical Key extraction employing DM
\cite{lindenbaum2015musical}, and many more. 

Frameworks such as \cite{PCA,MDS,LLE,GL,Lafon,p-laplacian,tsne,umap} do not
consider the possibility to exploit more than one view representing
the same process. Such multiple views can be obtained, for example, when the same underlying process is observed using several different modalities, or measured with different instrumentations. An additional view can provide meaningful insights regarding the dynamical process behind the data, whenever the data is indeed generated or governed by such a latent process. 

This study is devoted to the development of a framework for obtaining a low-dimensional parametrization from multiple measurements organized as ``views''. Our approach essentially relies on quantifying of the speed of a random walk restricted to ``hop'' between views. The analysis of such a random walk provides a natural representation of the data using joint l organization of the multiple measurements. 


The problem of learning from multiple views has been studied in several domains. Some prominent statistical approaches, for addressing this problem are Bilinear Models \cite{Bilinear}, Partial Least Squares \cite{PLS} and Canonical
Correlation Analysis (CCA) \cite{CCA}. These methods are powerful for
learning the relations among the different views, but are restricted to linear transformation for each view. Some methods such as \cite{lai2000kernel,bach2002kernel,Kumar2,Boots,DeSa,sun2016multiview} extend the CCA to nonlinear transformations by introducing a kernel, and demonstrate the advantages of fusing multiple views for clustering and classification. A Hessian multiset canonical correlations is presented in \cite{hCCA}, the authors propose to overcome limitations of the Laplacian by defining local Hessians on the samples to capture structures of multiple views. 
Markov based methods such as \cite{CDDM,Zhou,lederman2014common} use diffusion distances for classification, clustering or retrieval tasks. The
``agreement'' (also called ``consensus'') between different views is used in \cite{moshe} to extract the geometric
information from all views. 
A sparsity-based learning algorithm for cross-view dimensionality reduction is proposed in \cite{sparseMV}. A neural network which extracts maximally correlated representations is studied in \cite{deepcca}. A review comprehensive article \cite{MVreview} presents recent progress and challenges in this domain.


In this work we extend the concepts established in \cite{DeSa,lindenbaum2015learning} to devise a diffusion-based learning framework for fusing multiple views, by seeking the implied underlying low-dimensional structure in the ambient space.
Our contributions can be summarized as follows.
\begin{enumerate}
\item
We present a natural generalization of the DM method \cite{Lafon} for handling multiple views. The generalization is attained by combining the intrinsic relations
within each view with the mutual relations between views, so as to construct a multi-view kernel matrix. The proposed kernel defines a cross-view diffusion process, and related diffusion distances, which impose a structured random walk between the various views. In addition, we show that the spectral decomposition of the proposed kernel can be used to find an embedding of the data that offers improved exploitation of the information encapsulated in the multiple views.  
\item
For the coupled views setting, we analyze theoretical properties of the proposed method. The associated parametrization is justified by being recast as a minimizer of a kernel-based objective function. Then, we relate the spectrum of the proposed kernel to the spectrum of a single-view kernel which is based on na\"{i}ve concatenation of the views. Finally, under some simplification, we find the infinitesimal generator of the proposed kernel.
\item 
For practical use, we present an automated method for setting the kernel bandwidth parameter and a proposed procedure for recursively augmenting the representation with each new data point.  
\item 
We demonstrate the applicability of our multi-view method to classification, clustering and manifold learning. Furthermore, by fusing data from multiple seismic sensors we demonstrate an automatic extraction of latent seismic parameters using real-world data.

\end{enumerate}

The paper is structured as follows. Some essential background is provided in
section \ref{setup}. In section \ref{SecMulti} we formulate the multi-view dimensionality reduction problem and discuss prior work. Then, in Section \ref{OurProposed} we present our proposed multi-view DM method and discuss some of its basic properties. Section \ref{sec:CoupledV} studies additional theoretical properties of the proposed kernel for the particular (more simple) case of a coupled setting (where only two views are available). Section
\ref{SecExp} presents the experimental results. Potential applications are described in Section \ref{sec:application} and concluding remarks appear in Sections \ref{SecConc} and \ref{sec:appendix} (respectively).

\section{Background}
\label{setup}
\subsection{General dimensionality reduction}
Consider a high-dimensional dataset $\mX=\{\vx_1,\vx_2,\ldots,\vx_M\}\in\Rset^{D\times M}$, $\vx_i\in\Rset^D, i=1,2,\ldots,M$. The goal is to find a low-dimensional representation $\mZ=\{\vz_1,\vz_2,\ldots,\vz_M\}\in\Rset^{r\times M}$, $\vz_i\in\Rset^r, i=1,2,\ldots,M$, such that $r\ll D$, while latent ``inner relations" (if any) among the multidimensional data points are preserved as closely as possible, in some sense. This problem setup is based on the assumption that the data is represented (viewed) in a single vector space (single view).

\subsection{Diffusion Maps (DM)}
\label{SecDiff} DM \cite{Lafon} is a dimensionality reduction method which aims at extracting the intrinsic geometry of the data. The DM framework is highly effective when the data is densely sampled from some low-dimensional manifold, so that its ``inner relations" are the local connectivities (or proximities) on the manifold. Given a high-dimensional dataset $\mX$, the DM framework consists of the following steps:

\begin{enumerate}
\item A kernel function $\nK:\mX\times \mX\rightarrow\Rset$ is chosen, so as to construct a matrix $\mK\in\Rset^{M\times M}$ with elements $K_{ij}\defeq\nK(\vx_i,\vx_j)$, satisfying the following properties: (i) Symmetry: $\mK=\mK^\tps$; (ii) Positive semi-defeniteness: $\mK\succeq\mo$, namely $\forall\vv\in\Rset^M:\;\;\vv^\tps\mK\vv\ge 0$; and (iii) Non-negativity: $\mK\ge\mo$, namely $K_{i,j}\ge 0$ $\forall i,j\in[1,M]$. These properties guarantee that the matrix $\mK$ has real-valued eigenvectors and non-negative eigenvalues. A {\em Gaussian kernel} is a common example, in which $\nK(\vx_i,\vx_j)\defeq \exp\left\{-\frac{\|\vx_i-\vx_j\|^2}{2\sigma_x^2}\right\}$, where $\|\cdot\|$ denotes the $L_2$ norm and $\sigma_x^2$ is a user-selected width (scale) parameter.
\item By normalizing the rows of $\mK$, the matrix
    \begin{equation}
    \mP^x\defeq\mD^{-1}\mK\in\Rset^{M\times M}
    \end{equation}
    is obtained, where $\mD\in\Rset^{M\times M}$ is a diagonal matrix with $D_{i,i}=\sum_jK_{i,j}$. $\mP^x$ can be interpreted as the transition probabilities of a (fictitious) Markov chain on $\mX$, such that $\left[(\mP^x)^t\right]_{i,j}\defeq p_t(\vx_i,\vx_j)$ (where $t$ is an integer power) describes the implied probability of transition from point $\vx_i$ to point $\vx_j$ in $t$ steps.
	\item {Spectral decomposition is applied to $\mP^x$, obtaining a set of $M$ eigenvalues $\{\lambda_m\}$ (in descending order) and associated normalized eigenvectors $\{\vpsi_m\}$ satisfying $\mP^x\vpsi_m=\lambda_m\vpsi_m$, $m=0,\ldots,M-1$, or $\mP^x\mPsi=\mPsi\mLam$, where $\mPsi$ and $\mLam$ are (resp.) the eigenvectors and diagonal eigenvalues matrices.}
	\item{
	A new representation is defined for the dataset $\mX$, representing each $\vx_i$ by the $i$-th row of $(\mP^x)^t\mPsi=\mPsi\mLam^t$, namely:
    \begin{equation}
    \mPsi_t(\vx_i):\;\;\vx_i\mapsto\left[\lambda_1^t\psi_1[i], \ldots,\lambda_{M-1}^t\psi_{M-1}[i]\right]^\tps\in\Rset^{M-1},
    \;\;i\in[1,M]
    \end{equation}
       where $t$ is a selected number of steps and $\psi_m[i]$ denotes the $i$-th element of $\vpsi_m$. Note that the trivial eigenvector $\myvec{\psi}_0=\myvec{1}$ (with corresponding eigenvalue $\lambda_0=1$) was omitted from the representation as it does not carry information about the data. The main idea behind this representation is that the Euclidean distance between two data points in the new representation equals a weighted $L_2$ distance between the conditional probability vectors $p_t(\vx_i,:)$ and $p_t(\vx_j,:)$, $i,j\in[1,M]$ (the $i$-th and $j$-th rows of $(\mP^x)^t$). This weighted Euclidean distance is referred to as the {\em diffusion distance}, denoted
    \begin{multline}
    \label{EqDist}
    \nD_t^2(\vx_i,\vx_j)\defeq\|\mPsi_t(\vx_i)-\mPsi_t(\vx_j)\|^2\\
    =\sum_{m=1}^{M-1}\lambda_m^{2t}(\psi_m[i]-\psi_m[j])^2=\|p_t(\vx_i,:)-p_t(\vx_j,:)\|_
    {{\boldmath W}^{-1}}^2,
    \end{multline}
    where $\mW=\mD/\Tr\{\mD\}$ and where $\|\vz^T\|_{\boldmath Q}\defeq \vz^\tps\mQ\vz$ denotes a $\mQ$-weighted Euclidean norm (the last equality is shown in \cite{Lafon}).}
	\item{A desired accuracy level $\delta\ge 0$ is chosen for the diffusion distance defined by Eq.\ \eqref{EqDist}, such that $r(\delta,t)=\max\{\ell:|\lambda_\ell|^t>\delta|\lambda_1|^t\}$. The new (truncated) $r(\delta,t)$-dimensional mapping, which leads to the desired representation $\mZ$, is then defined as
    \begin{equation}
    \mPsi_t^\delta(\vx_i):\;\;\vx_i\mapsto\left[\lambda_1^t\psi_1[i],\lambda_2^t\psi_2[i].\ldots,\lambda_r^t\psi_r[i]\right]^\tps\defeq\vz_i\in\Rset^{r(\delta,t)}.
    \end{equation} }
\end{enumerate}
This dimensionality reduction approach was found to be useful in various applications in diverse fields. However, as previously noted, it is limited to a single-view representation.

\section{Multi-view dimensionality reduction - Problem formulation and prior work}
\label{SecMulti}
Assume now that we are given multiple sets (views) of observations  $\mymat{X}^\ell\text{ }, \ell=1,...,L$. Each
view is a high-dimensional dataset ${\mymat{X}^\ell = \lbrace{ {
			\myvec{x}_1^\ell,\myvec{x}_2^\ell,...,\myvec{x}_M^\ell  }}
	\rbrace \in {\mathbb{R}^{D_\ell\times M}} }$, where $D_\ell$ is the dimension of each feature space. Note that a bijective correspondence between views is assumed. These views can be a result of different measurement devices or different types of features extracted from raw data. For each view $\ell=1,...,L$, we seek a lower-dimensional representation that preserves the ``inner relations" between multidimensional data points within each view $\myvec{X}^\ell$, as well as among all views $\{ \mymat{X}^1,...,\mymat{X}^L\}$. Our goal is to find $L$ representations $\myvec{\Phi}_\ell(\mymat{X}^1,...,\mymat{X}^L):\Rset^{D_\ell}\rightarrow {\mathbb{R}^{r}}  $, such that $r \ll D_\ell,\;\;\ell=1,...,L$.

Before turning to present our proposed approach, we describe existing alternative approaches for incorporating multiple views. In particular, we detail here the methods which we would later use as benchmarks for comparison to our proposed framework. All of these methods begin by computing the kernel matrix $\mK^\ell$ for each ($\ell$-th) view ($\ell=1,\ldots L)$, and differ in the way of combining these kernels for generating the DM.
\subsection{Kernel Product DM (KP)}
\label{sec:KernelProd}
A na\"{i}ve generalization of DM \cite{Lafon} may be computed using an element-wise kernel product, namely by constructing  $\mymat{K}^{{\circ}}\defeq \mymat{K}^1
\circ \mymat{K}^2 \circ ...\circ \mymat{K}^L\in\Rset^{M\times M}$, where $\circ$ denotes Hadamard's (element-wise) matrix product, such that ${K}_{i,j}^{\circ}\defeq {K}_{i,j}^1
\cdot {K}_{i,j}^2\cdot...\cdot {K}_{ij}^L$, followed by row-normalization: $\mP=\left(\mD^\circ\right)^{-1}\mK^\circ\in\Rset^{M\times M}$,
where $\mD^\circ$ is diagonal, with $D_{i,i}^{\circ}\defeq\underset{j}{\sum}{K_{i,j}^{\circ}}$. 

Note that in the special case of using Gaussian kernels with equal width parameters $\sigma$, the resulting matrix $\mymat{K}^{\circ}$ is
	equal to the matrix $\mymat{K}^w$ constructed from the concatenated observations vector
	\begin{equation}
	    \myvec{w}_{i}=\left[\left(\myvec{x}^1_{i}\right)^\tps,..., \left(\myvec{x}^L_{i}\right)^\tps\right]^\tps
	\end{equation} 
	such that 
	\begin{equation}
	\label{EquationPDH}
	K_{i,j}^w=\exp\left\{ -\frac{\left\|\myvec{w}_i-\myvec{w}_j\right\|^2}
	{2\sigma^2}\right\}.
	\end{equation}

\subsection{Kernel Sum DM (KS)}
An average diffusion process was used in \cite{Zhou}, where the {\em sum kernel} is defined as
\begin{equation}
\label{EquationPDS}
\mymat{K}^{+}\defeq\sum^L_{\ell=1}\mymat{K}^\ell,
\end{equation}
followed by row-normalization: $\mP^+=\left(\mD^+\right)^{-1}\mK^+$, where $D^+_{i,i}=\underset{j}{\sum}{K_{i,j}^{+}}$.
The implied random walk sums up (and normalizes) the step probabilities from each view.

\subsection{{Kernel Canonical Correlation Analysis (KCCA)}} 
The frameworks \cite{lai2000kernel,bach2002kernel} extend the well-known Canonical Correlation Analysis  (CCA) by applying a kernel function prior to the application of CCA. Kernels $\myvec{K}^1$ and $\myvec{K}^2$ are constructed for each view, and the canonical vectors $\myvec{v}_1$ and $\myvec{v}_2$ are obtained by solving the following generalized eigenvalue problem
\begin{equation} \label{eq:KCCA}
\begin{bmatrix}  \mymat{0}_{M \times M} & {\mymat{K}^1\cdot \mymat{K}^2} \\ {{\mymat{K}^2\cdot \mymat{K}^1}} & \mymat{0}_{M \times M} \end{bmatrix} \begin{pmatrix}
{\myvec{v}_1} \\ {\myvec{v}_2}
\end{pmatrix}= \rho \cdot \begin{bmatrix}  (\mymat{K}^1+\gamma\myvec{I})^2 & {\mymat{0}_{M \times M}} \\ {\mymat{0}_{M \times M}} &(\mymat{K}^2+\gamma\myvec{I})^2  \end{bmatrix} \begin{pmatrix}
{\myvec{v}_1} \\ {\myvec{v}_2}
\end{pmatrix},
\end{equation} where $\gamma \myvec{I}$ are regularization terms added to prevent overfitting and thus improve generalization of the method. Usually the Incomplete Cholesky Decomposition (ICD) \cite{lai2000kernel,bach2002kernel,kershaw1978incomplete} is used to reduce the run time required for solving (\ref{eq:KCCA}). For clustering tasks, $K$-means clustering is applied to the set of generalized eigenvectors.

\subsection{{Spectral clustering with two views}}
The approach in \cite{DeSa} generalizes the traditional normalized graph Laplacian for two views. Kernels $\myvec{K}^1$ and $\myvec{K}^2$ are computed in each view and multiplied, yielding $\mymat{W}=\myvec{K}^1\cdot \myvec{K}^2$, from which 
\begin{equation} \label{EQKDeSa} \mymat{A}\defeq \begin{bmatrix}  \mymat{0}_{M \times M} & {\mymat{W}} \\ {\mymat{W}^T} & \mymat{0}_{M \times M} \end{bmatrix}\in\Rset^{2M\times 2M} \end{equation}
is obtained. Symmetric normalization is applied by using the diagonal matrix $\mymat{\bar{D}}$ with diagonal elements
\begin{math}
{\bar{D}}_{i,i}=\underset{j}{\sum}{{{W}}_{i,j}}
\end{math}, such that the normalized fused kernel is defined as \begin{equation} \label{eq:DeSa}
\myvec{\bar{A}}=\myvec{\bar{D}}^{-0.5}\cdot\myvec{A}\cdot \myvec{\bar{D}}^{-0.5}.
\end{equation}
Assuming that the data can be clustered into $N_C$ clusters, and denoting the $2M$ eigenvectors of $\mymat{\bar{A}}$ as $\myvec{\phi}_i, i=1,...,2M$ (in descending order of the corresponding eigenvalues), the mapping
\begin{equation}\label{eq:DeSa2}
\mPhi[i] \defeq \frac{1}{s[i]}\left[\phi_1[i],...,\phi_{N_C}[i]\right]^\tps\in\Rset^{N_C}
\end{equation} (where $s[i]= \sum^{N_C}_{j=1}(\phi^2_j[i])$) is useful for applying $K$-means clustering.
The work in \cite{DeSa} is focused on spectral clustering, however, a similar version of the kernel from Eq.\ (\ref{eq:DeSa}) is suitable for manifold learning, as we demonstrate in the sequel. In section \ref{SecExp} this approach is referred to as de Sa's.

As We show in the next section, our proposed approach essentially uses a stochastic matrix version of this kernel, and extends the construction to multiple views. We shall show that our stochastic matrix version is useful, as it provides various theoretical justifications for the implied multi-view diffusion process.

\section{Our Proposed Approach for Multi-view Diffusion Maps}
\label{OurProposed} 
Here we propose our generalization of the DM framework for handling
a multi-view scenario. This is done by imposing an implied (fictitious) random walk model
using the local connectivities between data points within all
views. Our way to generalize the DM framework is by restraining
the random walker to ``hop'' between different views in each step. The construction requires to choose a set of $L$ symmetrical positive
semi-definite, non-negative kernels, one for each view
\begin{math}{{\cal{K}}^l :
	\myvec{X}^l\times{\myvec{X}^l}\rightarrow{\Rset} \text{ } ,\;\;l=1,...,L
}
\end{math}. We use the Gaussian kernel function, so that the $(i,j)$-th element of each $\mK^l\in\Rset^{M\times M}$ is given by 
\begin{equation} \label{EQK}{ K^l_{i,j}=\exp\left\{
	-\frac{||\myvec{x}^l_i-\myvec{x}^l_j||^2}{2 \sigma_l^2}\right\}  },\;\;\;i,j=1,\ldots,M,\;\;l=1,\ldots,L,
\end{equation} 
where $\left\{\sigma_l^2\right\}_{l=1}^L$ are a set of selected parameters.
The decaying property of the Gaussian is useful, as it removes the influence of large Euclidean distances. 
The {\em multi-view kernel} is formed by constructing the following matrix
\begin{equation} \label{EQKMAT}
\mymat{\widehat{K}}= \begin{bmatrix}  \mymat{0}_{M \times M} & {\mymat{K}^1\mymat{K}^2}&  {\mymat{K}^1\mymat{K}^3}&...& {\mymat{K}^1\mymat{K}^L} \\
\mymat{K}^2\mymat{K}^1 & \mymat{0}_{M \times M} & {\mymat{K}^2\mymat{K}^3}&...& {\mymat{K}^2\mymat{K}^L}\\ \mymat{K}^3\mymat{K}^1 & {\mymat{K}^3\mymat{K}^2} &
\mymat{0}_{M \times M} &...& {\mymat{K}^3\mymat{K}^L}\\:&:&:&...&:\\\mymat{K}^L\mymat{K}^1 & {\mymat{K}^L\mymat{K}^2} & {\mymat{K}^L\mymat{K}^3} &...&
{\mymat{0}_{M \times M}}.
\end{bmatrix}\in\Rset^{LM\times LM}. \end{equation}
Finally, using the diagonal row-normalization matrix $\mymat{\widehat{D}}\in\Rset^{LM\times LM}$ with
\begin{math}
{\widehat{D}}_{i,i}=\underset{j}{\sum}{{\widehat{K}}_{i,j}}
\end{math}, the normalized row-stochastic matrix is defined as
\begin{equation}
\label{phat}
\mymat{\widehat{P}}={\mymat{\widehat{D}}}^{-1}\mymat{\widehat{K}}\in\Rset^{LM\times LM}.
\end{equation}
We refer to the $(l,m)$-th block of $\widehat{\mP}$ as the square $M\times M$ matrix starting at\\
$[1+(l-1)M,1+(m-1)M],\;\;l,m=1,...,L$. Thus, the $(i,j)$-th element of the $(l,m)$-th block describes a (fictitious) probability of
transition from $\vx_i^l$ to $\vx_j^m$. This construction takes into account all possibilities to ``hop'' between views, under the constraint that staying in the same view (namely, a transition from $\vx_i^l$ to $\vx_j^l$) is forbidden.

\subsection{Probabilistic interpretation of $\widehat{\mymat{P}}^t$}
\label{sec:prob}
Subsequent to our proposed construction (Eqs.\ (\ref{EQK}), (\ref{EQKMAT}) and
(\ref{phat})), each element of the power-$t$ matrix $\widehat{\mP}^t$,
\begin{equation}
\label{eq:genelement}
\left[\widehat{\mP}^t\right]_{i+(l-1)M,j+(m-1) M}\defeq\widehat{p}_t(\myvec{x}^l_i,\myvec{x}^m_j)
\end{equation}
denotes the probability of transition from $\vx_i^l$ to $\vx_j^m$ in $t$ time-steps. Note that, as mentioned above, due to the block-off-diagonal structure of $\widehat{\mP}$, which forbids a transition into the same view within a single time-step, we have $\widehat{p}_1(\myvec{x}^l_i,\myvec{x}^l_j)=0,\;\;l=1,\ldots,L$, although for $t>1$ $\widehat{p}_t(\myvec{x}^l_i,\myvec{x}^l_j)$ may be non-zero.

\subsubsection{Smoothing effect $t=1$}
For simplicity let us examine the term $\widehat{p}_t(\myvec{x}^l_i,\myvec{x}^m_j),l\neq m$ for $t=1$. The transition probability for $t=1$ is \[{{\widehat{p}}}_1(\myvec{x}^l_i,\myvec{x}^m_j)=\frac{\sum_{s}{K^l_{i,s}}K^m_{s,j}}{{\widehat{D}_{i,i}}}.\]
This probability takes into consideration all the various connectivities of node $\myvec{x}^l_{i}$ to node $\myvec{x}^l_{s}$ and the connectivities of the corresponding node $\myvec{x}^m_{s}$ to the destination node $\myvec{x}^m_{j}$. The proposed multi-view approach has a ``smoothing effect" in terms of the transition probabilities, meaning that the probability of transitioning from $\myvec{x}^l_i$ to $\myvec{x}^m_j$ could be positive even if $K^l_{i,j}=0$ and $K^m_{i,j}=0$: Assume that a non-empty subset ${\cal{S}}=\{s_1,...,s_F \}\subseteq [1,M]$ exists, such that $K^l_{i,s_f}>0$ and $K^m_{s_f,j}>0,f=1,..,F$. Then by definition of the multi-view probability we have ${\widehat{p}}_1(\myvec{x}^l_i,\myvec{x}^m_j)>0$. Note that although with the Gaussian kernel all elements of $\mK^l$ are positive, the smoothing effect can still be significant when despite negligibly low values in $K^l_{i,j}$ and in $K^m_{i,j}$, ${\widehat{p}}_1(\myvec{x}^l_i,\myvec{x}^m_j)$ can become relatively high.

\begin{figure}[ht]
	\centering
	
	{\includegraphics[width=6cm]{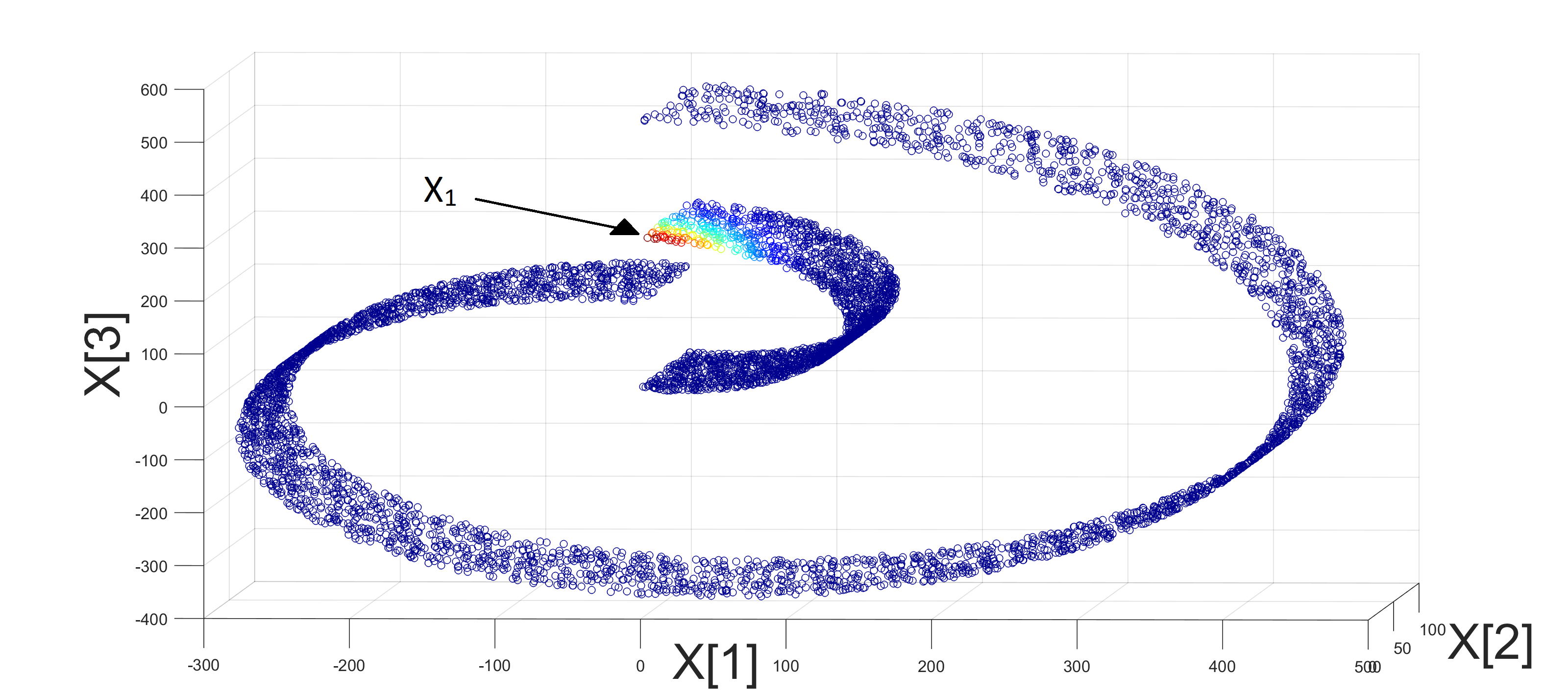}}
	{\includegraphics[width=6cm]{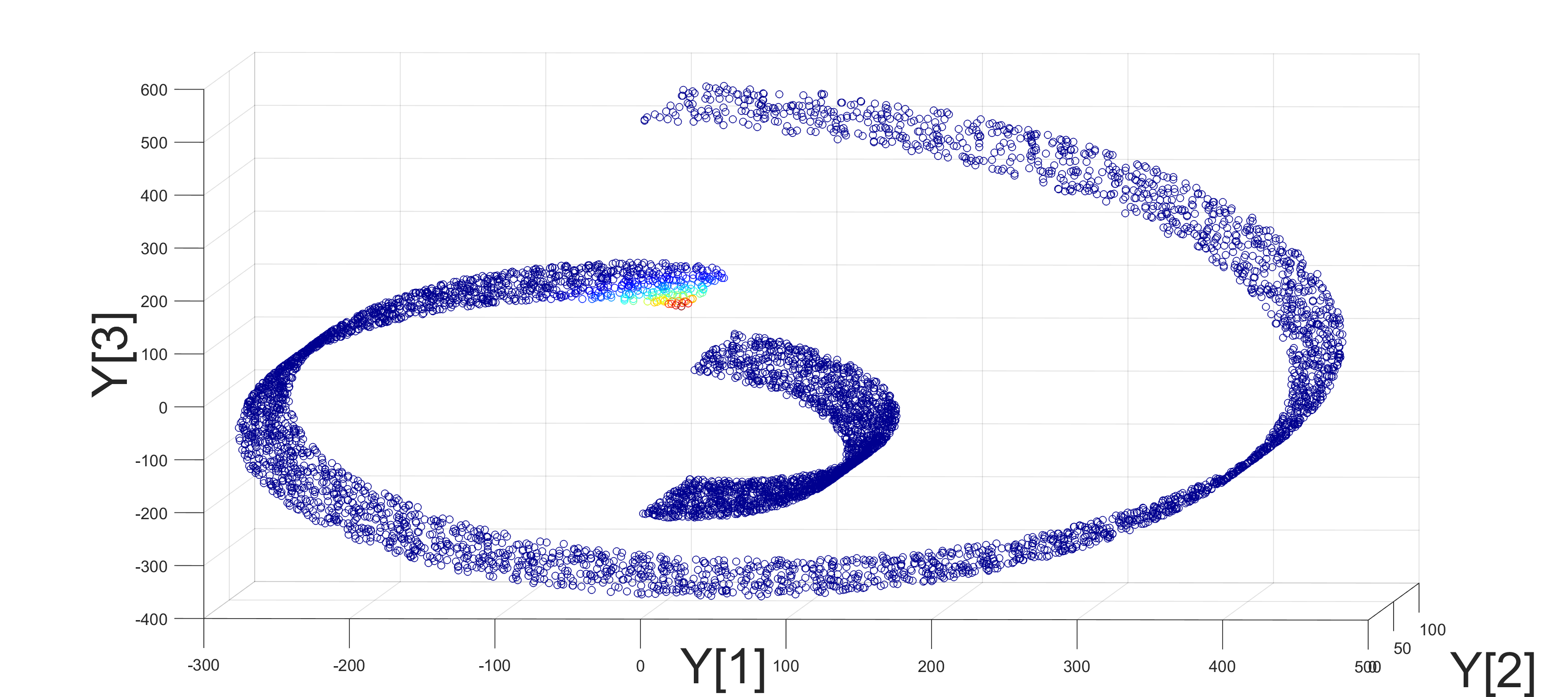}}
	{\includegraphics[width=6cm]{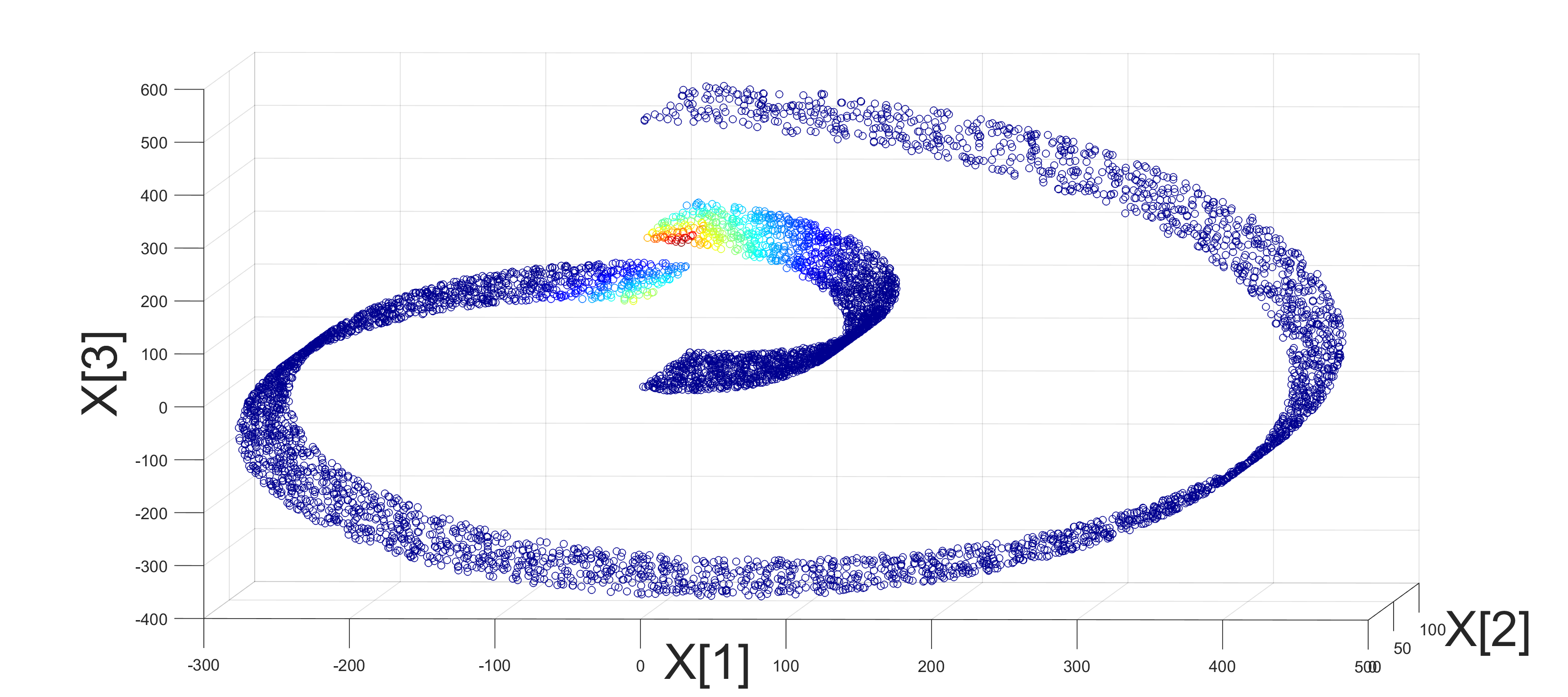}}
	{\includegraphics[width=6cm]{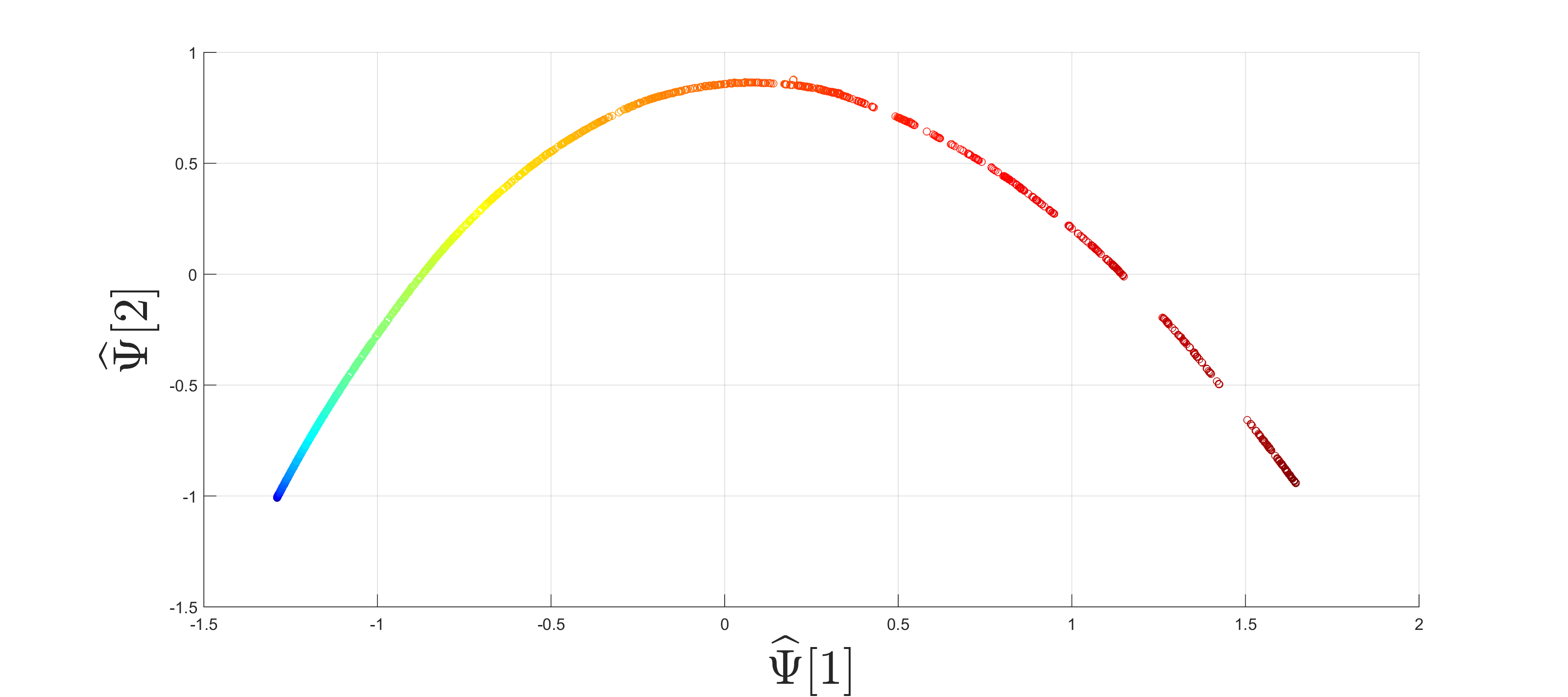}}
	\caption{Top-left: A non-smooth Swiss Roll sampled from View-I ($\myvec{X}^1$),
		colored by the single-view probability of transition ($t=1$) from $\myvec{x}_1$ to
		$\myvec{x}_:$. Top-right: A second Swiss Roll, sampled from View-II ($\myvec{X}^2$), colored by the single-view probability of transition ($t=1$) from $\myvec{y}_1$ to
		$\myvec{y}_:$. Bottom-left: the first Swiss Roll, colored by the multi-view probabilities of transition ($t=1$) from $\myvec{x_i}$ to $\myvec{y}_:$. In the top-left figure, we highlight $\myvec{x}_1$ using an arrow. Bottom-right: a low-dimensional representation extracted based on the multi-view transition matrix $\widehat{\myvec{P}}$ (Eq.\ \ref{phat}).} \label{Fig1}
\end{figure}

Figure \ref{Fig1} illustrates the multi-view transition probabilities compared to a single-view approach using two ($L=2$) deformed
``Swiss Roll" manifolds. In each view, there is no probability of transition from one side of its observed ``gap" to the other side. And yet, the multi-view (one-step) transition probability is non-zero for points at both sides of the gaps. This smoothing effect occurs because the gaps are located at a different position (near different points) on each view, thus allowing the multi-view kernel to smooth out the nonlinear gaps. 

\subsubsection{Increasing the diffusion step $t$}
\begin{figure}[ht]
	\centering
	
	{\includegraphics[width=8cm]{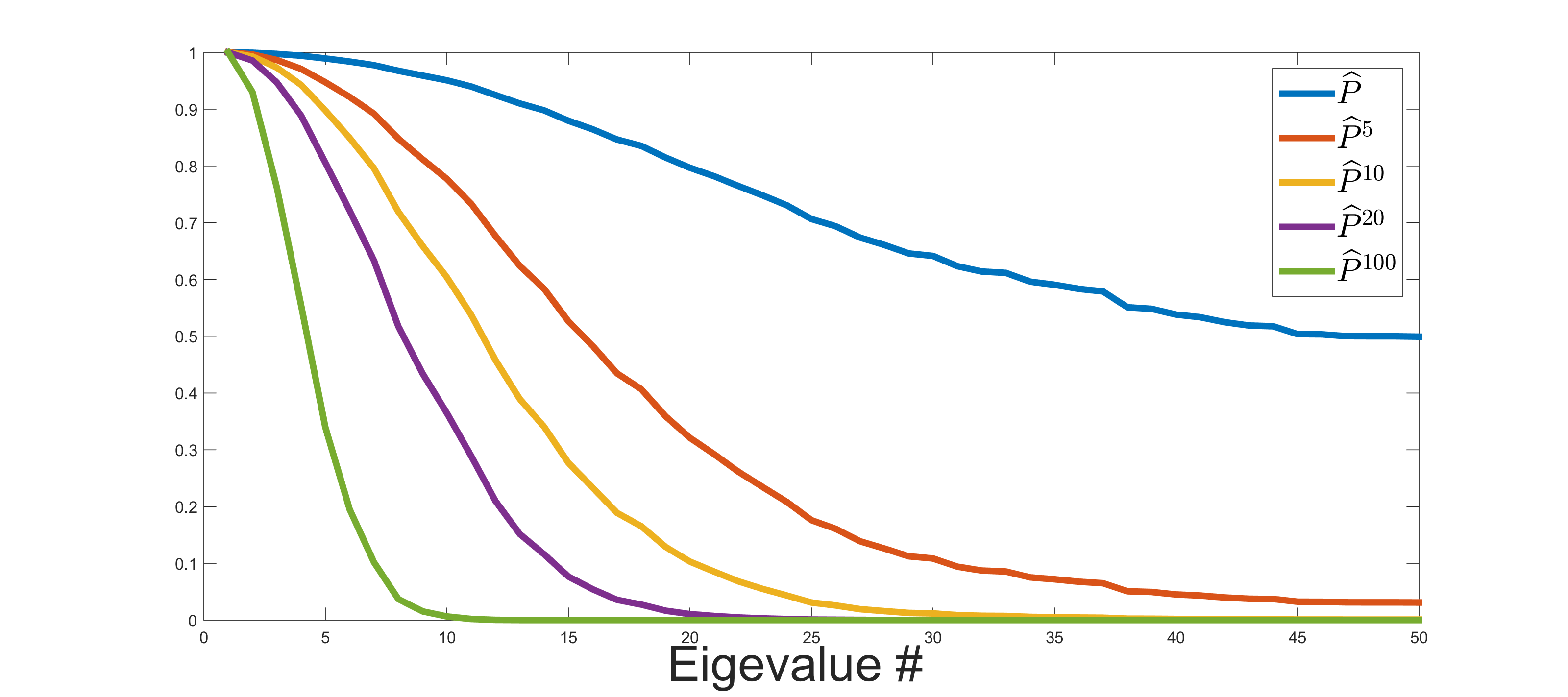}}
	
	\caption{The decay of the eigenvalues for increasing powers of the matrix $\widehat{\myvec{P}}$.} \label{FigTime}
\end{figure}

Under the stochastic Markov model assumption, raising  $\widehat{\myvec{P}}$ to a higher power (by increasing the diffusion step $t$) spreads the probability mass function along its rows based on the connectivities in all views. As described in \cite{Lafon}, this probability spread reduces the influence of eigenvectors associated with high-indexed (smaller) eigenvalues on the diffusion distance (Eq.\ (\ref{DIV1})). This implies that the eigenvectors corresponding to low-indexed (large) eigenvalues have a low-frequency content, whereas the eigenvectors corresponding to the high-indexed (small) eigenvalues describe the oscillatory behavior of the data \cite{Lafon}. In Fig.\ \ref{FigTime}, we present the eigenvalues of the matrix $\widehat{\myvec{P}}^t$ with different values of $t$. For this experiment we generated $L=3$ Swiss Rolls with $M=1,\!200$ data points each. It is evident that the numerical rank of $\widehat{\myvec{P}}^t$ decreases for higher values of $t$.

\subsection{Multi-view diffusion distance}
\label{multi viewDiffusionDistance} In a variety of real-world data
types, the Euclidean distance between observed data points (vectors) does not provide sufficient
information about the intrinsic relations between these vectors, and is highly sensitive to non-unitary transformations. Common tasks such as classification, clustering or system
identification often require a measure for the intrinsic connectivity
between data points, which is only locally expressed
by the Euclidean distance in the high-dimensional ambient space. The
multi-view diffusion kernel (defined in section \ref{OurProposed})
describes all the small local connections between data points.
The row stochastic matrix $\mymat{\widehat{P}}^t$ (Eq.\ (\ref{phat}))
accounts for all possible transitions between data points in $t$ time steps while hopping between the views. For a fixed value $t>0$, two
data points are \textit{intrinsically similar} if the conditional
distributions
${\widehat{\myvec{p}}}_t(\myvec{x}_i,:)=\myvec{{[\widehat{P}^t]}}_{i,:}$
and
${\widehat{\myvec{p}}}_t(\myvec{x}_j,:)=\myvec{{[\widehat{P}^t]}}_{j,:}$
are similar. This type of similarity measure indicates that the
points $\myvec{x_i}$ and $\myvec{x_j}$ are similarly connected (in the sense of similar probabilities of transitions) to several
mutual points. Thus, they are connected by some geometrical path. In
many cases, a small Euclidean distance can be misleading due to
the fact that two data points can be ``close'' without having any
geodesic path that connects them. Comparing the
transition probabilities is more robust, as it takes into consideration all of the local connectivities between the compared points. Therefore, even if two points seem far from one another in the Euclidean sense, they may still share many common neighbors and thus be ``close" in the sense of having a small diffusion distance.

Based on this observation, by expanding the single-view
construction given in \cite{Lafon}, we define the weighted inner view
diffusion distances for the first view as
\begin{equation}\label{DIV1}
{\mathcal D}_t^2(\ux_i^1,\ux_j^1)\defeq
		\sum_{k=1}^{L\cdot M}
		\frac{1}{\tilde{\phi}_0(k)}
		\left(\left[\widehat{\mP}^t\right]_{i,k}-
		\left[\widehat{\mP}^t\right]_{j,k}\right)^2
		=
		\left\|(\ue_i-\ue_j)^\tps\widehat{\mP}^t\right\|^2_{\scriptsize {\widehat{\mymat{D}}^{-1}}},
\end{equation} 
where $1\leq i,j \leq M$, $\myvec{{e}}_i$ is the $i$-th column of an $L\cdot M\times L \cdot M$ identity matrix, $\tilde{\myvec{\phi}}_0$ is the (non-normalized) first left eigenvector of $\mymat{\widehat{P}}$, whose $k$-th element is  \begin{math}
\tilde{\phi}_0(k)={\widehat{D}_{k,k}} \end{math}. 
A similarly weighted norm is defined for the $l$-th view
\begin{equation} \label{DIV2}
{\mathcal D}_t^2(\ux_i^l,\ux_j^l)\defeq
		\sum_{k=1}^{L\cdot M}
		\frac{1}{\tilde{\phi}_0(k)}
		\left(\left[\widehat{\mP}^t\right]_{i+\tilde{l},k}-
		\left[\widehat{\mP}^t\right]_{j+\tilde{l},k}\right)^2
		=
		\left\|(\ue_{\tilde{l}+i}-\ue_{\tilde{l}+j})^\tps\widehat{\mP}^t\right\|^2_{\scriptsize {\widehat{\mymat{D}}^{-1}}},
\end{equation} 
where $\tilde{l}=(l-1)\cdot M$. The main advantage of these distances (Eqs.\ (\ref{DIV1})
and (\ref{DIV2})) is that they can be expressed in terms of the
eigenvalues and eigenvectors of the matrix
$\mymat{\widehat{P}}$. This insight allows us to use a
representation (defined in section \ref{MVparapmeterization} below) where
the induced Euclidean distance is proportional to the diffusion
distances defined in Eqs.\ (\ref{DIV1}) and (\ref{DIV2}). Indeed, let $\widehat{\mP}=\mPsi\mLam\mPhi^\tps$ denote the eigenvalues decomposition of $\widehat{\mP}$, where $\mPsi=\left[\psi_0,\ldots,\psi_{L\cdot M-1}\right]$ and $\mPhi=\left[\phi_0,\ldots,\phi_{L\cdot M-1}\right]$ denote the matrices of (normalized) right- and left-eigenvectors (resp.), and $\mLam\defeq\Diag(\lambda_0,\ldots,\lambda_{L\cdot M-1})$ denotes the diagonal matrix of respective eigenvalues. Then

\begin{T1} \label{T2}
	The inner view diffusion distance defined by Eqs.\ (\ref{DIV1}) and
	(\ref{DIV2}) can also be expresses as
	\begin{equation}\label{DIV3}{ { {{{\cal{D}}_t^2}}( \myvec{x}^l_i,\myvec{x}^l_j)=\sum_{{k}=1}^{L\cdot M-1}\lambda_{k}^{2t}\left(\psi_{k}\left[i+\tilde{l}\right]-
	\psi_{k}\left[j+\tilde{l}\right]\right)^2} , i,j=1,...,M},
	\end{equation} 
	where $\tilde{l}=(l-1)\cdot M$.
\end{T1}
\begin{proof}
Using the eigenvalues decomposition of $\hmP$ we have
\begin{equation}
    \hmP^t\hmD^{-1}\left(\hmP^t\right)^\tps=\mPsi\mLam^t\mPhi^\tps\hmD^{-1}\mPhi\mLam^t\hmPsi^\tps.
\end{equation}
Define the symmetric matrix $\hmP_s\defeq\hmD^{-1/2}\hmK\hmD^{-1/2}=\hmD^{1/2}\hmP\hmD^{-1/2}$, and note that this matrix is algebraically similar to $\hmP$. Therefore, both matrices share the same set of eigenvalues $\mLam$. Additionally, let $\mPi$ denote the (left- and right-) orthonormal eigenvectors matrix of $\hmP_s$, namely $\hmP_s=\mPi\mLam\mPi^\tps$. The left- and right-eigenvectors matrices of $\hmP=\hmD^{-1/2}\hmP_s\hmD^{1/2}$ are then easily identified as $\mPhi=\hmD^{1/2}\mPi$ and $\mPsi=\hmD^{-1/2}\mPsi$ (resp.), so that $\mPhi^\tps\hmD^{-1}\mPhi=\mPi^\tps\mPi=\mI$. Therefore,
\begin{multline}
{\mathcal D}_t^2(\ux_i^l,\ux_j^l)
		=\left\|(\ue_{\tilde{l}+i}-\ue_{\tilde{l}+j})^\tps\hmP^t\right\|^2_{\scriptsize{\widehat{\mymat{D}}^{-1}}}
		=(\ue_{\tilde{l}+i}-\ue_{\tilde{l}+j})^\tps\hmP^t\hmD^{-1}\left(\hmP^t\right)^\tps(\ue_{\tilde{l}+i}-\ue_{\tilde{l}+j})\\
		=(\ue_{\tilde{l}+i}-\ue_{\tilde{l}+j})^\tps\mPsi\mLam^{2t}\mPsi^\tps(\ue_{\tilde{l}+i}-\ue_{\tilde{l}+j})
		=\sum_{{k}=1}^{L\cdot M-1}\lambda_{k}^{2t}\left(\psi_{k}\left[i+\tilde{l}\right]-
	\psi_{k}\left[j+\tilde{l}\right]\right)^2.
\end{multline}
	The term with $k=0$ can be excluded from the sum since $\myvec{\psi_0=\myvec{1}}$ (an all-ones vector) is always the first right-eigenvector (with eigenvalue $1$) for any stochastic matrix, so the term corresponding to $k=0$ in the sum would vanish anyway.
\end{proof}

\subsection{Multi-view Data Parametrization}
\label{MVparapmeterization}
Tasks such as classification, clustering
or regression in a high-dimensional feature space are
considered to be computationally expensive. In addition, the
performance in such tasks is highly dependent on the distance
measure used. As explained in section \ref{multi viewDiffusionDistance}, distance measures in the original
ambient space are often meaningless in many real life situations.
Interpreting Theorem \ref{T2} in terms of Euclidean distance enables
us to define mappings for every view $\mymat{X}^l,l=1,...,L$, using
the right eigenvectors of $\widehat{\mymat{P}}$ (Eq.\ (\ref{phat}))
weighted by $\lambda^t_i$. The representation for instances in $\mymat{X}^l$ is given
by
\begin{equation}
\label{Map1}
{ \myvec{\widehat{\Psi}}_t{(\myvec{x}^l_i)}:   \myvec{x}^l_i
	\longmapsto
	\begin{bmatrix} { \lambda_1^{t}\psi_1[i+\tilde{l}]} , {.} {.} {.} ,
	
	{\lambda_{M-1}^{t}\psi_{M-1}[i+\tilde{l}]}
	
	\end{bmatrix}^T \in{\mathbb{R}^{M-1}} },
\end{equation} where $\tilde{l}=(l-1)\cdot M$.
These $L$ mappings capture the intrinsic geometry of the views as well
as the mutual relations between them. As shown in \cite{LafonPHD}, the set of eigenvalues ${\lambda_m}$ has a decaying property
such that
\begin{math}{1=\lvert{\lambda_0} \rvert \geq
	\lvert{\lambda_1}\rvert\geq\text{...}\geq\lvert{\lambda_{M-1}}\rvert
}
\end{math}. Exploiting the decaying property enables us to represent data up to a dimension $r$ where $r \ll D_1,...,D_l$. The dimension $r \equiv r(\delta)$ is
determined by approximating the diffusion distance (Eq.\ (\ref{DIV3}))
up to a desired accuracy $\delta$ (we elaborate on this issue in
section \ref{SECDEC}). The reduced dimension version of ${ \myvec{\widehat{\Psi}}_t{(\myvec{X})}}$ is denoted by ${ \myvec{\widehat{\Psi}}^r_t{(\myvec{X})}}$.

Using the inner view diffusion distances defined in Eqs.\
(\ref{DIV1}) and  (\ref{DIV2}), we define a multi-view
diffusion distance as the sum of inner views
distances,
\begin{equation}{ {
		{{{\cal{D}}_t}^{(MV)}}^2(i,j)\defeq\sum_{l=1}^L\left|{\myvec{\widehat{\Psi}}^r_t{(\myvec{x}^l_i)}}-{\myvec{\widehat{\Psi}}^r_t{(\myvec{x}^l_j)}}\right|^2
}} .
\end{equation}
This distance is the induced Euclidean distance in a space constructed from the concatenation of all low-dimensional multi-view mappings
\begin{equation}\widehat{\myvec{\Psi}}_t({\myvec{X}})=
\left[ \myvec{\widehat{\Psi}}^r_t{(\myvec{X}^1)};\myvec{\widehat{\Psi}}^r_t{(\myvec{X}^2)};...;\myvec{\widehat{\Psi}}^r_t{(\myvec{X}^L)}\right]\in\Rset^{L\cdot r\times M}.
\end{equation} 
This mapping is used in section \ref{sec:Clust} for the experimental evaluation of clustering.

\subsection{Multi-view kernel bandwidth}
\label{sec:setting_epsilon}
When constructing the Gaussian kernels $\mymat{K}^l,l=1,...,L$, in Eq.\ (\ref{EQK}), the values of the scale (width) parameter $\sigma^2_l$ have to be set. Setting these values to be too small
may result in very small local
neighborhoods that are unable to capture the local structures around the data points.
Conversely, setting the values to be too large
may result in a fully connected graph that may generate
a too coarse description of
the data.
Several approaches have been proposed in the literature for determining the kernel width (scale):

In \cite{keller2010audio}, a max-min measure is suggested such that the scale becomes
\begin{equation} \label{eq:MaxMin}
\sigma_{{l}}^2={{C}}\cdot \underset{j}{\max} \left[ \underset{i,i\neq j}{\min} (||\myvec{x}^l_i-\myvec{x}^l_j||^2)\right].
\end{equation}
Based on empirical results, the authors in \cite{keller2010audio} suggest to set ${{C}}$ within the range $[1,1.5]$. The specific choice of $C$ should be sufficient to capture all local connectivities.  This single-view approach could be relaxed in the context of our multi-view scenario. The multi-view kernel $\mymat{\widehat{K}}$ (Eq.\ \ref{EQKMAT}) consists of products of single-view kernel matrices $\mymat{K}^l,l=1,...,L$ (Eq.\ \ref{EQK}). The diagonal values of each kernel matrix $\mymat{K}^l$ are all $1$'s, therefore, in order to have connectivity in all rows of $\mymat{\widehat{K}}$ a connectivity in only one of the views is sufficient. By connectivity, we mean that there is at least one nonzero value in the affinity kernel. This insight suggests that a smaller value for the parameter ${{C}}$ could be used in the multi view setting. In \cite{dov2016kernel}, the authors provide a probabilistic interpretation of the choice of $C$, and present a grid-search algorithm for optimizing the choice of $C$ such that each point in the dataset is connected to at least one other point.

\begin{figure}
	
	\centering
	
	{\includegraphics[width=5.8cm]{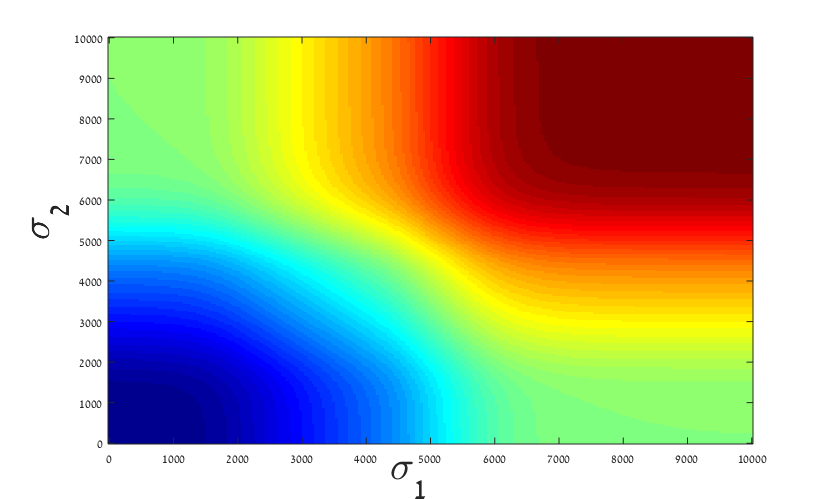}}
	{\includegraphics[width=5.8cm]{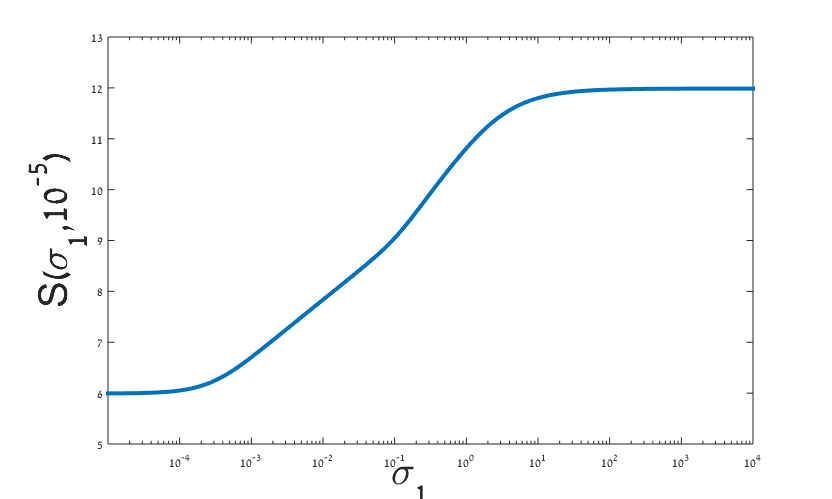}}
	\caption{Left: an example of the two dimensional function $S(\sigma_l,\sigma_m)$. Right: a slice at the first row ($\sigma_2=10^{-5}$). The asymptotes are clearly visible in both figures. Algorithm \ref{alg:Singer} exploits the multi-view to set a small scale parameter for both views. } \label{fig:LogLog}
\end{figure}

Another scheme, proposed in \cite{Singer}, aims to find a range of values for $\sigma_l$. The idea is to compute the kernel $\mymat{K}^l$ (Eq.\ (\ref{EQK})) for various values of $\sigma$ and search for the range of values where the Gaussian bell shape is more pronounced. To find such range of valid values for $\sigma$, first  a logarithmic function is applied to the sum of all elements of the kernel matrix, next the range is identified as the maximal range where the logarithmic plot is linear. We expand this idea for a multi-view scenario based on the following algorithm: 
\begin{algorithm}[H] 
	\caption{Multi-view kernel bandwidth selection}
    \textbf{Input:} Multiple sets of observations (views) $\myvec{X}^l,l=1,...,L $.\\
    \textbf{Output:} Scale parameters $\{ \sigma_1,...,\sigma_L   \}$.
	\begin{algorithmic}[1]
		\STATE Compute Gaussian kernels        \begin{math}{\myvec{K}^l(\sigma_l) } ,l=1,...,L
		\end{math}      for several values of \begin{math} {\sigma}_l\in [10^{-5},10^5] \end{math}.
		\STATE Compute for all pairs $l\neq m$:    \begin{math}{S^{lm}(\sigma_l,\sigma_m)=\underset{i}{{\sum}}\underset{j}{{\sum}}K^{lm}_{i,j}(\sigma_l,\sigma_m) }
		\end{math}, where $\myvec{K}^{lm}(\sigma_l,\sigma_m)=\myvec{K}^{l}(\sigma_l)\cdot \myvec{K}^{m}(\sigma_m) $.
		\FOR {$l=1:L$}
		\STATE Find the minimal value for $\sigma_l$ such that $\myvec{S}^{lm}(\sigma_l,\sigma_m)$ is linear for all $m\neq l$. 
		\ENDFOR
		
	\end{algorithmic}
	\label{alg:Singer}
\end{algorithm}
Note that the two dimensional function $S^{lm}(\sigma_l,\sigma_m)$ consists of two asymptotes, $S^{lm}(\sigma_l,\sigma_m)\overset{\sigma_l,\sigma_m\rightarrow 0}{\longrightarrow} \text{log}(N)$, and $S^{lm}(\sigma_l,\sigma_m)\overset{\sigma_l,\sigma_m\rightarrow \infty}{\longrightarrow} \text{log}(N^3)=3\text{log}(N)$, since for $\sigma_l,\sigma_m\rightarrow 0$, both $\myvec{K}^l$ and $\myvec{K}^m$ approach the Identity matrix, and for $\sigma_l,\sigma_m\rightarrow \infty$, both $\myvec{K}^l$ and $\myvec{K}^m$ approach all-ones matrices.  
An example of the plot $S^{lm}(\sigma_l,\sigma_m)$ for two views ($L=2$) is presented in Fig.\ \ref{fig:LogLog}. The range of each $\sigma_l$ should reflect the asymptotic behaviour, and may be determined empirically by choosing $\min(\sigma_l) \ll median \{ ||\myvec{x}^l_i-\myvec{x}^l_j||\}$ and $\max(\sigma_l) \gg \text{median} \{ ||\myvec{x}^l_i-\myvec{x}^l_j|| \}$. In practice the range $[10^{-5},10^5] $ is sufficient. 
\subsection{Computational Complexity}
Let us now consider the computational complexity required for extracting $\myvec{\Phi}_l(\mymat{X}^1,...,\mymat{X}^L)\in {\mathbb{R}^{r \times M}}  $ from 
${\mymat{X}^l = \lbrace{ {
			\myvec{x}_1^l,\myvec{x}_2^l, \myvec{x}_3^l,...,\myvec{x}_M^l  }}
	\rbrace \in {\mathbb{R}^{D_l\times M}} }$. 
The computational complexity of computing $L$ kernels is ${\cal{O}}(\sum_l M^2 D_l)$. Computing the proposed normalized kernel $\widehat{\myvec{P}}$ (Eq.\ \ref{phat}) adds a complexity of ${\cal{O}} (\frac{L(L-1)}{4} M^3)$. The final step requires the spectral decomposition of $\widehat{\myvec{P}}$, thus it is the most computationally expensive step and adds a complexity of ${\cal{O}} (L^3M^3)$. However, due to the low rank, sparse nature of $\widehat{\myvec{P}}$, this spectral decomposition could be approximated using random projection methods such as in \cite{aizenbud2016matrix,holmes2007fast,holmes2009quic}. Using random projection the complexity of the spectral decomposition is improved to ${\cal{O}}(L^2M^2 \log{r})$. For the optional step proposed in Algorithm \ref{alg:Singer}, if $n_0$ values are chosen for $\sigma$, the additional complexity cost is of $\mathcal{O}(M L n_0)$.

\section{Coupled views $L=2$}
\label{sec:CoupledV}
In this section we provide analytical results for the special case of a coupled data set (i.e $L=2$). Some of the results could be expanded to a larger number of views but not in a straightforward manner.
To simplify the notation in the rest of this section we denote $\myvec{X}\defeq \myvec{X}^1$ and $\myvec{Y}\defeq \myvec{X}^2$.

\subsection{Coupled mapping}
The mappings provided by our approach (Eq.\ (\ref{Map1})) are justified by the relations given by Eq.\
(\ref{DIV3}). In this subsection we provide some intuition on the relation between the mappings of $\myvec{X}$ and $\myvec{Y}$. We focus on the
analysis of a $1$-dimensional mapping for each view. Let
$\myvec{\rho}^x\defeq\myvec{\rho}(\mymat{X})=[\rho(\myvec{x}_1),\rho(\myvec{x}_2),...,\rho(\myvec{x}_M)]^\tps$
and
$\myvec{\rho}^y\defeq\myvec{\rho}(\mymat{Y})=[\rho(\myvec{y}_1),\rho(\myvec{y}_2),...,\rho(\myvec{y}_M)]^\tps$
denote such $1$-dimensional mappings (one for each view) 
and define $\mymat{K}^z\defeq\mymat{K}^x\cdot\mymat{K}^y$, where $\mymat{K}^x, \mymat{K}^y$ are computed from $\mymat{X}$ and $\mymat{Y}$ (resp.) based on Eq.\ (\ref{EQK}).
The following theorem characterizes the desirable $1$-dimensional mappings $\myvec{\rho}^x$ and $\myvec{\rho}^y$.
\begin{T1} \label{T3}

A $1$-dimensional zero-mean representation which minimizes the following objective function
\begin{multline}
\label{EQargMIN}
\min_{\myvec{\rho}^x, \myvec{\rho}^y}
\sum_{i,j} (\rho(\myvec{x}_i)-\rho(\myvec{y}_j))^2{K}_{i,j}^z\\
{\rm s.t.}\;\;\|\myvec{\rho}^x\|^2+\|\myvec{\rho}^y\|^2=1,\;\sum_i\left(\rho(\myvec{x}_i)+\rho(\myvec{y}_i)\right)=0,
\end{multline} 
is obtained by setting $\widehat{\myvec{\rho}}\defeq[\myvec{\rho}^{x\tps}\;\myvec{\rho}^{y\tps}]^\tps=\myvec{\psi}_1$, where $\myvec{\psi}_1$ is the first non-trivial normalized eigenvector of the graph Laplacian $\widehat{\myvec{D}}-\widehat{\myvec{K}}$. The scaling constraint avoids arbitrary scaling of the representation, while the zero-mean constraint (implying orthogonality of $\widehat{\myvec{\rho}}$ to the constant all-ones vector $\myvec{1}$) eliminates any presence of a trivial (constant) component in the representations.
\end{T1}
Note that $\myvec{K}^z$ is an affinity measure of points based on $\myvec{K}^x \cdot\myvec{K}^y$. This means that a small value of ${K}_{i,j}^z$ indicates weak
connectivity between data points $i$ and $j$ (possibly through an intermediate point $\ell\in[1,M]$), implying that the distance between
$\rho(\ux_i)$ and $\rho(\uy_j)$ can be
large. Conversely, if ${K}_{i,j}^z$ is large, indicating
strong connectivity (possibly through an intermediate point) between points $i$ and $j$, the distance between
$\rho(\ux_i)$ and $\rho(\uy_j)$ should be
small when aiming to minimize the objective function.

\begin{proof}
By expanding the objective function in Eq.\ (\ref{EQargMIN}) we get
\begin{multline}
\sum_{i,j}(\rho(\myvec{x}_i)-\rho(\myvec{y}_j))^2{K}_{i,j}^z=
\sum_{i,j}\rho^2(\myvec{x}_i)K_{i,j}^z+
\sum_{i,j}\rho^2(\myvec{y}_j)K_{i,j}^z-
2\sum_{i,j}\rho(\myvec{x}_i)\rho(\myvec{y}_j){K}_{i,j}^z\\
=
\sum_i\rho^2(\myvec{x}_i)\sum_jK^z_{i,j}+
\sum_j\rho^2(\myvec{y}_j)\sum_iK^z_{i,j}-
\sum_{i,j}\rho(\myvec{x}_i)\rho(\myvec{y}_j){K}_{i,j}^z-
\sum_{i,j}\rho(\myvec{x}_j)\rho(\myvec{y}_i){K}_{j,i}^z\\
=
\sum_i\rho^2(\myvec{x}_i)D_{i,i}^{\rm rows}+
\sum_j\rho^2(\myvec{y}_j)D_{j,j}^{\rm cols}-
\sum_{i,j}\rho(\myvec{x}_i)\rho(\myvec{y}_j){K}_{i,j}^z-
\sum_{j,i}\rho(\myvec{x}_j)\rho(\myvec{y}_i){K}_{j,i}^z\\
=\begin{bmatrix}  \mymat{\rho}^{x\tps} & \mymat{\rho}^{y\tps}  \end{bmatrix}\Bigg[ \begin{bmatrix}  \mymat{D}^{\rm rows} & \mymat{0}_{M \times M} \\ \mymat{0}_{M \times M} & \mymat{D}^{\rm cols} \end{bmatrix}- \begin{bmatrix}  \mymat{0}_{M \times M} & {\mymat{K}^z} \\ ({\mymat{K}^z})^\tps & \mymat{0}_{M \times M} \end{bmatrix} \Bigg] \begin{bmatrix}  \mymat{\rho}^x \\ \mymat{\rho}^y  \end{bmatrix},
\end{multline}
where ${D_{i,i}^{\rm rows}}=\sum_{j=1}^M
{{K^z_{i,j}}}$ and ${D_{j,j}^{\rm cols}}=\sum_{i=1}^M {{K^z_{i,j}}}$ are
diagonal matrices.

The same minimization problem (Eq.\ \ (\ref{EQargMIN})) can therefore be rewritten as
\begin{equation}\label{mini-D0}
\min_{\hat{\myvec{\rho}}} \widehat{\myvec{\rho}}^\tps(\widehat{\myvec{D}}-\widehat{\myvec{K}}) \widehat{\myvec{\rho}}\;\;\;{\rm s.t.} \;\; \|\widehat{\myvec{\rho}}\|=1,\;\widehat{\myvec{\rho}}^\tps\myvec{1}=0.
\end{equation}
Without the orthogonality constraint, this minimization problem could be solved by finding the minimal
eigenvalue of $(\widehat{\myvec{D}}-\widehat{\myvec{K}})
\widehat{\myvec{\rho}}^T=\bar{\lambda}
\widehat{\myvec{\rho}}^T$.
This eigenproblem has a trivial solution which is the all-ones eigenvector $\myvec{\psi}_0=\myvec{1}$ with $\bar{\lambda}=0$. However, to satisfy the orthogonality constraint, we must use a different eigenvector (which would naturally be orthogonal to $\myvec{\psi}_0$ due to the symmetry of the matrix), leading to the second-smallest (smallest non-zero) eigenvalue $\lambda_1$ with its corresponding eigenvector $\myvec{\psi}_1$.
\end{proof}

\subsection{Spectral decomposition}
\label{SecSpec} In this section, we show how to efficiently compute
the spectral decomposition of $\mymat{\widehat{P}}$ (Eq.\
(\ref{phat})) when only two view exist ($L=2$). As already mentioned above, the matrix $\mymat{\widehat{P}}$ is algebraically
similar (conjugation) to the symmetric matrix
${\mymat{\widehat{P}}_s\defeq{\mymat{\widehat{D}}}^{1/2}
	\mymat{\widehat{P}} {\mymat{\widehat{D}}}^{-1/2}=
	\mymat{\widehat{D}}^{-1/2} \mymat{\widehat{K}}
	{\mymat{\widehat{D}}}^{-1/2} }$. Therefore, both
$\mymat{\widehat{P}}$ and $\mymat{\widehat{P}_s}$ share the same set
of eigenvalues. Due to symmetry of the matrix
$\mymat{\widehat{P}}_s$, it has a set of $2M$ real eigenvalues $\{
\lambda_i \} _{i=0} ^{2M-1} \in \mathbb{R}$ and corresponding real
orthogonal eigenvectors $\{\myvec{\pi}_m \} _{m=0} ^{2M-1} \in
\mathbb{R}^{2M}$, thus, $\mymat{\widehat{P}_s}=\mymat{\Pi \Lambda
	\Pi^T}$. By denoting $\mymat{\Psi}=\hmD^{-1/2}\mPi$
and $\mymat{\Phi}=\hmD^{1/2}\mPi$, we conclude that
the set $\{ \myvec{\psi}_m, \myvec{\phi}_m \} _{m=0} ^{2M-1} \in
\mathbb{R}^{2M}$ are the right and the left eigenvectors of
$\mymat{\widehat{P}}=\mymat{\Psi\Lambda}\mymat{\Phi}^T$,
respectively, satisfying $\mymat{\psi_i}^T
\mymat{\phi_j}=\delta_{i,j}$ (Kronecker's delta function). In the sequel, we use the symmetric
matrix $\mymat{\widehat{P}}_s$ to simplify the analysis.

To avoid the spectral decomposition of a $2M \times 2M$ matrix
$\widehat{\mymat{P}}_s$, the spectral decomposition of
$\mymat{\widehat{P}}_s$ can be computed using the Singular Value
Decomposition (SVD) of the matrix
$\bar{\mymat{K}}^{z}=({\mymat{D}^{\rm rows}})^{-1/2}\mymat{K}^{z}({\mymat{D}^{\rm cols}})^{-1/2}$
of size $M \times M$ where ${D_{i,i}^{\rm rows}}=\sum_{j=1}^M
{{K^z_{i,j}}}$ and ${D_{j,j}^{\rm cols}}=\sum_{i=1}^M {{K^z_{i,j}}}$ are
diagonal matrices. Theorem \ref{T1} enables us to form the
eigenvectors of $\mymat{\widehat{P}}$ as a concatenation of the
singular vectors of $\mymat{K}^z=\mymat{K}^x\cdot \mymat{K}^y$.
\begin{T1} \label{T1} By using the left and right singular vectors of $\mymat{K^{z}=V\Sigma U^T}$, the eigenvectors and the eigenvalues of $\widehat{\myvec{K}}$ are given
	explicitly by
	\begin{equation} \label{EQSVD} {\mymat{\Pi}= \frac{1}{\sqrt{2}}\begin{bmatrix}
		\mymat{V}  & \mymat{V}\\
		\mymat{U}  & \mymat{-U} \\
		\end{bmatrix} ,  {\mymat{\Lambda}= \begin{bmatrix}
			\mymat{\Sigma}  & \mymat{0}_{M \times M}\\
			\mymat{0}_{M \times M}  & \mymat{-\Sigma} \\
			
			\end{bmatrix}. }} \end{equation}
\end{T1}

\begin{proof} Both $\mymat{V}$ and $\mymat{U}$ are orthonormal sets, therefore, ${\myvec{u}_i}^T\myvec{u}_j=\Delta_{i,j}$, and ${\myvec{v}_i}^T\myvec{v}_j=\Delta_{i,j}$, thus,
	the set $\{ \myvec{\pi}_m \}$ is orthonormal. Therefore,
	\mymat{\Pi \Pi^T=I}. By direct substitution of Eq.\
	(\ref{EQSVD}), $\mymat{\Pi \Lambda \Pi}^T$ can be computed explicitly
	as
	\begin{multline}
	\mymat{\Pi \Lambda \Pi}^T= \frac{1}{{2}}\begin{bmatrix}
	\mymat{V}  & \mymat{V}\\
	\mymat{U}  & \mymat{-U} \\
	\end{bmatrix} \begin{bmatrix}  \mymat{\Sigma} & \mymat{0}_{M \times M} \\ {\mymat{0}_{M \times M}} & \mymat{-\Sigma} \end{bmatrix} \begin{bmatrix}
	\mymat{V^T}  & \mymat{U^T}\\
	\mymat{V^T}  & \mymat{-U^T} \\
	\end{bmatrix} \\
	= \frac{1}{{2}} \begin{bmatrix}  \mymat{V\Sigma} & \mymat{-V\Sigma} \\ {\mymat{U\Sigma}} & \mymat{U\Sigma} \end{bmatrix} \begin{bmatrix}
	\mymat{V^T}  & \mymat{U^T}\\
	\mymat{V^T}  & \mymat{-U^T} \\
	\end{bmatrix}
	=
	\frac{1}{{2}} \begin{bmatrix} \mymat{0}_{M \times M} & {2\mymat{K}^{z}} \\ {(2\mymat{K}^{z})^T} & \mymat{0}_{M \times M} \end{bmatrix}={\centering {\mymat{\widehat{K}}}},
	\end{multline}.
\end{proof}
Thus the proposed mapping in Eq.\ (\ref{Map1}) could be computed for $L=2$ using the SVD of $\myvec{K}^z$, Eq.\ (\ref{EQSVD}) and $\mymat{\Psi}=\mymat{\widehat{D}^{-1/2}\Pi}$.

\subsection{Cross-view (CV) diffusion distance}
\label{CVDD} In some physical systems the observed dataset $\mX$ may depend on some underlying parameter, say
$\alpha$. Under this model, multiple snapshots (views) can typically be obtained for
various values of $\alpha$, each denoted $\mX^\alpha$. An example of such a scenario
occurs in hyper-spectral images that change over time. 
Quantifying the amount of change in the datasets due to a change in $\alpha$ in such models is often desired, but if the datasets are high-dimensional, this can be quite a challenging
task. This scenario was recently studied in \cite{Hirn}, where the DM
framework was applied to each fixed value of $\alpha$.
Then, by using the extracted low-dimensional mappings, the Euclidean
distance enables to quantify the extent of changes due to $\alpha$.
However, this approach may be rather unstable, since every small change in the data can
result in different mappings and the mappings are extracted
independently of any mutual influence. 

Our multiview approach, on the other hand, incorporates the mutual relations of data
within each view, as well as the relations between views. This point of view
facilitates a more robust measure of the extent  of variation
between two datasets that  correspond to a small variation in
$\alpha$. To this end, we define a new diffusion distance, which
measures the relation between two views, i.e. between all the data
points obtained for different values of $\alpha$. We measure the distance between all the coupled data points
among the mappings of the snapshots $\myvec{X}^{\alpha_l}$ and $\myvec{X}^{\alpha_m}$ by using the expression
\begin{equation}{ \label{ECVDD} { {{{\cal{D}}_t}^{\rm (CV)}}^2(\myvec{X}^{\alpha_l},\myvec{X}^{\alpha_m})\defeq \sum \limits_{i=1}^{M}\left\|{\myvec{\widehat{\Psi}}_t{\left(\myvec{x}^{\alpha_l}_i\right)}}-{\myvec{\widehat{\Psi}}_t{\left(\myvec{x}^{\alpha_m}_i\right)}}\right\|^2}} .
\end{equation}
Our kernel matrix is a product of the Gaussian kernel matrices in
each view. If these values of the kernel matrices ($\mymat{K}^{x^{\alpha_l}},
\mymat{K}^{x^{\alpha_m}}$) are similar, this corresponds to similarity between
the views' inner geometry. The right- and left-singular vectors of the
matrix $\mymat{K}^{x^{\alpha_l}}\mymat{K}^{x^{\alpha_m}}$ will be similar, thus,
${{{\cal{D}}_t}^{\rm (CV)}}$ will be small.

\begin{T1}
	\label{T5} Using Gaussian kernels with identical $\sigma$ values, the cross manifold distance (defined in Eq.\ (\ref{ECVDD})) is
	invariant to orthonormal transformations between the ambient spaces
	$\mymat{X}^{{\alpha_l}}$ and $\mymat{X}^{{\alpha_m}}$.
\end{T1}
\begin{proof}
	Denote an orthonormal transformation matrix $\mymat{R}: \mymat{X}^{{\alpha_l}} \rightarrow \mymat{X}^{{\alpha_m}}$, w.l.o.g. by $\myvec{x}^{{\alpha_m}}_i=\mymat{R}\myvec{x}^{{\alpha_l}}_i$. Then
	\begin{multline}
	K^{x^{\alpha_m}}_{i,j}=\exp\left\{ -\frac
	{\|\myvec{x}^{\alpha_m}_i-\myvec{x}^{{\alpha_m}}_j\|^2}
	{2\sigma^2_{{{m}}}}\right\}
	=\exp\left\{-\frac
	{\|\mymat{R}\myvec{x}^{\alpha_l}_i-\mymat{R}\myvec{x}^{\alpha_l}_j\|^2}
	{2\sigma^2_m}\right\}\\
	=\exp\left\{-\frac
	{\|\myvec{x}^{\alpha_l}_i-\myvec{x}^{\alpha_l}_j\|^2}
	{2\sigma_{l}^2}\right\}={K^{x^{\alpha_l}}_{i,j}}.
	\end{multline}
	The penultimate transition is due to the orthonomality of $\mymat{R}$ and to the identical $\sigma_l=\sigma_m$ values.
	Therefore, the matrix $\mymat{K}^z=(\mymat{K}^{x^{\alpha_m}})^2$  from Eq.\ (\ref{EQK}) is symmetric and its right- and left-singular vectors are
	equal, i.e.
	$\mymat{U=V}$ in Eq.\ (\ref{EQSVD}). This induces a repetitive form
	in $\mymat{\Psi}=\widehat{\mymat{D}}^{{-1}/{2}}\mymat{\Pi}\rightarrow \psi_l[i]=\psi_l[M+i],\text{ } 1 \leq i,l \leq M-1 \rightarrow {\Psi_t{(\myvec{x}^{\alpha_l}_i)}}={\Psi_t{(\myvec{x}^{\alpha_m}_i)}}$,
	thus, ${{{\cal{D}}_t}^{(CM)}}^2(\myvec{X}^{\alpha_l},\myvec{X}^{\alpha_m})=0$.
	
\end{proof}

\subsection{Spectral decay of $\mymat{\widehat{K}}$}
\label{SECDEC}
The power of kernel based methods for dimensionality reduction stems
from the spectral decay of the kernel matrix' eigenvalues. In this
subsection we study the relation between the spectral decay of the
Kernel Product (Eq.\ (\ref{EquationPDH})) and our multi-view kernel
(Eq.\ (\ref{phat})). 

\begin{T1}
	\label{T6}
	All $2M$ eigenvalues of $\mymat{\widehat{P}}$ (Eq.\ (\ref{phat})) are real-valued
	and bounded, $|\lambda_i| \leq 1$, $i=0,...,2M-1$.
	
\end{T1}

\begin{proof}
    \footnote[1]{A similar proof is given in https://sites.google.com/site/yoelshkolnisky/teaching} 
	As shown in section \ref{SecSpec}, $\mymat{\widehat{P}}$ is
	algebraically similar to a symmetric matrix, thus its eigenvalues
	are guaranteed to be real-valued. Denote by $\lambda$ and $\myvec{\psi}$
	an eigenvalue and an eigenvector, resp., such that
	$\lambda\myvec{\psi}=\mymat{\widehat{P}}\myvec{\psi}$.\\ 
	Define 	$i_0\defeq\arg\max_i |\psi[i]|$ (the index of the largest element of $\myvec{\psi}$).
	The	maximal value $\psi[i_0]$ can be computed using
	$\mymat{{\widehat{P}}}$ from Eq.\ (\ref{phat}),
	\begin{equation}
	    \lambda
	\psi[i_0]={\sum\limits_{j=0}^{2M-1}{{\widehat{P}}_{i_0j}\psi[j]}}
	\Rightarrow
	|\lambda|=\left|{\sum\limits_{j=0}^{2M-1}{{\widehat{P}}_{i_0j}\frac{\psi[j]}{\psi[i_0]}}}\right|
	\leq{\sum\limits_{j=0}^{2M-1}{{\widehat{P}}_{i_0j}\frac{|\psi[j]|}{|\psi[i_0]|}}}
	\leq{\sum\limits_{j=1}^{2M}{{\widehat{P}}_{i_0j}}}=1.
	\end{equation}
	The first inequality is due to the triangle inequality and the
	second equality is due to the definition of $i_0$.
	
\end{proof}
Although, according to this Theorem, the eigenvalues are all real-valued and bounded,
their mere boundedness is generally insufficient for dimensionality reduction.
Dimensionality reduction is meaningful only in the presence of a significant
spectral decay.

\begin{D1}
	\label{D1}
	Let $\cal{M}$ be a manifold. The intrinsic dimension $d$ of the manifold  is a positive integer determined by how many independent ``coordinates'' are needed to describe $\cal{M}$. Using a parametrization to describe a manifold, the dimension of $\cal{M}$ is the smallest integer $d$ such that a smooth map $\myvec{f}(\myvec{\xi})=\cal{M}$ describes the manifold, where $\myvec{\xi} \in {\mathbb{R}}^d$.
\end{D1}
Our framework employs a Gaussian kernel, the spectral decay of
which was studied in \cite{Lafon}. We use Lemma \ref{L1}
(which is based on Weyl's asymptotic law and appears in \cite{LafonPHD}) to evaluate the spectral decay of our kernel.

\begin{L1}
	\label{L1} Assume that the data is sampled from a manifold with
	intrinsic dimension $d \ll M$. Let $\mymat{K^{\circ}}\in\mathbb{R}^{M\times M}$ (a kernel constructed based on the concatenation of views) denote the kernel with an exponential decay as a
	function of the Euclidean distance. For $\delta > 0$, the number of
	eigenvalues of $\mymat{K}^{\circ}$  above $\delta$ is proportional
	to $(\log(\frac{1}{\delta}))^d$.
\end{L1}

Assume that the eigenvalues of $\mymat{K}^\circ$ are arranged in descending order of their absolute values, $|\lambda_0|\ge|\lambda_1|\ge\cdots\ge|\lambda_{M-1}|$, set $\delta\in(0,1)$ and define 
$r_{\delta}\defeq\max\{\ell \in [1,M]:|\lambda_{\ell-1}| > \delta \}$ (denoting the number of eigenvalues
of $\mymat{K^{\circ}}$ above $\delta$).
Recall now that $\mymat{K}^{\circ}=\mymat{K}^x\circ\mymat{K}^y$ corresponds to a single DM view (formed of the concatenation of two views)
as addressed in \cite{Lafon}. Theorem \ref{SpectralDecay} relates the
spectral decay of our proposed kernel $\widehat{\mymat{P}}$ (Eq.\ (\ref{phat})) to the decay of the Kernel Product-based DM ($\mymat{P}^{\circ})$.

\begin{L1}
	\label{L2}
	Let $\mymat{A,B}\in \mathbb{R}^{M\times M}$ be any two positive semi-definite (PSD) matrices. Then for any $1 \leq k \leq M-1$
	\begin{equation} 
	\label{E17} 
	\prod\limits_{{\ell}=k}^{M-1}{\lambda_{\ell}(\mymat{A\cdot B})}\leq \prod\limits_{{\ell}=k}^{M-1} {{\lambda_{\ell}(\mymat{A\circ B})}}  
	\end{equation}
    where $\lambda_\ell(\cdot)$ denotes the $\ell$-th eigenvalue (in descending order) of the enclosed matrix.	
\end{L1}
This inequality is proved in \cite{Ando} and \cite{Visick}.

\begin{T1}
	\label{SpectralDecay}
The product of the last $M-1-r_{\delta}$ eigenvalues of
	$\mymat{K}^z$ is smaller or equal to $\delta^{M-1-r_{\delta}}$.
	Formally,
	${\prod\limits_{{\ell}=r_{\delta}}^{M-1}{\lambda_{\ell}(\mymat{K}^x\cdot \mymat{K}^y)}\leq
		\delta^{M-1-r_{\delta}}}$.
\end{T1}

\begin{proof}
	Substitute the PSD matrices $\mymat{A}=\mymat{K}^x$ and
	$\mymat{B}=\mymat{K}^y$ in Lemma \ref{L2} and choose
	${\ell}=r_{\delta}$ in Eq.\ \ref{E17} to obtain
	\begin{equation}
	{\prod\limits_{{\ell}=r_{\delta}}^{M-1}{\lambda_{\ell}(\mymat{K}^x \cdot 
			\mymat{K}^y)}\leq \prod\limits_{{\ell}=r_{\delta}}^{M-1}
		{{\lambda_{\ell}(\mymat{K}^{\circ})}}\leq \delta^{M-1-r_{\delta}}}.
	\end{equation}
\end{proof}

Relying on the spectral decay of the kernel matrix, we can approximate Eq.\
(\ref{DIV3}) by neglecting all eigenvalues smaller than $\delta$. Thus, we can compute a low
dimensional mapping such that
\begin{equation}{ \widehat{\myvec{\Psi}}^r_t{(\myvec{x}_i)}:   \myvec{x}_i  \longmapsto \begin{bmatrix}
	{ \lambda_1^{t}\psi_1[i]} ,
	{ \lambda_2^{t}\psi_2[i]} ,
	{ \lambda_3^{t}\psi_3[i]} ,
	{.}
	{.}
	{.}   ,
	
	{\lambda_{r-1}^{t}\psi_{r-1}[i]}\\
	
	\end{bmatrix}^T \in{\mathbb{R}^{r-1}} }.
\end{equation}
The following Lemma introduces an error bound for using this low-dimensional mapping:
\begin{L1}
\label{boundDr}
	The truncated diffusion distance up to coordinate $r$ defined as
	\begin{equation}
	[{\cal{D}}^r_t(\myvec{x}_i,\myvec{x}_j)]^2\defeq\left\|\widehat{\myvec{\Psi}}^r_t{(\myvec{x}_i)}-\widehat{\myvec{\Psi}}^r_t{(\myvec{x}_j)}\right\|^2=\sum^r_{s=1}\lambda_s^{2t}(\psi_s[i]-\psi_s[j])^2,		\end{equation} 
	is bounded by the inner view diffusion distance (defined in Eq.\ (\ref{DIV1}))
	\[ 2\cdot\left[\sum^{M-1}_{s=1}\lambda_s^{2t}(\psi_s[i]-\psi_s[j])^2- \delta^{2t}\cdot\left(\frac{1-\Delta_{i,j}}{\widehat{D}^{\rm min}_{i,j}}\right)\right] 
	\leq \left[{\cal{D}}^r_t(\myvec{x}_i,\myvec{x}_j)\right]^2\leq 2\cdot\sum^{M-1}_{s=1}\lambda_s^{2t}(\psi_s[i]-\psi_s[j])^2 , \] 
	where $\widehat{D}^{\rm min}_{i,j}$ is the minimal value of $\widehat{D}_{i,i}\text{ and }\widehat{D}_{j,j}$, $\Delta_{i,j}$ is the Kronecker delta function and $\delta$ is the accuracy threshold.
\end{L1}
\begin{proof}
	For the last inequality, clearly 	\[[{\cal{D}}^r_t(\myvec{x}_i,\myvec{x}_j)]^2\leq [{\cal{D}}^{2M}_t(\myvec{x}_i,\myvec{x}_j)]^2=2\cdot\sum^{M-1}_{s=1}\lambda_s^{2t}(\psi_s[i]-\psi_s[j])^2,\] the equality is a result of the repetitive form of $\mymat{\Psi}$ and $\mymat{\Lambda}$, which were defined in Theorem \ref{T1} for $L=2$. Note that $s=0$ was excluded from the sum as $\myvec{\psi}_0=\myvec{1}$ is constant.
	For the first inequality, using $\mymat{\Psi}=\widehat{\mymat{D}}^{-1/2}\mymat{\Pi}$ where $\mymat{\Pi}$ is an orthonormal basis defined in Eq.\ (\ref{EQSVD}), we get
	\[\mymat{\Psi\Psi}^T=\widehat{\mymat{D}}^{-1/2}\mymat{\Pi}\mymat{\Pi}^T\widehat{\mymat{D}}^{-1/2}=\widehat{\mymat{D}}^{-1}, \] 
	which means that 
	\begin{equation}
	\sum^{2M-1}_{s=0}(\psi_s[i]-\psi_s[j])^2=\frac{1}{\widehat{D}_{i,i}}+\frac{1}{\widehat{D}_{j,j}}-\frac{2\Delta_{i,j}}{\widehat{D}_{i,i}}.
	\end{equation} 
	Using the definition of the truncated diffusion distance we have
	\begin{multline}
    \left[{\cal{D}}^r_t(\myvec{x}_i,\myvec{x}_j)\right]^2=\sum^{2M-1}_{s=0}\lambda_s^{2t}(\psi_s[i]-\psi_s[j])^2-\sum^{2M-1}_{s=r+1}\lambda_s^{2t}(\myvec{\psi}_s[i]-\myvec{\psi}_s[j])^2\\
    \geq 2\sum^{M-1}_{s=1}\lambda_s^{2t}(\psi_s[i]-\psi_s[j])^2-\delta^{2t}\sum^{2M-1}_{s=0}(\psi_s[i]-\psi_s[j])^2\\
    \geq 2\left[\sum^{2M-1}_{s=1}\lambda_s^{2t}(\psi_s[i]-\psi_s[j])^2-\delta^{2t}\cdot\left(\frac{1-\Delta_{i,j}}{\widehat{D}^{\rm min}_{i,j}}\right)\right].
	\end{multline}
	 In the same way, a similar bound for the truncated diffusion distance between $\myvec{y}_i$ and $\myvec{y}_j$ can be derived.
	
\end{proof}
The dimension $r$ is typically determined by looking at the decay rate of the eigenvalues and on choosing a threshold level $\delta$. In this subsection, we demonstrated how the dimension $r$ and the threshold $\delta$ are related to the approximation of the diffusion distance. By removing all coordinates with eigenvalues smaller than $\delta$, the error of the diffusion distance is bounded according to Lemma \ref{boundDr}. The decaying property of the eigenvalues could be quantified by the numerical rank of the Gaussian kernel matrix. Next, we present some existing results that shed some light onto the practicality of the truncation of the diffusion coordinates up to dimension $r$. In practice, there usually is no single optimal value for $r$, and the number of necessary coordinates depends on the application.
\subsubsection{Intrinsic Dimension and Numerical Rank}
 In \cite{bermanis2013multiscale} Bermanis {\em et al.} prove that the numerical rank of a Gaussian kernel matrix is typically independent of its size. Their results state that the numerical rank is proportional to the volume of the data. By using boxes with side-length of $\epsilon^{D/2}= \sigma^D$ ($D$ is the dimension of the data), it is shown that the numerical rank is bounded from above by the minimal number of such cubes required to cover the data. As the number of eigenvalues $|\lambda_\ell | > \delta$ is proportional to the numerical rank, this means that for a prescribed $\delta$, the number of coordinates $r_{\delta}$ required is independent of the number of samples $N$.
 
Practically speaking, this result is useful if the numerical rank is small. Furthermore, as there are several methods for estimating the numerical rank, using the numerical rank to choose $r_{\delta}$ is not very expensive. Another practical method, which we found useful for setting $r$ without directly choosing $\delta$, is based on estimating the intrinsic dimension of the data $d$. Various methods have been proposed for estimating the intrinsic dimension of high-dimensional data. Here, we provide a brief description of a few. Popular methods, such as \cite{fukunaga1971algorithm,verveer1995evaluation} apply local PCA to the data, and estimate the intrinsic dimension as the median of the number of principal components corresponding to singular values larger than some threshold. Based on our experience, we found that a method called Dimensionality from Angle and Norm Concentration (DANCo) \cite{danco} provides a robust estimation of the intrinsic dimension $d$. DANCo generates artificial data of dimension $d$ and attempts to minimize the Kullback–Leibler divergence between the estimated probability distribution functions (pdf-s) of the observed data and the artificially-generated samples.

\subsection{Out-of-sample extension}

\label{SecGH}
To extend the diffusion coordinates to new data points without re-applying a large-scale eigendecomposition \cite{keller2010audio}, the Nystr\"{o}m extension \cite{nystrom} is widely used. Here we formulate the extension method for a multi-view scenario. Given the data sets $\myvec{X}$ and $\myvec{Y}$ and new points $\tilde{\myvec{x}} \notin \myvec{X} \text{ and }\tilde{\myvec{y}}\notin \myvec{Y}$, we want to extend the multi-view diffusion mapping to $\tilde{\myvec{x}}\text{ and }\tilde{\myvec{y}}$ without re-applying the batch procedure. First, we describe the explicit form for the eigenvalue problem for two views $\myvec{X}$ and $\myvec{Y}$. The eigenvector $\myvec{\psi}_k$ and its associated eigenvalue $\lambda_k$ satisfiy $\lambda_k\myvec{\psi}_k= \widehat{\myvec{P}}\myvec{\psi}_k$. Substituting the definition of $\widehat{\myvec{P}}$ from Eqs.\ (\ref{EQK}) and (\ref{phat}) we get that

\[	\lambda_k\myvec{\psi}_k=\widehat{\myvec{D}}^{-1} \begin{bmatrix}  \mymat{0}_{M \times M} & {\mymat{K}^x\cdot\mymat{K}^y} \\ {\mymat{K}^y\cdot \mymat{K}^x} & \mymat{0}_{M \times M} \end{bmatrix}\cdot \myvec{\psi}_k\in\mathbb{R}^{2M},\] 
due to the block form of the matrix $\widehat{\myvec{P}}$ 
\[\lambda_k\myvec{\psi}^x_k[i]=\sum_{j}{{\widehat{p}}}(\myvec{x}_i,\myvec{y}_j)\myvec{\psi}^y_k[j]\in\mathbb{R}^M,  \]
\[\lambda_k\myvec{\psi}^y_k[i]=\sum_{j}{{\widehat{p}}}(\myvec{y}_i,\myvec{x}_j)\myvec{\psi}^x_k[j]\in\mathbb{R}^M, \]
where $\myvec{\psi}^x_k[i] \defeq  \myvec{\psi}_k[i], i=1,...,M$ and  $\myvec{\psi}^y_k[i] \defeq  \myvec{\psi}_k[i+M], i=1,...,M$. The transition matrices are \[{{\widehat{p}}}(\myvec{x}_i,\myvec{y}_j)=\frac{\sum_{s}{K^x_{i,s}}K^y_{s,j}}{{\widehat{D}_{i,i}^{\rm rows}}}\text{, and }{{\widehat{p}}}(\myvec{y}_i,\myvec{x}_j)=\frac{\sum_{s}{K^y_{i,s}}K^x_{s,j}}{{\widehat{D}_{i,i}^{\rm cols}}}.\]

The Nystr\"{o}m extension is is similarly obtained by computing a weighted sum of the original eigenvectors. The weights are computed by applying the kernel $\cal{K}$ to the extended data points, followed by row-normalization. For the proposed mapping, the extension is defined by
\begin{equation}\label{ExtensionX} \hat{\psi}_k(\myvec{\tilde{x}})=\frac{1}{\lambda_k}\underset{{j}}{\sum}{\hat{p}({\myvec{\tilde{x}}},\myvec{y}_j)\psi^y_k [j] } \end{equation}
\begin{equation}\label{ExtensionY} \hat{\psi}_k(\myvec{\tilde{y}})=\frac{1}{\lambda_k}\underset{{j}}{\sum}{\hat{P}({\myvec{\tilde{y}}},\myvec{x}_j)\psi^x_k [j]}\end{equation}
where the ``missing" probabilities $\hat{p}(\tilde{\ux},\uy_j)$ and $\hat{p}(\tilde{\uy},\ux_j)$ are approximated using the original kernel function as
\begin{equation}
    \hat{p}(\tilde{\ux},\uy_j)={\sum_{s}{\exp \bigg\{ -\frac{||\tilde{\myvec{x}}-\myvec{x}_s||^2}{2\sigma^2_x}\bigg\}}  \exp\bigg\{-\frac{||\myvec{y}_s-\myvec{y}_j||^2}{2\sigma^2_y}\bigg\}  } \frac{ 1 }{\widehat{D}_{j+M,j+M} } 
\end{equation}
\begin{equation}
    \hat{p}(\tilde{\uy},\ux_j)={\sum_{s}{\exp \bigg\{ -\frac{||\tilde{\myvec{y}}-\myvec{y}_s||^2}{2\sigma^2_y}\bigg\}} }     \exp\bigg\{-\frac{||\myvec{x}_s-\myvec{x}_j||^2}{2\sigma^2_x}\bigg\}   \frac{ 1 }{\widehat{D}_{j+M,j+M} }  
\end{equation}
and where $\psi_k^x[j]=\psi_k[j]$ and $\psi_k^y[j]=\psi_k[m+j]$. 

The new mapping vector for the new data point is then given by                                          
\begin{equation}{ \label{PsiHat} {\hat{ {{{\Psi}}}}}({\myvec{\tilde{x}})= \begin{bmatrix}
		{ \lambda_1\hat{\psi}_1({\myvec{\tilde{x}}})},
		{ \lambda_2\hat{\psi}_2({\myvec{\tilde{x}}})},
		{ \lambda_3\hat{\psi}_3({\myvec{\tilde{x}}})},
		{.}
		{.}
		{.},
		{\lambda_{M-1}\hat{\psi}_{M-1}({\myvec{\tilde{x}}})}
		
		\end{bmatrix} } \in {\mathbb{R}^{M-1}}}.
\end{equation}    \
The new coordinates in the diffusion space are approximated and the new data points $\myvec{\tilde{x}},\myvec{\tilde{y}}$ have no effect on the original map's structure.

\subsection{Infinitesimal generator}
In the subsequent analysis we only consider kernel functions operating on the scaled difference between their two arguments, namely kernel functions which can be written as 
\begin{equation}
\mymat{\cal{K}}(\ux_i,\ux_j)=K\left(\frac{\ux_i-\ux_j}{\sqrt{\epsilon}}\right),
\end{equation}
where $\epsilon$ is a positive constant, and $K(\uz):\Rset^D\mapsto\Rset$ is a non-negative symmetric function, namely $R(\uz)\ge 0$ and $R(\uz)=R(-\uz)$ $\forall \uz\in\Rset^D$. For example, the Gaussian kernel (see subsection \ref{SecDiff}) $\nK(\vx_i,\vx_j)= \exp\left\{-\frac{\|\vx_i-\vx_j\|^2}{2\sigma_x^2}\right\}$ satisfies this property with $\epsilon=\sigma^2$.

\label{inf_gen} A family of differently normalized diffusion operators was introduced in
\cite{Lafon}. If
appropriate limits are taken such that $M\rightarrow \infty$,
$\epsilon\rightarrow 0$, then from
\cite{Lafon} it follows that the DM kernel operator will converge to one of the
following differential operators: 1.\ Normalized graph Laplacian; 2.\
Laplace-Beltrami diffusion; or 3.\ Heat kernel equation. These are
proved in \cite{Lafon}. The operators are all special cases of the
diffusion equation. This convergence provides not only a physical justification
for the DM framework, but allows in some cases to distinguish
between the geometry and the density of the data points. In this
subsection we study the asymptotic properties of the proposed kernel
$\widehat{\myvec{K}}$ (Eq.\ (\ref{EQKMAT})), limiting the discussion to only two views, i.e. $L=2$.

We are interested in understanding the properties of the eigenfunctions of the proposed multi-view kernel  $\mymat{{\widehat{P}}}$ (Eqs.\ (\ref{EQKMAT}), (\ref{phat})) for two views. We assume that there is some unknown mapping $\myvec{\beta}: \mathbb{R}^d \rightarrow  \mathbb{R}^d$ from view $\myvec{X}$ to view $\myvec{Y}$ that satisfies $\myvec{y}_i=\myvec{\beta}(\myvec{x}_i),i=1,...,M$. Each view-specific kernel applies the same function, namely ${K}^x(\uz)={K}^y(\uz)=K(\uz)$, and $K(\uz)$ is normalized such that $\int_{\mathbb{R}^d} K(\myvec{z})\dz=1$. Note that the Gaussian kernel can be propely normalized to satisfy this requirement. The analysis relates to data points $\{ \myvec{x}_1,...,\myvec{x}_M \} \in \mathbb{R}^D$ sampled from a uniform distribution over a bounded domain in ${\mathbb{R}^d}$. The image of the function $\myvec{\beta}$ is a bounded domain in $\mathbb{R}^d$ with distribution ${\alpha}(\myvec{\myvec{z}})$.

\begin{T1}
	\label{T9}
	The infinitesimal generator induced by the proposed kernel matrix $\widehat{\mymat{K}}$ (Eq.\ (\ref{EQKMAT})) after row-normalization, denoted in here as $\widehat{\mymat{P}}$, converges when
	$M\rightarrow \infty, \epsilon \rightarrow 0$ (with $\epsilon=\sigma_x^2=\sigma_y^2$) to a ``cross domain Laplacian operator''. The functions $f(\myvec{x})$ and $g(\myvec{y})$ converge to eigenfunctions of $\mymat{{\widehat{P}}}$. These functions
	are the solutions of the following diffusion-like equations:
	\begin{align} (\widehat{\myvec{P}}f)(\myvec{x}_i)&=g(\myvec{\beta(x_i)})+ {{\epsilon}} \triangle \gamma(\myvec{\beta(\myvec{x}_i)})/\alpha(\myvec{\beta(x_i)}) +\mathcal{O}(\epsilon^{3/2}), \\
	(\widehat{\myvec{P}}g)(\myvec{y}_i)&= f(\myvec{\beta^{-1}}(\myvec{y}_i))+{{\epsilon}} \triangle \eta(\myvec{\beta^{-1}(\myvec{y}_i)})/\alpha(\myvec{\beta(y_i)}+\mathcal{O}(\epsilon^{3/2}),  \end{align} where the functions $\gamma, \eta$ are defined as $\gamma(\myvec{z})\defeq g(\myvec{z})\alpha(\myvec{z}),\eta(\myvec{z})\defeq f(\myvec{z})\alpha(\myvec{z})$.
\end{T1}
The proof of theorem \ref{T9} is deferred to the Appendix. Although the interpretation of the result is hardly intuitive, it provides evidence that the mapping $\myvec{f}$ and $\myvec{g}$ are indeed coupled and are related to the second derivative on the manifold. However, the distribution of the points $\alpha$ has an impact on the mapping, which we hope to reduce in future research.

\subsection{The convergence rate}
In Theorem \ref{T9}, we assume the number of data points $M\rightarrow \infty$ while the scale parameter $\epsilon\rightarrow 0$. In practice we cannot expect to have an infinite number of data points. It was shown in \cite{Lafon,Belkin} (and elsewehere) that a single-view graph Laplacian converges to the laplacian operator on a manifold. It is demonstrated in \cite{Hein,SingerConvergence} that the variance of the error for such an operator decreases as $M\rightarrow \infty$, but increases as $\epsilon\rightarrow 0$. The study in \cite{Hein} proves that for a uniform distribution of data points, the variance of the error is bounded by $\mathcal{O}(\frac{1}{M^{1/2}\epsilon^{1+d/4}},\epsilon^{1/2})$. This bound was improved in \cite{SingerConvergence} by an asymptotic factor of $\sqrt{\epsilon}$ based on the correlation between $\myvec{D}^{-1}$ and $\myvec{K}$.

We now turn our attention to the variance of the multi-view kernel for a finite number of points. Given $\myvec{x}_1,....,\myvec{x}_M$ independent uniformly distributed data points sampled from a bounded domain in $\mathbb{R}^d$, define the multi-view Parzen Window density estimator by 
\begin{equation} \label{eq:Parzen}
\dot{\myvec{K}}_{M,\epsilon}(\myvec{x}) \defeq \frac{1}{M^2} \sum_{\ell=1}^M \sum^M_{j=1} \frac{1}{\epsilon^d} K\left(\frac{\myvec{x}-\myvec{x}_{\ell}}{\sqrt{\epsilon}}\right) K\left(\frac{\myvec{y}_{\ell}-\myvec{y}_{j}}{\sqrt{\epsilon}}\right).
\end{equation}
We are interested in finding a bound for the variance of $\dot{\myvec{K}}_{M,\epsilon}(\myvec{x})$ for a finite number of data points:
\begin{multline}
\var(\dot{\myvec{K}}_{M,\epsilon}(\myvec{x}))=\frac{1}{{M}^4\epsilon^{2d}} \cdot M \cdot \var\left( \sum^M_{\ell} K\left(\frac{\myvec{x}-\myvec{x}_{\ell}}{\sqrt{\epsilon}}\right) K\left(\frac{\myvec{y}_{\ell}-\myvec{y}_{j}}{\sqrt{\epsilon}}\right)  \right)\\ 
\leq\frac{1}{{M}^4\epsilon^{2d}} \cdot M^3 \cdot \var \left( K^{x}_{\epsilon} K^{y}_{\epsilon}\right)  \leq\frac{1}{{M}\epsilon^{2d}} \big[ \var\big(K^x_{\epsilon}\big) \cdot ||K^y_{\epsilon} \left|_{\infty} +  \var\big(K^y_{\epsilon}\big) \cdot ||K^x_{\epsilon} \right|_{\infty}\big]\\ 
\leq\frac{1}{{M}\epsilon^{2d}} \cdot[{\epsilon^{d/2}}  \cdot m_1 \cdot 1 + \epsilon^{d/2}\cdot m_2\cdot \alpha(\myvec{x})\big] \leq \frac{ m_1  +  m_2\cdot \alpha(\myvec{x})}{M\cdot \epsilon^{1.5d}},
\end{multline}
where the constants $m_1$ and $m_2$ are functions of the chosen kernels and of the pdf $\alpha(\cdot)$ of the points $\myvec{y}_i,i=1,...,M$. This bound helps to choose an optimal value for the scaling factor $\epsilon$ given the number of data points $M$ and the intrinsic dimension $d$.

\subsection{Generalized multi-view kernel}

One can consider a more general multi-view kernel which does not preclude a transition within views $\myvec{X}$ and $\myvec{Y}$ in each time step. Such a kernel will take the form
\begin{equation} \label{EQKMATextended}
\mymat{\widehat{\widehat{K}}}=  \begin{bmatrix}  \eta\cdot(\mymat{K}^x)^2 & 
(1-\eta)\cdot\mymat{K}^x\mymat{K}^y \\ (1-\eta)\cdot\mymat{K}^y\mymat{K}^x& \eta\cdot(\mymat{K}^y)^2 \end{bmatrix},\end{equation} where the parameter $\eta\in [0,1] $ prescribes the (implied) probability of within-view transitions. This kernel is normalized using the sum of rows diagonal matrix $\widehat{\widehat{\myvec{D}}}$, such that $\myvec{\widehat{\widehat{P}}}=\widehat{\widehat{\myvec{D}}}^{-1} \mymat{\widehat{\widehat{K}}}$.
For large values of $\eta$, the kernel favors the within-view transitions, thereby sharing characteristics with the single-view diffusion process. For small values of $\eta$, the kernel tends to behave like our multi-view kernel $\mymat{\widehat{K}}$ (Eq.\ (\ref{EQKMAT})).

\subsection{Additive noise}
In this sub section we explore the effect of noise by following the analysis presented in \cite{Boots}: Assuming that the noise is additive, we evaluate how the proposed method preserves the statistics of the latent model parameters $\myvec{\Theta}\in\mathbb{R}^{d\times M}$. We begin by considering the case of a single-view linear model, then consider a multi-view (linear model) and finally use the kernel trick to relax the linearity assumption. 

The single-view linear model is defined by $\myvec{x}_i= \myvec{A} \myvec{\theta}_i + \myvec{\xi}_i,i=1,...,M$, where the vectors $\myvec{\theta}_i \in \mathbb{R}^d,i=1,...,M$ (columns of $\mymat{\Theta}$) describe latent parameters, modeled as i.i.d., zero-mean random vectors with a finite covariance matrix $\myvec{\Sigma}_{\myvec{\theta}}$, and $\myvec{\xi}_i\in \mathbb{R}^d$ are i.i.d., zero-mean random ``noise" vectors with a finite covariance matrix $\myvec{\Sigma}_{\xi}$.  The observations are $\myvec{x}_i \in \mathbb{R}^D$ and the matrix $\myvec{A} \in \mathbb{R}^{D\times d}$ has full column rank. 

A standard approach for reducing the dimension is to apply PCA. However, in the presence of noise, even if the covariance of $\myvec{\Theta}$ is full rank PCA will provide a biased representation of $\myvec{\Theta}$. In applying PCA to the set $\myvec{X}$, one first computes the sample covariance $\frac{1}{M}\myvec{X}\myvec{X}^T$, which at the limit $M \rightarrow \infty$ converges to $\myvec{A}\myvec{\Sigma}_{\myvec{\theta}}\myvec{A}^T + \myvec{\Sigma}_ {\myvec{\xi}}$. This means that using the top principal components of $\myvec{X} $ embeds the data into a biased representation of $\myvec{\Theta}$. Nonetheless, by introducing a second set of measurements $\myvec{y}_i= \myvec{B} \myvec{\theta}_i + \myvec{\eta}_i,i=1,...,M$, where $\mymat{B}\in\mathbb{R}^{D\times d}$ is another full-rank matrix and $\myvec{\eta}_i$ are i.i.d.\ zero-mean noise vectors which are all uncorrelated with the respective $\myvec{\xi}_i$ (namely, $E\left[\myvec{\xi}_i \cdot \myvec{\eta}_i^\tps\right]=\mymat{0},i=1,...,M$), one can remove the bias term. Essentially, the corresponding additional measurements, along with the noise decorrelation assumption, provide the information required for retrieving an unbiased representation of $\myvec{\Theta}$. Removing the bias term is possible by applying CCA \cite{CCA2} to $\myvec{X}$ and $\myvec{Y}$. CCA not only removes the bias, but enables to uncover a reduced representation that preserves the statistics of $\myvec{\theta}$. This is because the sample cross covariance of $\myvec{X}$ and $\myvec{Y}$ is an unbiased estimate of $\myvec{A} \myvec{\Sigma}_{\myvec{\theta}} \myvec{B}^T$. The top $d$ right- and left-singular vectors of $\frac{1}{M} \myvec{XY}^T$ are used to embed the data into a $d$ dimensional space, such that the statistics of $\myvec{\Theta}$ are preserved. By denoting the top $d$ singular vectors as $( \hat{\myvec{U}}, \myvec{S}, \hat{\myvec{V}}^T )=\text{SVD}_d (\frac{1}{M} \myvec{XY}^T)$, the reduced representations are defined as $\myvec{\hat{U}}^T\myvec{X}$ and $\myvec{\hat{V}}^T\myvec{Y}$, respectively.

Although CCA is an effective, powerful tool in this context, it is based on a linear model, and thus limited to linear transformations of the data. A widely used solution to capture non-linear relations in the data are Kernel matrices \cite{weinberger2004learning,shiokawa2018application,zhang2018large}. Using a kernel matrix in the ambient space is a natural way to capture the sample-covariance in some unknown high-dimensional feature space \cite{scholkopf2001kernel}. Therefore, using a kernel generalizes the results provided by CCA to the nonlinear case. We now demonstrate how under mild assumptions the proposed kernel $\myvec{K}^z=\myvec{K}^x\myvec{K}^y$ mitigates the bias effect of the additive uncorrelated noise.

Assume that $\myvec{x}_i,\myvec{y}_i,i=1,...,M$ are noisy observations of the same low-dimensional latent variable $\myvec{\theta}_i$, such that $\myvec{x}_i=\myvec{r}(\myvec{\theta}_i) +\myvec{\xi}_i$ and $\myvec{y}_i=\myvec{h}(\myvec{\theta}_i)
+\myvec{\eta}_i$. The functions $\myvec{r,h}: \mathbb{R}^d  \rightarrow \mathbb{R}^D$. The affinity values of $\myvec{K}^x,\myvec{K}^y$ are the sample covariance in two high-dimensional feature spaces $\myvec{\Gamma,\Delta}$. The inaccesible feature maps for both views are defined by $\myvec{\gamma}(\myvec{x}_i)$ and $\myvec{\delta}(\myvec{y}_i)$. This means that by computing the SVD of $\myvec{K}^z=\myvec{K}^x \myvec{K}^y$,  we would essentially be applying CCA in the inaccessible features spaces $\myvec{\Gamma,\Delta}$. This implies that if the following conditions hold
\begin{enumerate}
	\item The functions $\myvec{r,h}\in \mathbb{C}^{\infty}$ are invertible.
	\item The feature representations $\myvec{\Gamma(\myvec{X})\text{ and }\Delta(\myvec{Y})}$ are centered.
	\item The noise terms are uncorrelated: $E \left[\myvec{\xi}_i \cdot \myvec{\eta}_i^\tps\right]=\mymat{0},i=1,...,M$.
	\item $E [\myvec{\gamma}(\myvec{x}_i) | \myvec{\theta}_i]=a_0 \myvec{\theta}_i +a_1$, with some real valued constants $a_0$ and $a_1$.	
	\item $E [\myvec{\delta}(\myvec{y}_i) | \myvec{\theta}_i]=b_0 \myvec{\theta}_i +b_1$, with some real valued constants $b_0$ and $b_1$. 
\end{enumerate}
then, applying SVD to $\myvec{K}^z$ enables to cancel out the view-specific noise terms. Condition (1) implies that the latent parameters $\myvec{\Theta}$ lie on a noisy manifold in both observed spaces. Based on condition (2) and on the choice of kernels, the empirical covariance matrices in the feature space are $\myvec{K}^x=\frac{1}{M}\myvec{\Gamma \Gamma}^T $ and $\myvec{K}^y=\frac{1}{M}\myvec{\Delta \Delta}^T$. We remind that $\myvec{\Gamma}$ and $\myvec{\Delta}$ are inaccessible and the kernels are computed based on $\myvec{X}\text{ and } \myvec{Y}$. As in the linear case, using $\myvec{K}^x$ or $\myvec{K}^y$ alone is insufficient for obtaining an unbiased estimate of the representation of $\myvec{\Theta}$.

We now turn our attention to demonstrate that the matrix $\myvec{K}^z=\myvec{K}^x\myvec{K}^y=\frac{1}{M^2}\myvec{\Gamma \Gamma^T}\myvec{\Delta \Delta^T}$ enables to remove the uncorrelated additive noise. A right eigenvector of $\myvec{K}^z$, denoted by $\myvec{\tau}_i$, satisfies
\begin{equation}\label{eq:tau}
\myvec{K}^z \myvec{\tau}_i =\frac{1}{M^2}\myvec{\Gamma \Gamma^T}\myvec{\Delta \Delta^T} \myvec{\tau}_i=\lambda_i \myvec{\tau}_i,\end{equation} by multiplying Eq.\ \ref{eq:tau} on the left by $\myvec{\Delta}^T$ we get  
\begin{equation}\label{eq:tau2} \frac{1}{M^2}\myvec{\Delta}^T \myvec{\Gamma \Gamma^T}\myvec{\Delta \Delta^T} \myvec{\tau}_i=\lambda_i \myvec{\Delta}^T \myvec{\tau}_i. \end{equation} Substituting $\myvec{u}_i=\myvec{\Delta}^T\myvec{\tau}_i$, $\bar{\myvec{\Sigma}}_{\myvec{\gamma}\myvec{\delta}}=\frac{1}{M}\myvec{\Gamma^T \Delta}$ and $\bar{\myvec{\Sigma}}_{\myvec{\delta}\myvec{\gamma}}=\frac{1}{M}\myvec{ \Delta^T \Gamma}$ yields
\begin{equation}\label{eq:tau3} \bar{\myvec{\Sigma}}_{\myvec{\delta} \myvec{\gamma}} \bar{\myvec{\Sigma}}_{\myvec{\gamma}\myvec{\delta}} \myvec{u}_i =\lambda_i \myvec{u}_i, \end{equation} so up to scaling the left- and right-singular vectors of $\bar{\myvec{\Sigma}}_{\myvec{\gamma}\myvec{\delta}}$ provide a low-dimensional representation that captures the statistics of $\myvec{\Theta}$. Thus, based on conditions (3-5), by taking the number of points to infinity, one can extract an unbiased estimate of the representation of $\myvec{\Theta}$.

\section{Experimental results}
\label{SecExp} In this section we present experimental results
to evaluate our proposed framework. First we empirically evaluate the theoretical properties derived in section \ref{sec:CoupledV}. Then, we demonstrate how the proposed framework can be used for learning coupled manifolds even in the presence of noise.
\subsection{Empirical evaluations of theoretical aspects}
In the first group of experiments we provide empirical evidence corroborating the theoretical analysis from Section \ref{sec:CoupledV}.
\subsubsection{Spectral decay}
\label{SECEXP2} 
\begin{figure}
	
	\centering
	
	{\includegraphics[width=6cm]{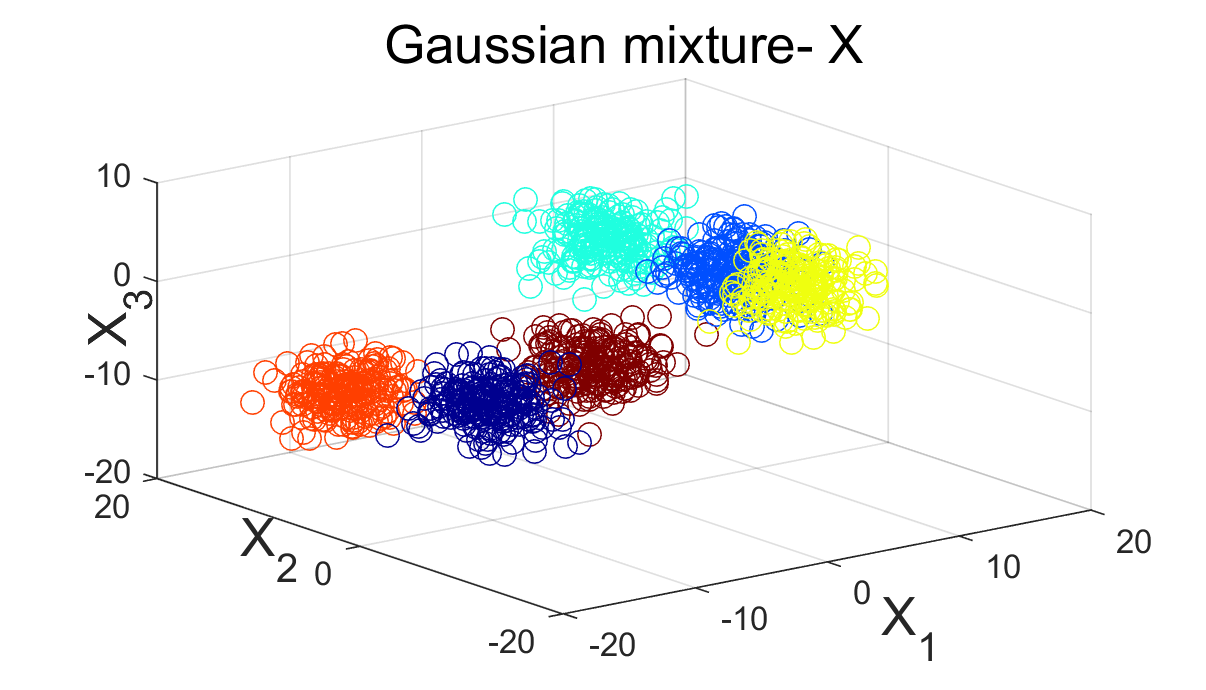}}
	{\includegraphics[width=6cm]{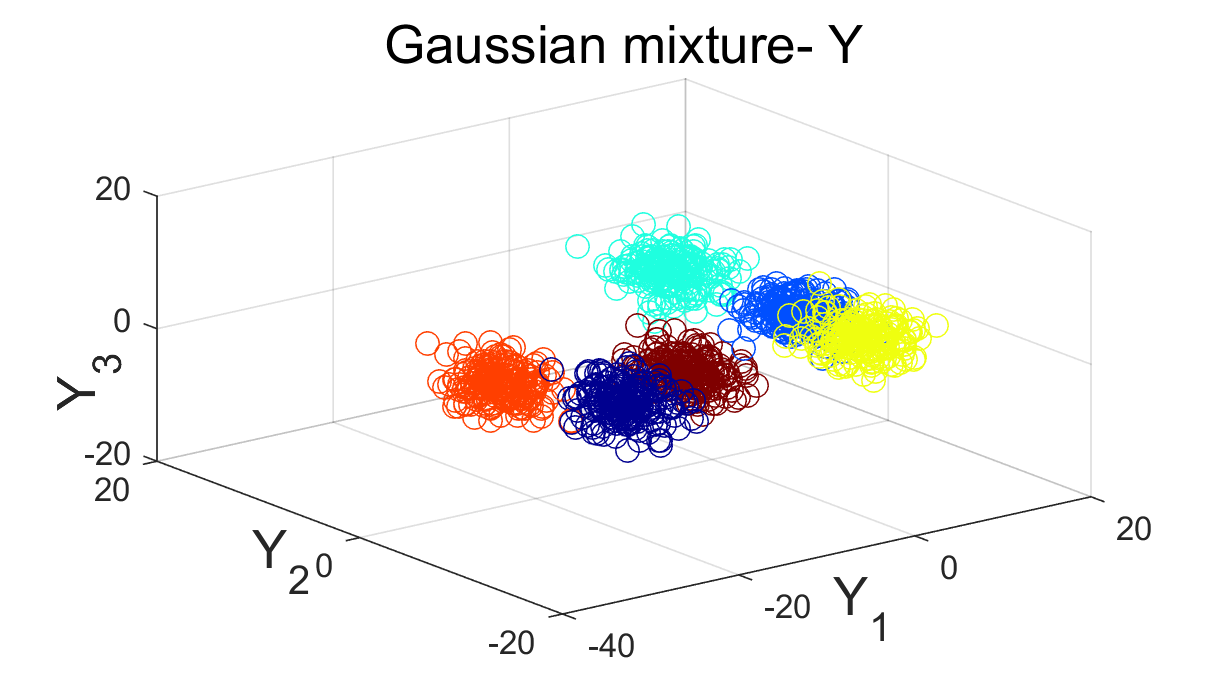}}
	
	\caption{The first 3 dimensions of the Gaussian mixture. Both views share the center of masses of
		the Gaussian spread. Left: first view denoted as $\mymat{X}$. Right: second view denoted as $\mymat{Y}$. The variance of the Gaussian in each dimension is 8.}
	\label{F4}
\end{figure}

In Section \ref{SECDEC}, an upper bound on the
eigenvalues' rate of decay for our multi-view-based approach (matrix
$\widehat{\mymat{P}}$ Eq.\ (\ref{phat})) was presented. In order to
empirically evaluate 
the spectral decay for $\widehat{\mymat{P}}$,
${\mymat{P}}^{\circ}$ (Eq.\ (\ref{EquationPDH}) and \cite{Lafon})
and ${\mymat{P}}^{+}$, we generated synthetically-clustered data drawn from Gaussian-Mixtures distributions. The following
steps describe the generation of both views, denoted
$(\mymat{X,Y})$ and referred to as View-I ($\mymat{X}$) and View-II
($\mymat{Y}$), resp.:
\begin{enumerate}
	\item {Six vectors $\myvec{\mu}_j \in {\mathbb{R}}^9, ~j=1, \ldots, 6$ were drawn from a Gaussian distribution $N(\myvec{0},8\cdot \mymat{I}_{9\times 9})$. These vectors would serve as
		the centers of masses of the generated classes.}
	\item {One hundred data points were drawn for each cluster $j=1,...,6$ from a Gaussian distribution $N(\myvec{\mu}_j,2 \cdot \mymat{I}_{9\times 9})$. Denote these $600$ data points $\mymat{X}\in\Rset^{9\times 600}$.}
	\item {One hundred additional data points were similarly drawn from each of the six Gaussian distributions $N(\myvec{\mu}_j,2 \cdot \mymat{I}_{9\times 9})$. Denote these $600$ data points$\mymat{Y}\in\Rset^{9\times 600}$.}
\end{enumerate}
The first $3$ dimensions of both views are depicted in Fig.\ \ref{F4}.
We compute the probability matrix for each view $\mymat{P}^{x}$ and
$\mymat{P}^{y}$, the Kernel Sum approach
probability matrix $\mymat{P}^{+}$, the
Kernel Product approach $\mymat{P}^{\circ}$ (Eq.\
(\ref{EquationPDH})) and the proposed approach
$\mymat{\widehat{P}}$. The eigendecomposition is computed for all
matrices. The resulting eigenvalues' decay rate are compared with
the eigenvalues product from both views. To get a fair comparison between all the methods,
we set the Gaussian scale parameters $\sigma_x$ and $\sigma_y$ in
each view  and then use these scales in all the methods. The
vectors' variance in the concatenation approach is the sum of
variances since we assume statistical independence. Therefore, the
following scale parameters $\sigma_{\circ}^2=\sigma_x^2+\sigma_y^2$
are used.

\begin{figure}
	
	\centering
	{\includegraphics[width=9cm]{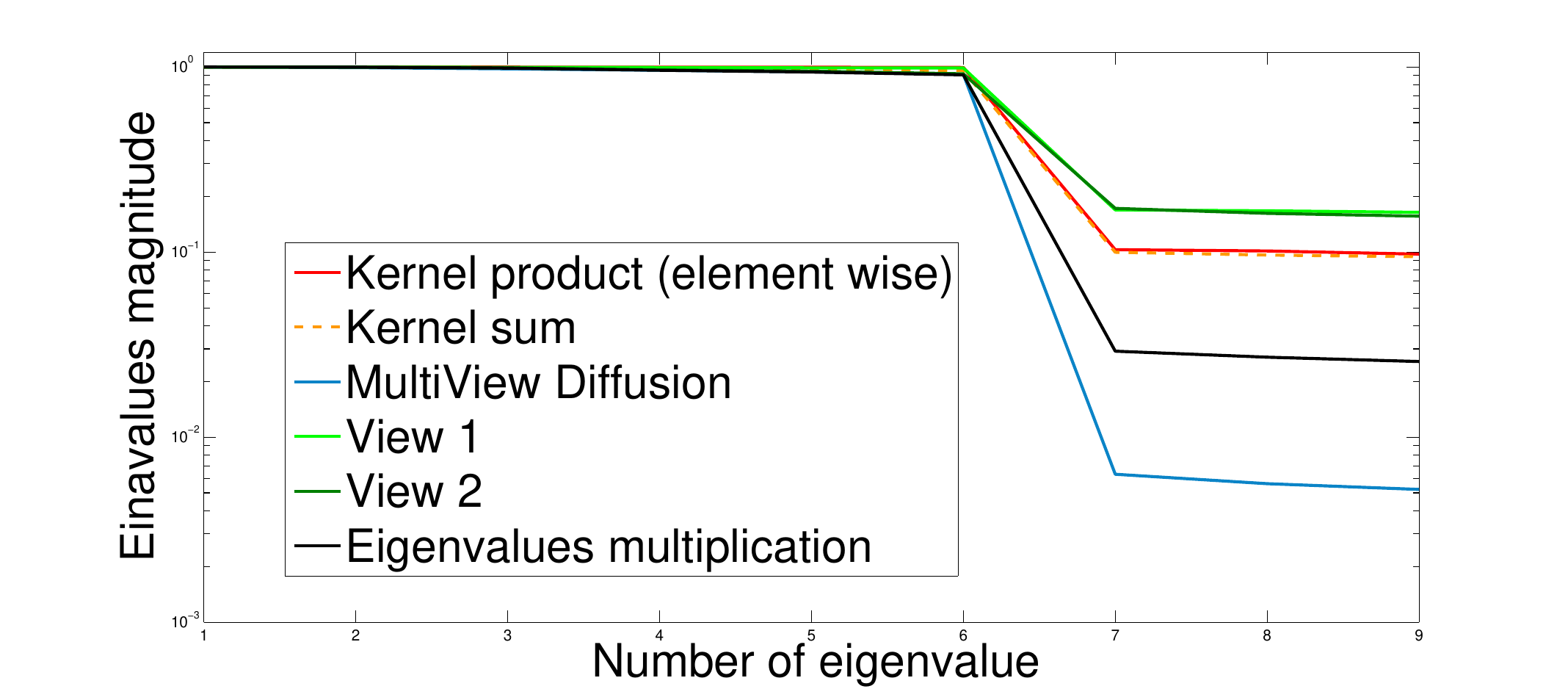}}
	\hfill
	{\includegraphics[width=9cm]{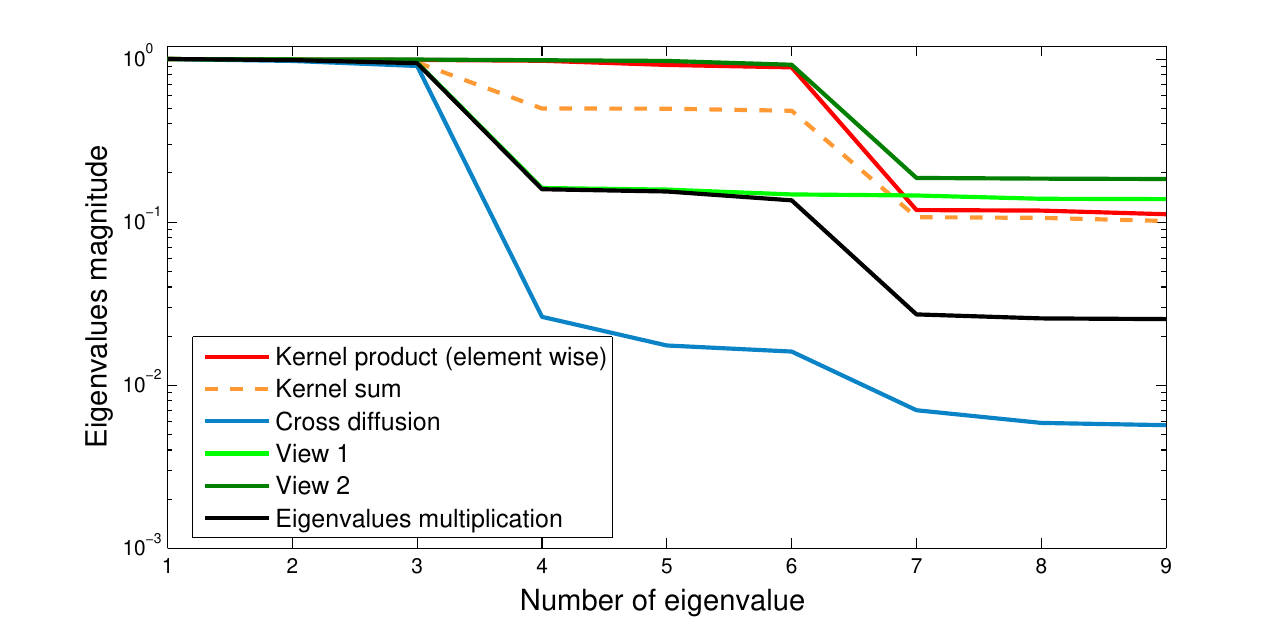}}
	
	\caption{Eigenvalues decay rate. Comparison between different
		mapping methods. Top: 6 clusters in each view. Bottom: 6 clusters in
		$\mymat{X}$ and 3 clusters in $\mymat{Y}$.} \label{F7}
\end{figure}

The experiment is repeated but this time $\mymat{X}$ contains $6$
clusters whereas $\mymat{Y}$ contains only $3$. For $\mymat{Y}$, we
use only the first $3$ centers of masses and generate 200 points in each
cluster. Figure \ref{F7} presents a logarithmic scale of the
spectral decay for eigenvalues extracted from all methods. It is
evident that our proposed kernel has the strongest spectral decay.
	\subsubsection{Cross view diffusion distance}
\begin{figure}
	
	\centering
	{\includegraphics[width=6cm]{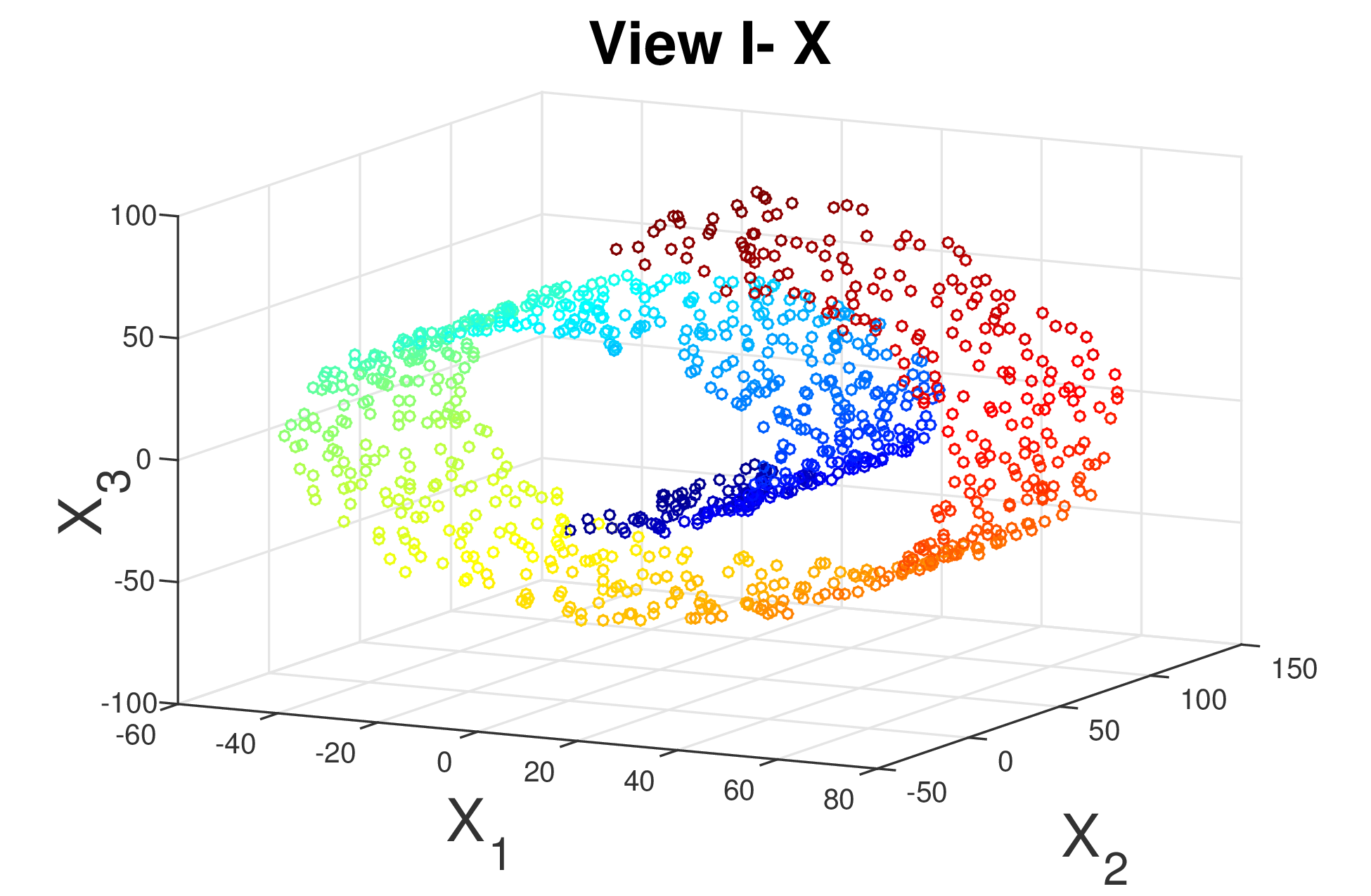}}
	{\includegraphics[width=6cm]{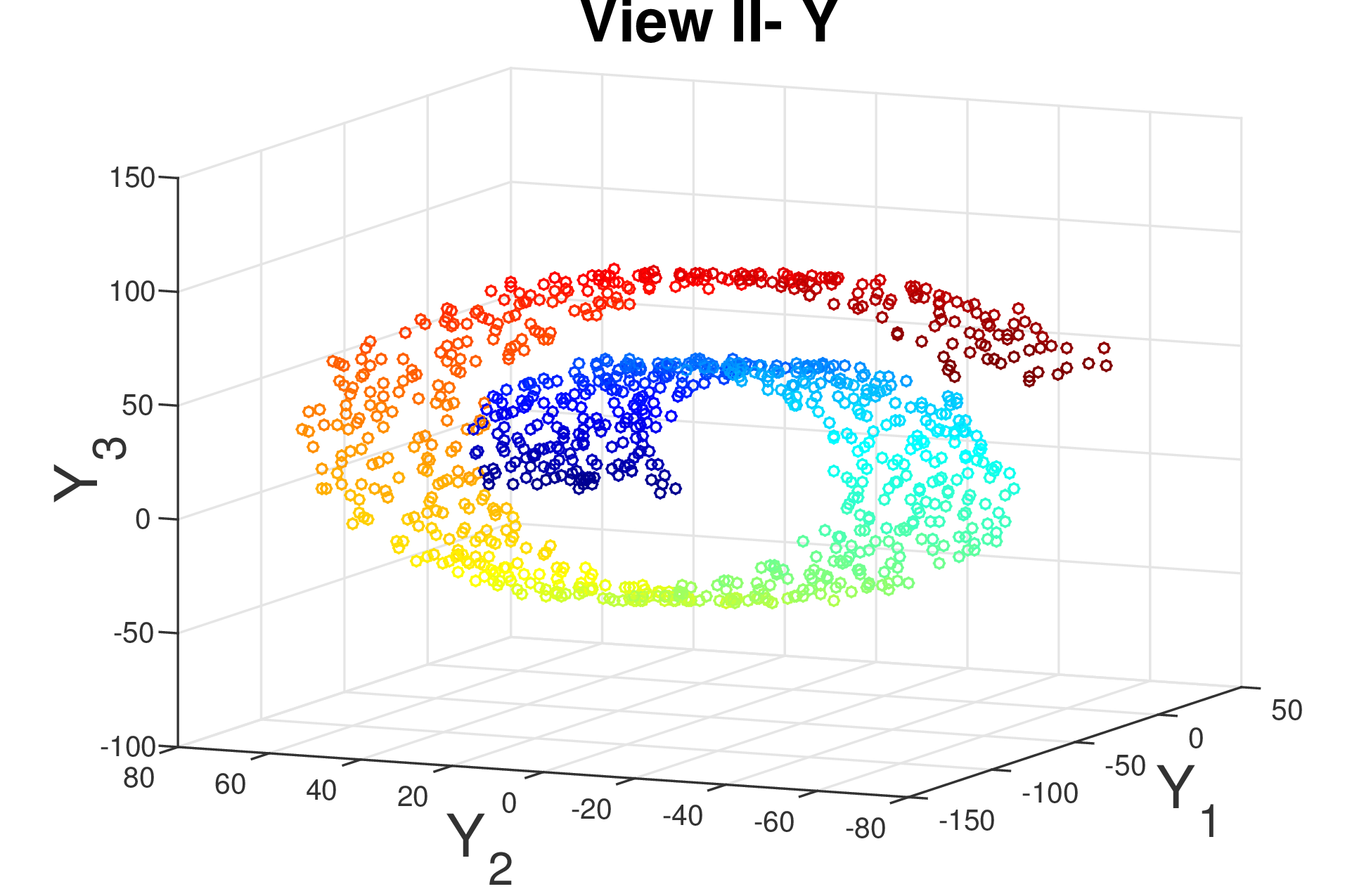}}
	
	\caption{The two Swiss Rolls, Left - a Swiss Roll generated by Eq.\ (\ref{EQSwissX}), Right - a Swiss Roll generated by Eq.\ (\ref{EQSwissY}).}
	\label{FSwiss}
\end{figure}

In this section, we examine the proposed Cross View Diffusion
Distance (Section \ref{CVDD}). A Swiss Roll is generated by using
the function
\begin{equation}
\label{EQSwissX}
\text{View I:   }
\mymat{x}_i=
\begin{bmatrix}
{x_i}[1]\\
{x_i}[2]\\
{x_i}[3]\\

\end{bmatrix}
=
\begin{bmatrix}
{6\theta_i\cos(\theta_i)}\\
{h_i}\\
{6\theta_i\sin(\theta_i)}\\

\end{bmatrix}
+\myvec{n}_i^{(1)},
\end{equation}
with ${\theta_i}=(1.5\pi) s_i $, $i=1,2,...,1,\!000$, where $s_i$ are $1000$
data points evenly spread along the segment  $[1,3]$, and where $\myvec{n}_i^{(1)}$ are i.i.d.\ zero-mean Gaussian noise vectors with covariance $\sigma_N^2\cdot\mymat{I}_{3\times 3}$. The second view is generated by applying an
orthonormal transformation to the (noiseless) Swiss Roll and similarly adding Gaussian
noise:
\begin{equation}
\label{EQSwissY}
\text{View II:   }
\mymat{y}_i=
\begin{bmatrix}
{y_i}[1]\\
{y_i}[2]\\
{y_i}[3]\\

\end{bmatrix}
=\mymat{R}
\begin{bmatrix}
{6\theta_i\cos(\theta_i)}\\
{h_i}\\
{6\theta_i\sin(\theta_i)}\\

\end{bmatrix}
+\myvec{n}_i^{(2)},
\end{equation}
where $\mymat{R}\in \mathbb{R}^{3 \times 3}$ is a random orthonormal
transformation matrix, and where $\myvec{n}_i^{(2)}$ are i.i.d.\ $N(\myvec{0},\sigma_N^2\cdot\mymat{I}_{3\times 3}$. The matrix $\mymat{R}$ is generated by independently drawing its elements from a standard Gaussian distribution, followed by applying the Gram-Schmidt orthogonalization procedure. The variables $h_i, i=1,...,1000$ are drawn from a uniform distribution in the interval $[0,100]$. An
example for both Swiss Rolls is shown in Fig.\ \ref{FSwiss}.

A standard DM is applied to each view and a $2$-dimensional
embedding of the Swiss Roll is extracted. The sum of distances
between all the data points in the embedding spaces is denoted as a
single-view diffusion distance (SVDD). The distance is computed
using the measure
\begin{equation}{ \label{SVDD} { {{{\cal{D}}_t}^{\rm (SV)}}^2(X,Y)=\sum \limits_{i=1}^{M}||{\Psi_t{(\myvec{x}_i)}}-{\Psi_t{(\myvec{y}_i)}}||^2}} ,
\end{equation} where $\Psi_t{(\myvec{x}_i)},\Psi_t{(\myvec{y}_i)}, i=1,...,M$ are the single-view diffusion mappings. 
Then, the proposed framework is applied to extract the coupled embedding.
A Cross View Diffusion Distance (CVDD) is computed using Eq.\ (\ref{ECVDD}).
This experiment was executed 100 times for various values of the Gaussian noise variance $\sigma_N^2$.

\begin{figure}
	
	\centering
	{\includegraphics[width=8cm]{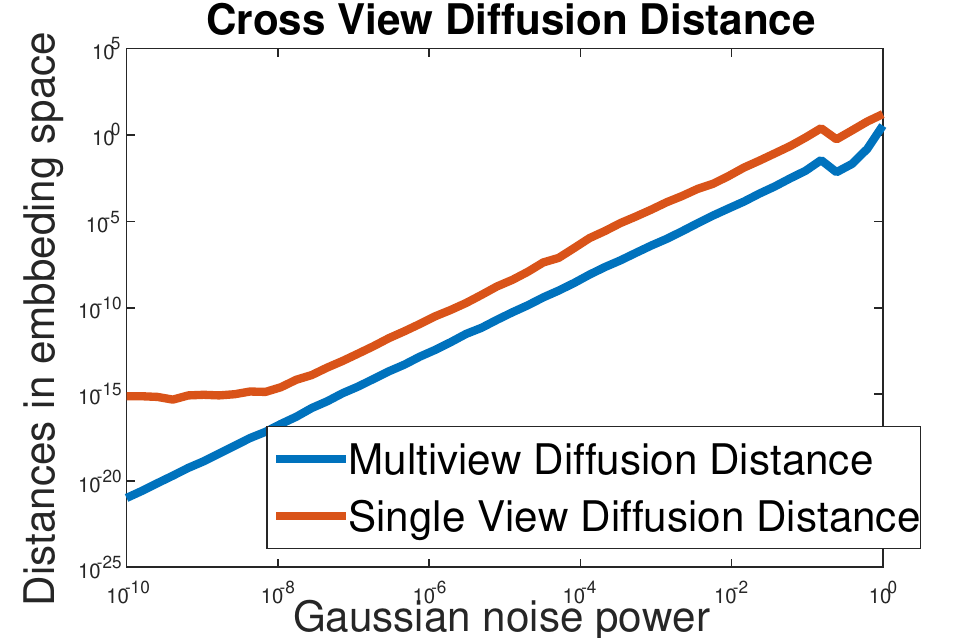}}

	\caption{Comparison between two cross view diffusion based distances. Simulated on two Swiss Rolls with additive Gaussian noise.
		The results are the median of 100 simulations. }
	\label{Forth}
\end{figure}

In about $10\%$ of the single-view trials the embeddings' axis are flipped. This generates a large SVDD although the embeddings share similar structures. In order to mitigate the effect of this type of errors we used the Median of the measures taken from the $100$ trialss, presented in Fig.\ \ref{Forth}.
	
\subsection{Manifold learning}
In this subsection we demonstrate how our proposed Multi-view DM framework allows to simultaneously extract $L$ low-dimensional representations for $L$ datasets.

\subsubsection{Artificial manifold learning}
The general DM approach is based on an underlying assumption, that the sampled
space describes a single low-dimensional manifold. However, this
assumption may be incorrect if the sampled space describes the
existence of redundancy in the manifold, or more generally, if the
sampled space can describe two or more manifolds generated by a
common physical process. In this subsection we consider such cases. We examine the extracted
embedding computed
using our method and compare it to the Kernel Product approach (Subsection \ref{sec:KernelProd}).
\\
\par
\begin{figure}

	\centering
	
	{\includegraphics[width=6cm]{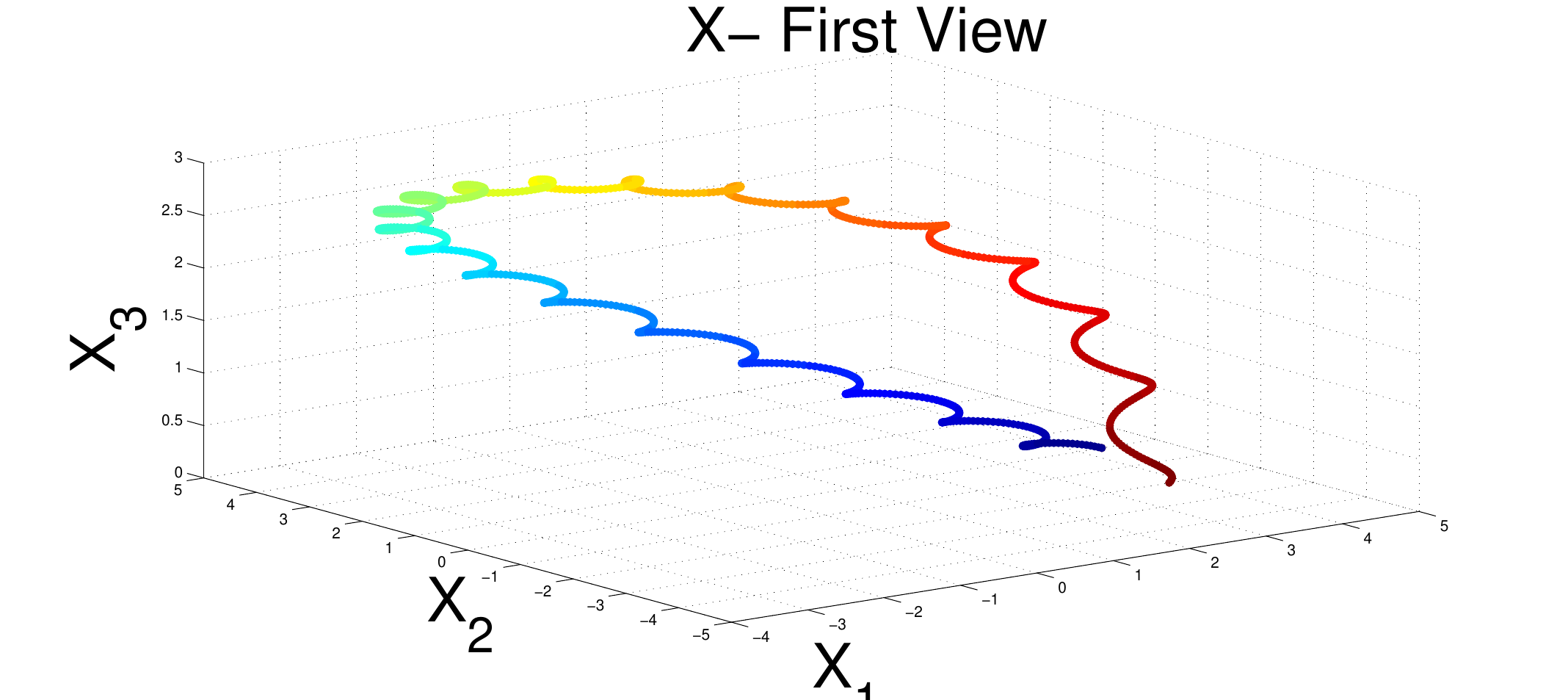}}
	{\includegraphics[width=6cm]{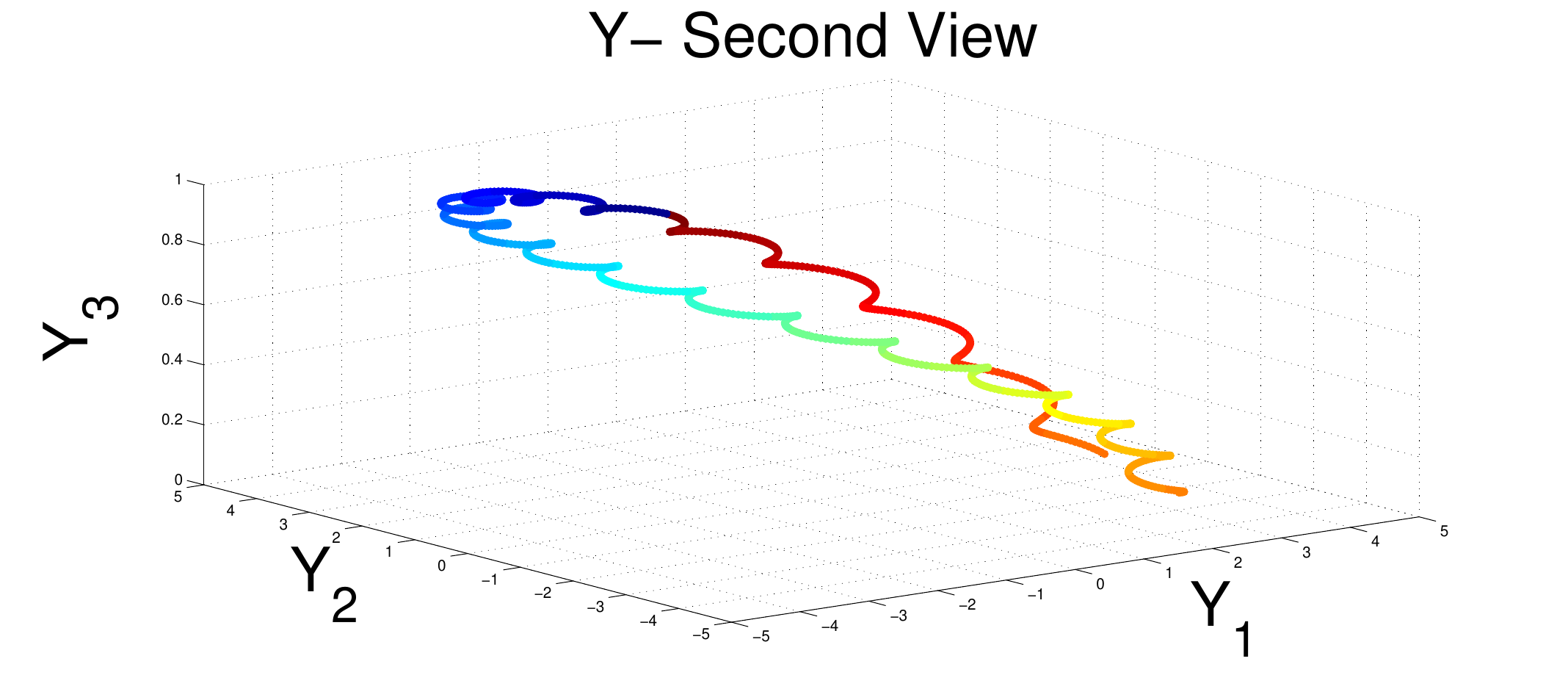}}
	
	\caption{Left: first Helix $\mymat{X}$ (Eq.\ (\ref{E19})). Right:
		second Helix  $\mymat{Y}$ (Eq.\ (\ref{E18})). Both manifolds have
		some circular structure governed by the angle parameter $a[i]$ and
		$b[i]$, $i=1,2,...,1,\!000$ colored by the points index $i$.}
	\label{ViewsI}
\end{figure}

\noindent
{\bf{Helix A}}\\
Two coupled manifolds with a common underlying open circular
structure are generated. The helix shaped manifolds were generated
by the application of a $3$-dimensional function to $M=1,\!000$ data points $\{a_i,b_i\}_{i=1}^M$, such that the $\{a_i\}$ are evenly spread in  $[0,2\pi]$
and ${b_i}=(a_i+0.5\pi)\mod2\pi $, $i=1,...,M$. The following functions
are used to generate the
datasets for View-I and View-II denoted as $\mymat{X}$ and
$\mymat{Y}$, resp.:

\begin{equation}
\label{E19}
\text{View I:   }
\mymat{x}_i=
\begin{bmatrix}
{x_i}[1]\\
{x_i}[2]\\
{x_i}[3]\\

\end{bmatrix}
=
\begin{bmatrix}
{4\cos(0.9a_i)+0.3\cos(20a_i)}\\
{4\sin(0.9a_i)+0.3\sin(20a_i)}\\
{0.1(6.3a_i^2-a_i^3)}\\

\end{bmatrix},i=1,2,...,1,\!000,
\end{equation}

\begin{equation}
\label{E18}
\text{View II:   }
\mymat{y}_i=
\begin{bmatrix}
{y_i}[1]\\
{y_i}[2]\\
{y_i}[3]\\

\end{bmatrix}
=
\begin{bmatrix}
{4\cos(0.9b_i)+0.3\cos(20b_i)}\\
{4\sin(0.9b_i)+0.3\sin(20b_i)}\\
{0.1(6.3b_i-b_i^2)}\\

\end{bmatrix},i=1,2,...,1,\!000.
\end{equation}
The resulting $3$-dimensional Helix-shaped manifolds $\mymat{X}$ and $\mymat{Y}$ are shown in Fig.\ \ref{ViewsI}.

\begin{figure}
	
	\centering
	
	{\includegraphics[width=6cm]{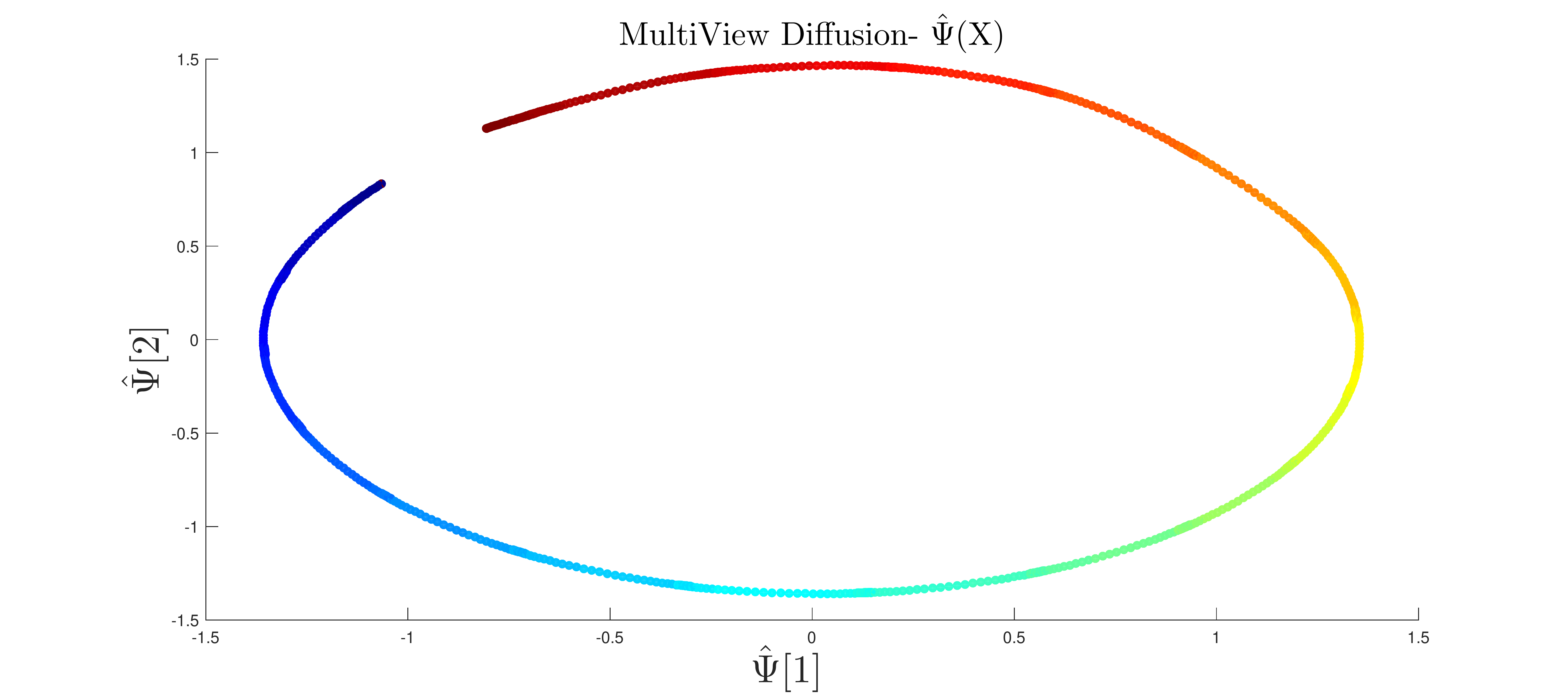}}
	{\includegraphics[width=6cm]{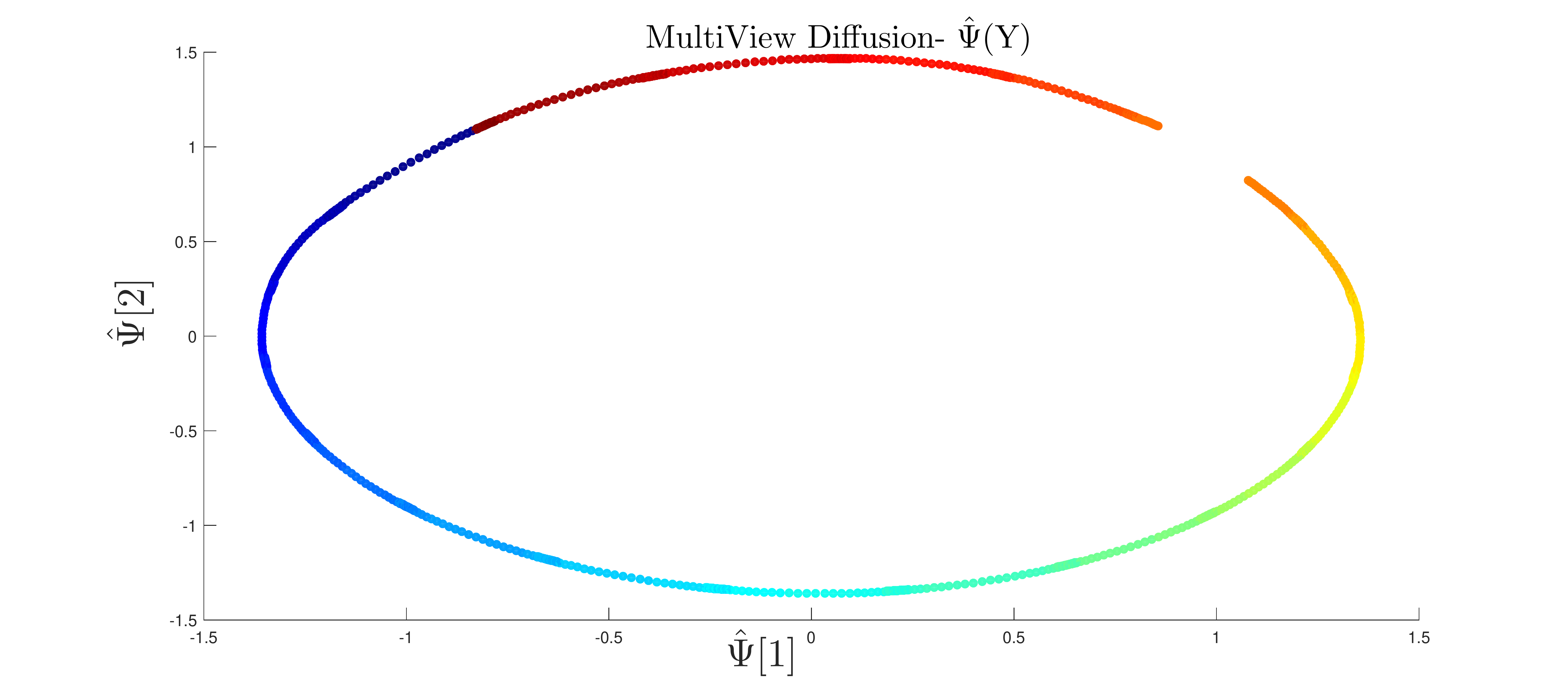}}
	\caption{Left: Multi-View based embedding of the first view
		$\widehat{\mymat{\Psi}}(\mymat{X})$. Right: Multi-View based
		embedding of the second view $\widehat{\mymat{\Psi}}(\mymat{Y})$.
		They were computed by using Eq.\ (\ref{Map1})),
		respectively.} \label{FigMap1}
\end{figure}

\begin{figure} 
	
	\centering
	\includegraphics[width=6cm]{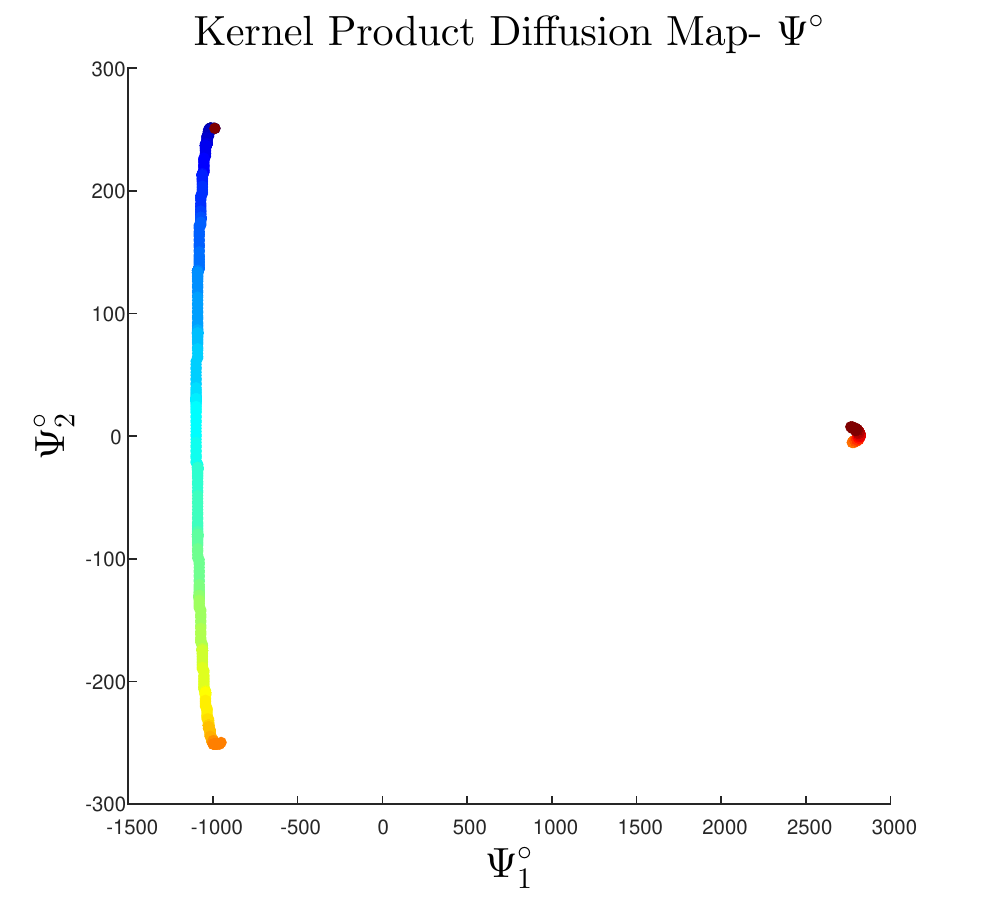}
	
	\caption{2-dimensional DM-based mapping of the Helix computed using the concatenated vector from both views that correspond to the kernel $\mymat{P}^{\circ}$ (Eq.\ (\ref{EquationPDH})).}
	\label{FigCon}
\end{figure}

The Kernel Product mapping (Eq.\ (\ref{EquationPDH})) separates the
manifold to a bow and a point as shown in Fig.\ \ref{FigCon}. This
structure neither represents any of the original structures nor
reveals the underlying parameters $a_i,b_i$. On the other hand, our
embedding (Eq.\ (\ref{Map1})) captures the two
structures. one for each view. As shown in Fig.\ \ref{FigMap1}, one
structure represents the angle of $a_i$ while the other represents
the angle of $b_i$. The Euclidean distance in the new spaces
preserves the mutual relations between data points based on the
geometrical relation in both views. Moreover, both manifolds are in
the same coordinate system and this is a strong advantage as it
enables to compare the manifolds in the lower-dimensional
space. The Euclidean distance in the new spaces preserves the mutual
relations between data points that are based on the geometrical
structure of both views.

\par
\begin{figure}

	\centering
	
	{\includegraphics[width=6cm]{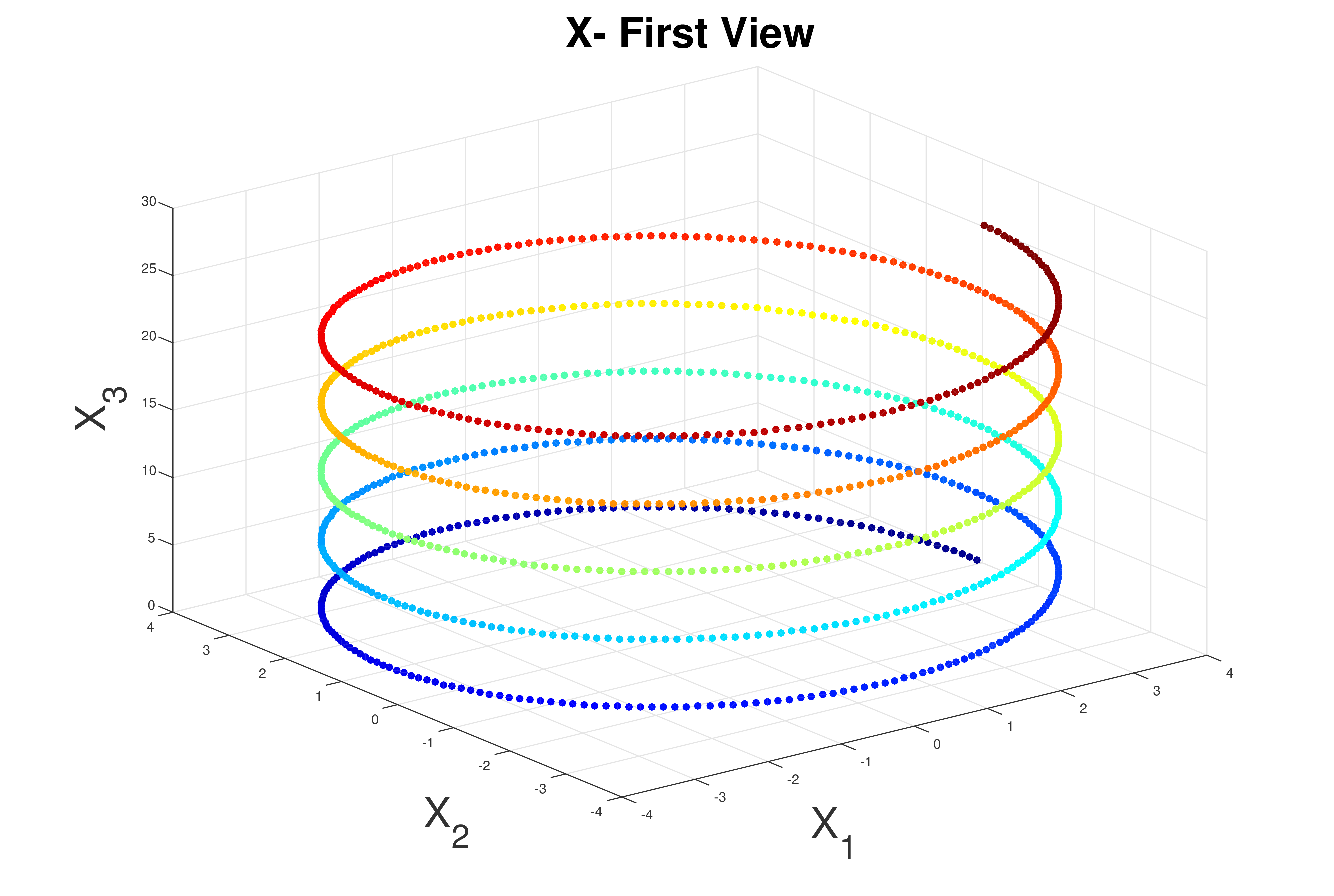}}
	{\includegraphics[width=6cm]{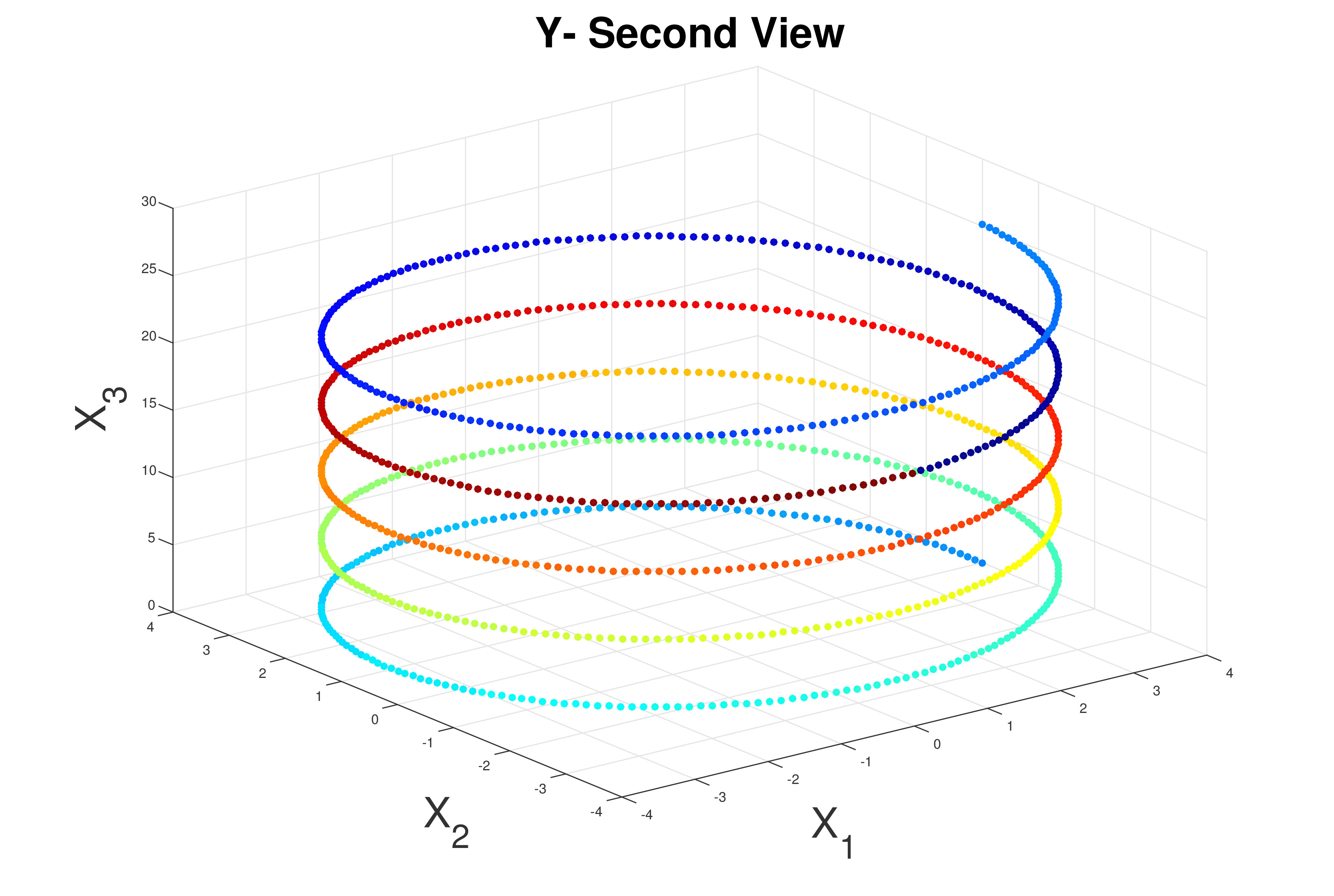}}
	
	\caption{Left: first Helix $\mymat{X}$ (Eq.\ (\ref{E20})). Right: second Helix  $\mymat{Y}$ (Eq.\ (\ref{E21})).
		Both manifolds have some circular structure governed by the angle parameter $a[i]$ and $b_i$, $i=1,2,...,1,\!000$, as colored by the point's index $i$.}
	\label{ViewsII}
\end{figure}

\noindent
{\bf{Helix B}}\\
The previous experiment was repeated using the following alternative functions:
\begin{equation}
\label{E20}
\text{View I:   }
\mymat{x}_i=
\begin{bmatrix}
{x_i}[1]\\
{x_i}[2]\\
{x_i}[3]\\

\end{bmatrix}
=
\begin{bmatrix}
{4\cos(5a_i)}\\
{4\sin(5a_i)}\\
{4a_i}\\

\end{bmatrix},
\end{equation}

\begin{equation}
\label{E21}
\text{View II:   }
\mymat{y}_i=
\begin{bmatrix}
{y_i}[1]\\
{y_i}[2]\\
{y_i}[3]\\

\end{bmatrix}
=
\begin{bmatrix}
{4\cos(5b_i)}\\
{4\sin(5b_i)}\\
{4b_i}\\

\end{bmatrix}.
\end{equation}
Again, $M=1,\!000$ points were generated using ${a_i}\in [0,2\pi]$, ${b_i}=(a_i+0.5\pi)\mod2\pi $, $i=1,...,M$. The
generated manifolds are presented in Fig.\ \ref{ViewsII}.

\begin{figure}
	
	\centering
	{\includegraphics[width=6cm]{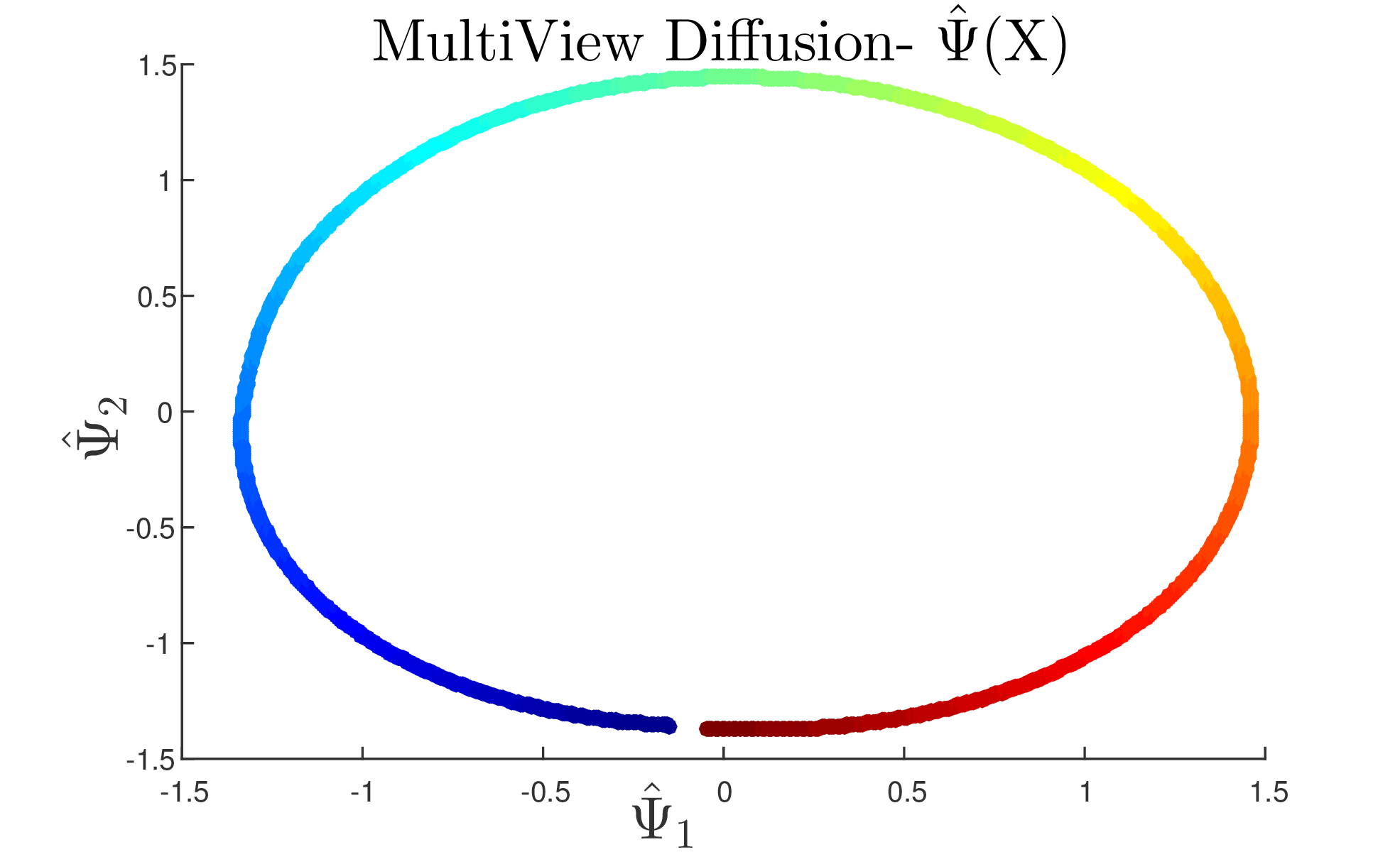}}
	{\includegraphics[width=6cm]{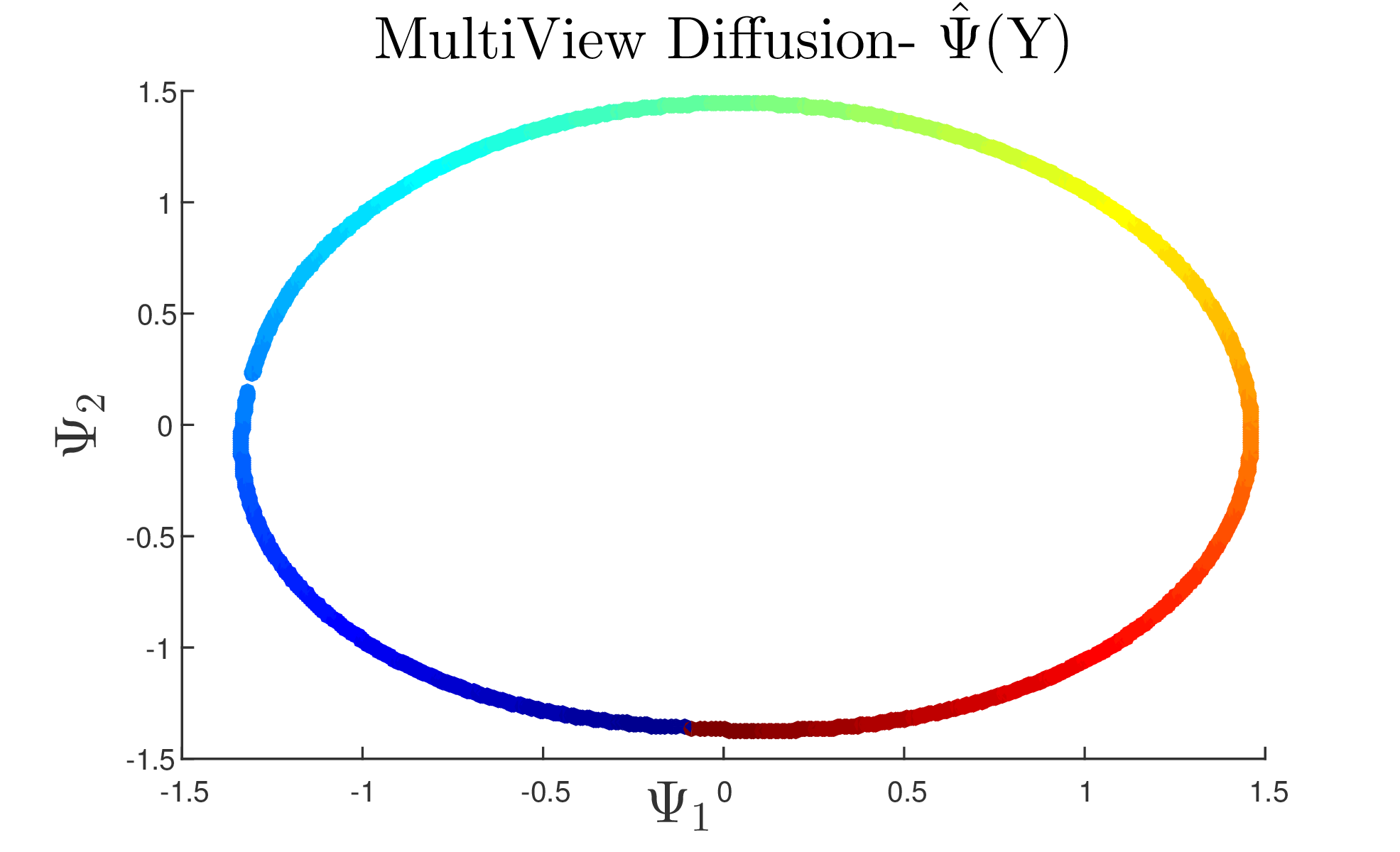}}
	\caption{The coupled mappings computed using our proposed parametrization in Eq.\ \ref{Map1}}
	\label{FigMapII}
\end{figure}

\begin{figure} 
	
	\centering
	\includegraphics[width=6cm]{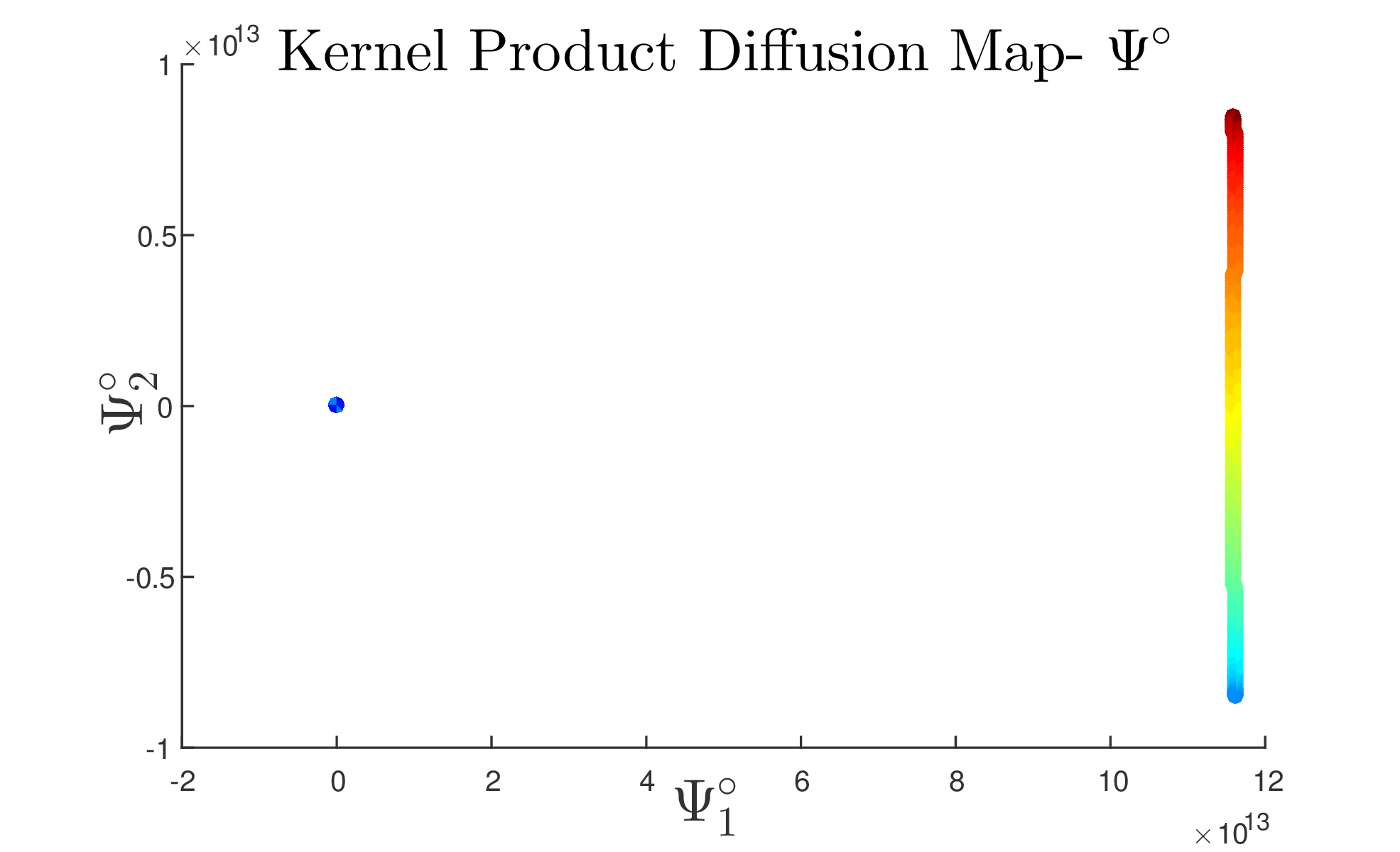}

	\caption{A 2-dimensional mapping, extracted based on $\mymat{P}^{\circ}$ (Eq.\ (\ref{EquationPDH})).}
	\label{FigConII}
\end{figure}
As can be viewed in Fig.\ \ref{FigMapII}, our proposed embeddings
(Eq.\ (\ref{Map1}) has successfully captured the
governing parameters $a_i$ and $b_i$. The Kernel Product based
embedding (Eq.\ (\ref{EquationPDH})), as evident in Fig.\
\ref{FigConII}, again separated
the data points into two unconnected structures that do not
represent well the parameters.
\subsubsection{MultiView video sequence}
Various examples involving datasets in diverse fields, such as images, audio, MRI (\cite{Keller},
\cite{lindenbaum2015musical} and \cite{{piella2014diffusion}}, resp.) have
demonstrated the power of DM for the extraction of underlying changing physical parameters from real datasets. In this experiment, the
multi-view approach is tested  on a real video data, in what can literally be termed a ``toy example": 

Two web cameras and a
toy train with preset tracks are used. The train's tracks have an
``eight'' shape structure. Extracting the underlying manifold from
the set of images enables to organize the images according to the
location along the train's path and thus reveals the true underlying
parameters of the processes.

\begin{figure} 
	\centering
	\includegraphics[width=6cm]{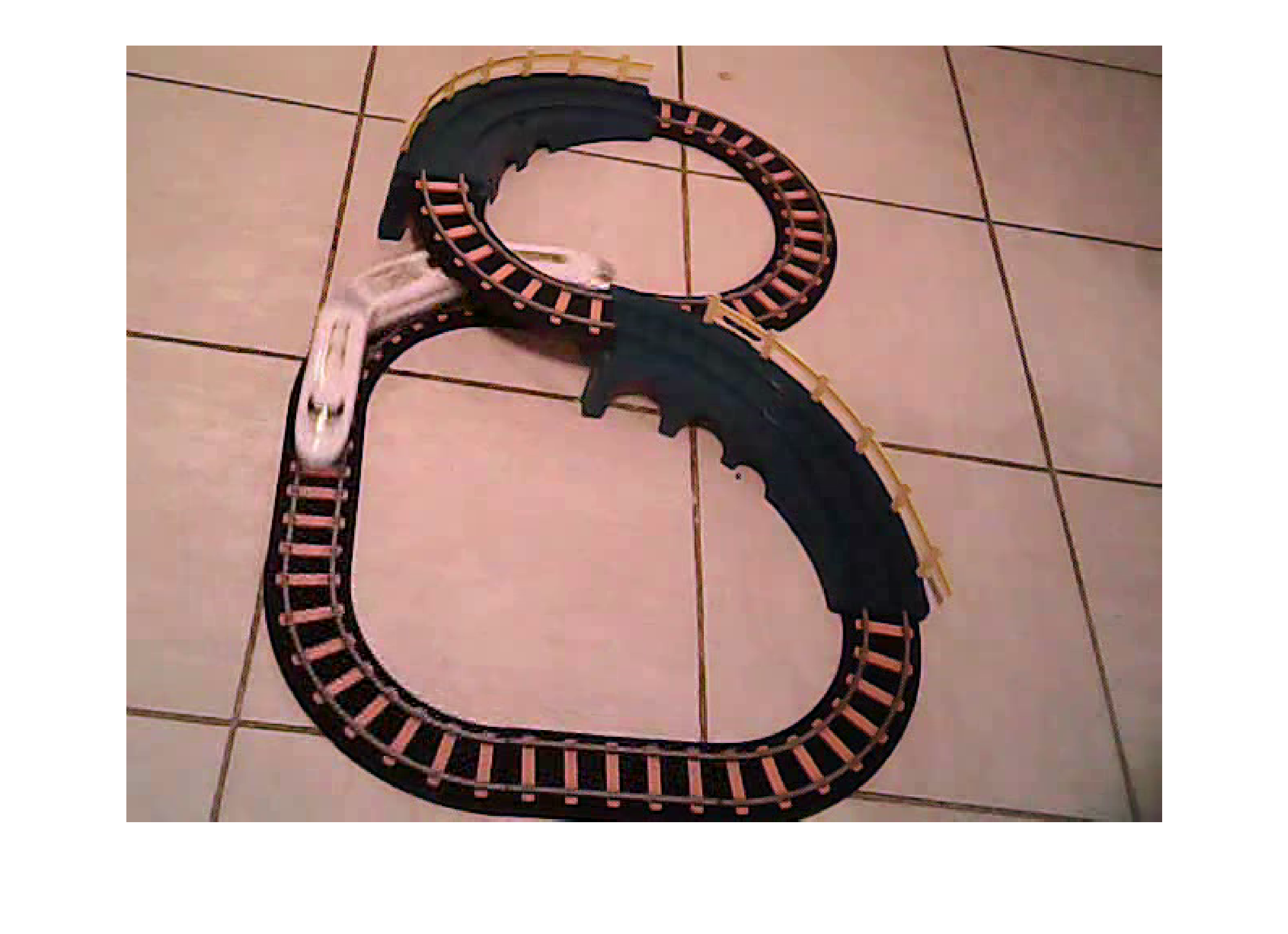}
	\includegraphics[width=6cm]{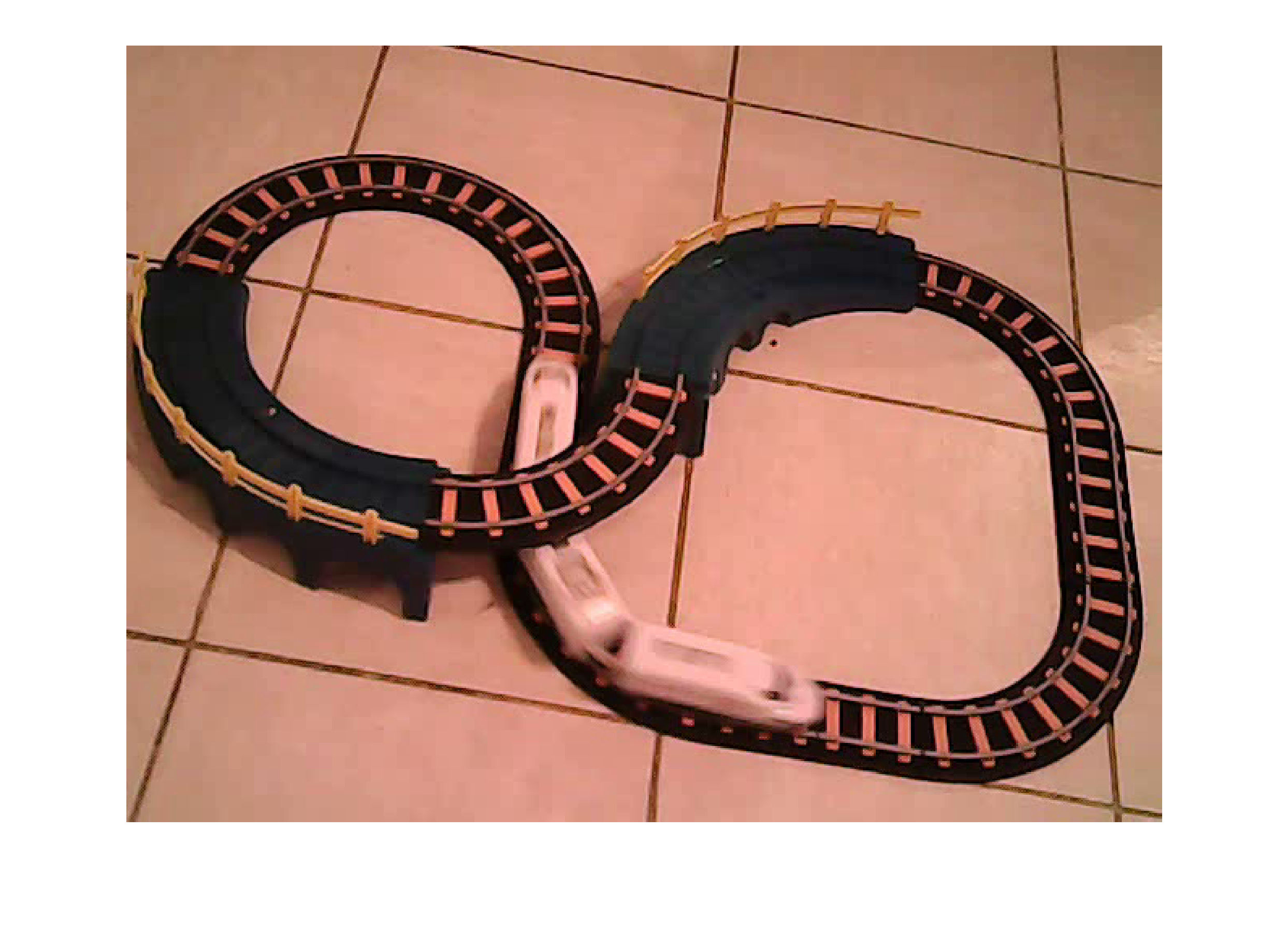}
	\caption{Left: a sample image from the first camera ($\mymat{X}$). Right: a sample image from the second camera ($\mymat{Y}$).}
	\label{FigTrain1}
\end{figure}

\begin{figure} 
	\centering
	\includegraphics[width=6cm]{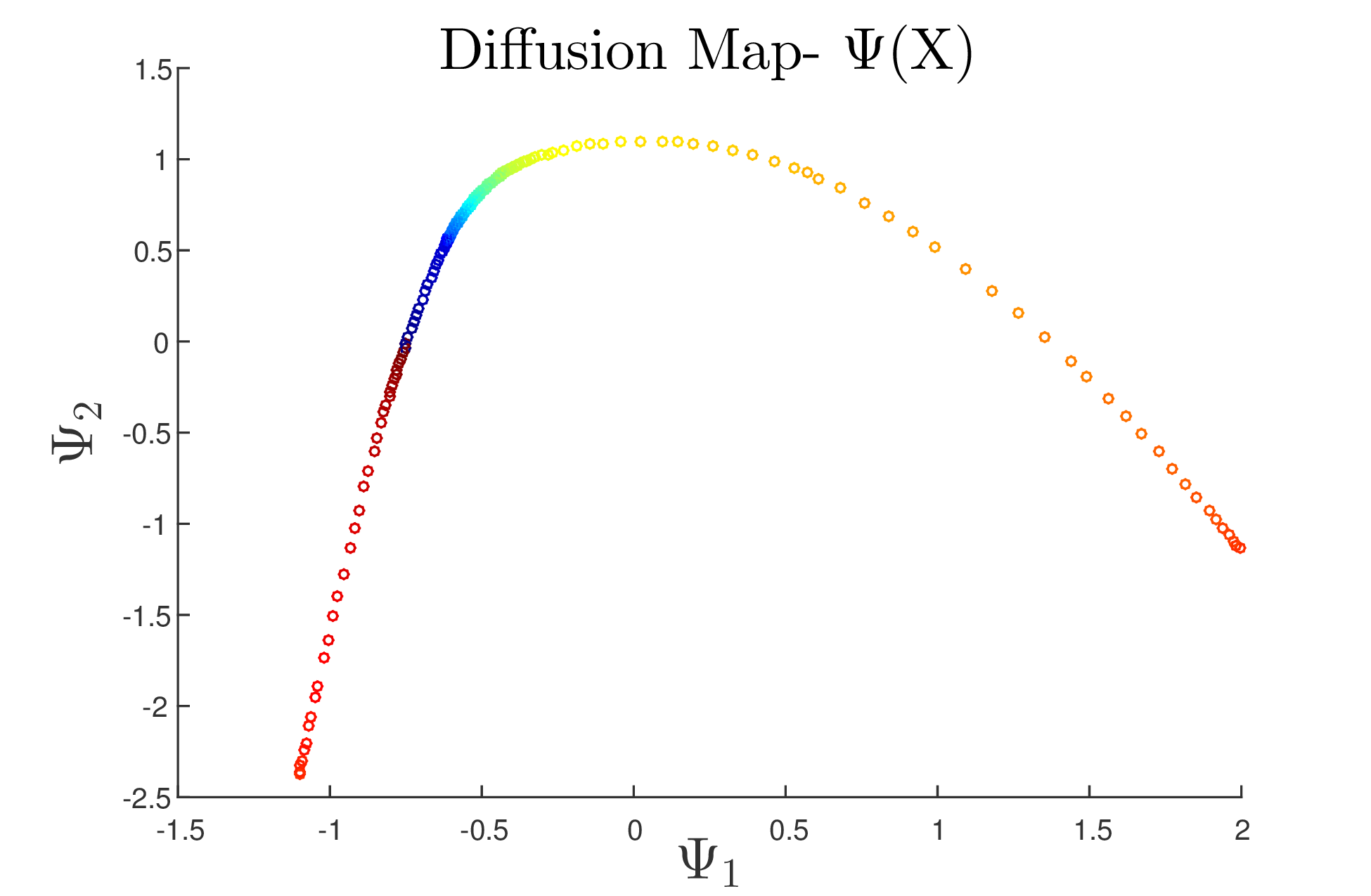}
	\includegraphics[width=6cm]{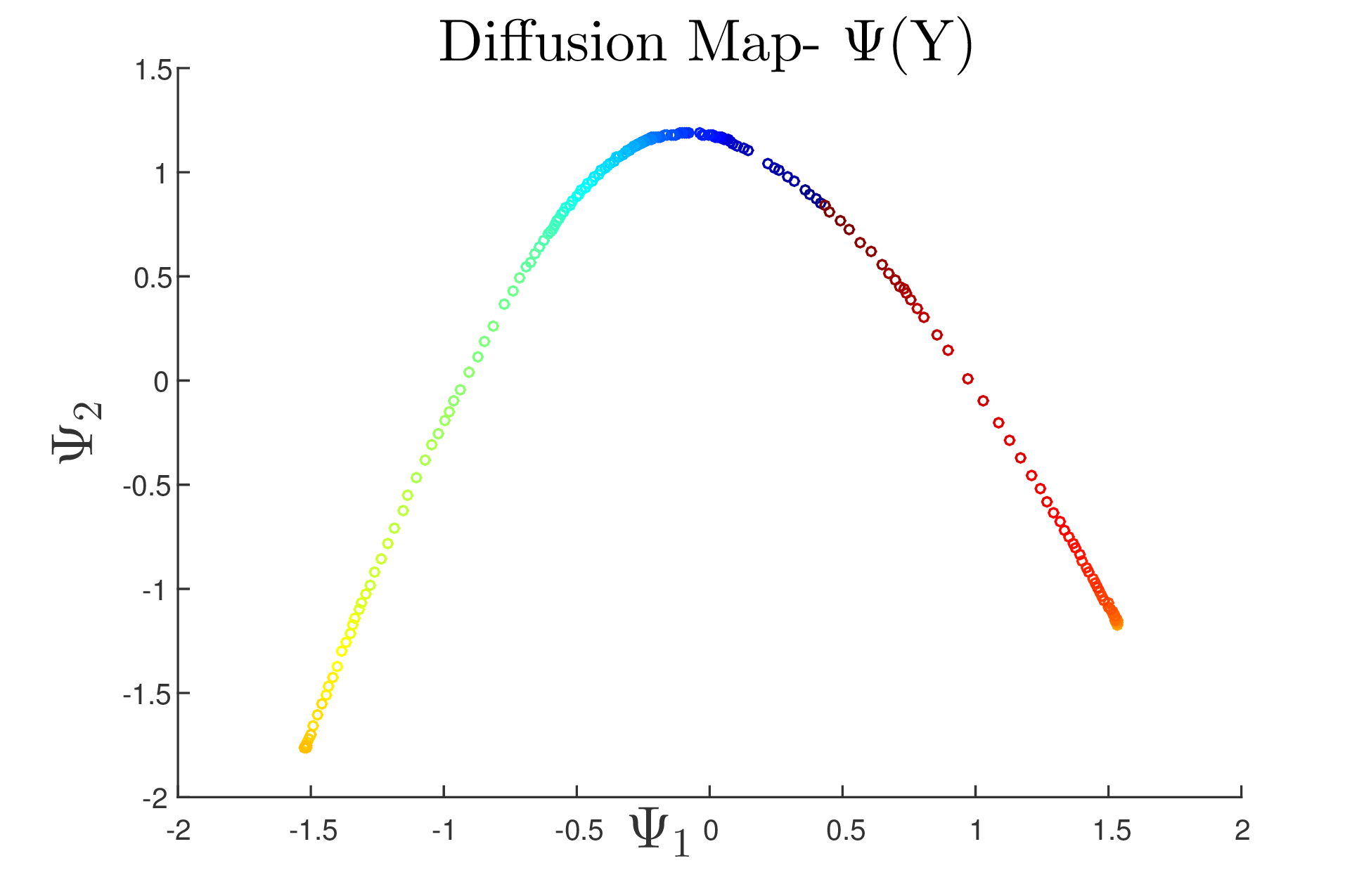}
	\caption{Left: DM-based single-view mapping $\Psi(\mymat{X})$. Right: DM-based single-view mapping $\Psi(\mymat{Y})$). The removed images caused a bow shaped structure.}
	\label{FigTrainSV}
\end{figure}

\begin{figure}
	\centering
	\includegraphics[width=6cm]{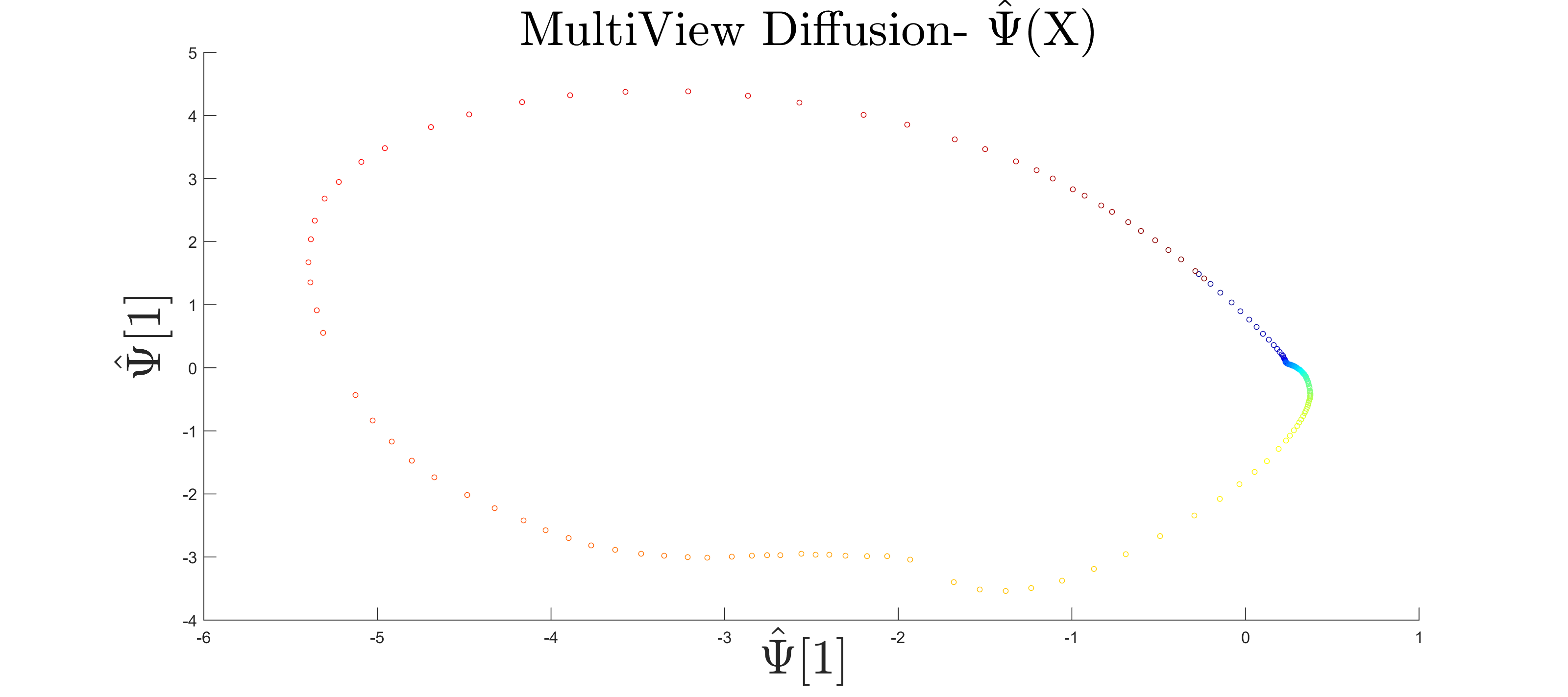}
	\includegraphics[width=6cm]{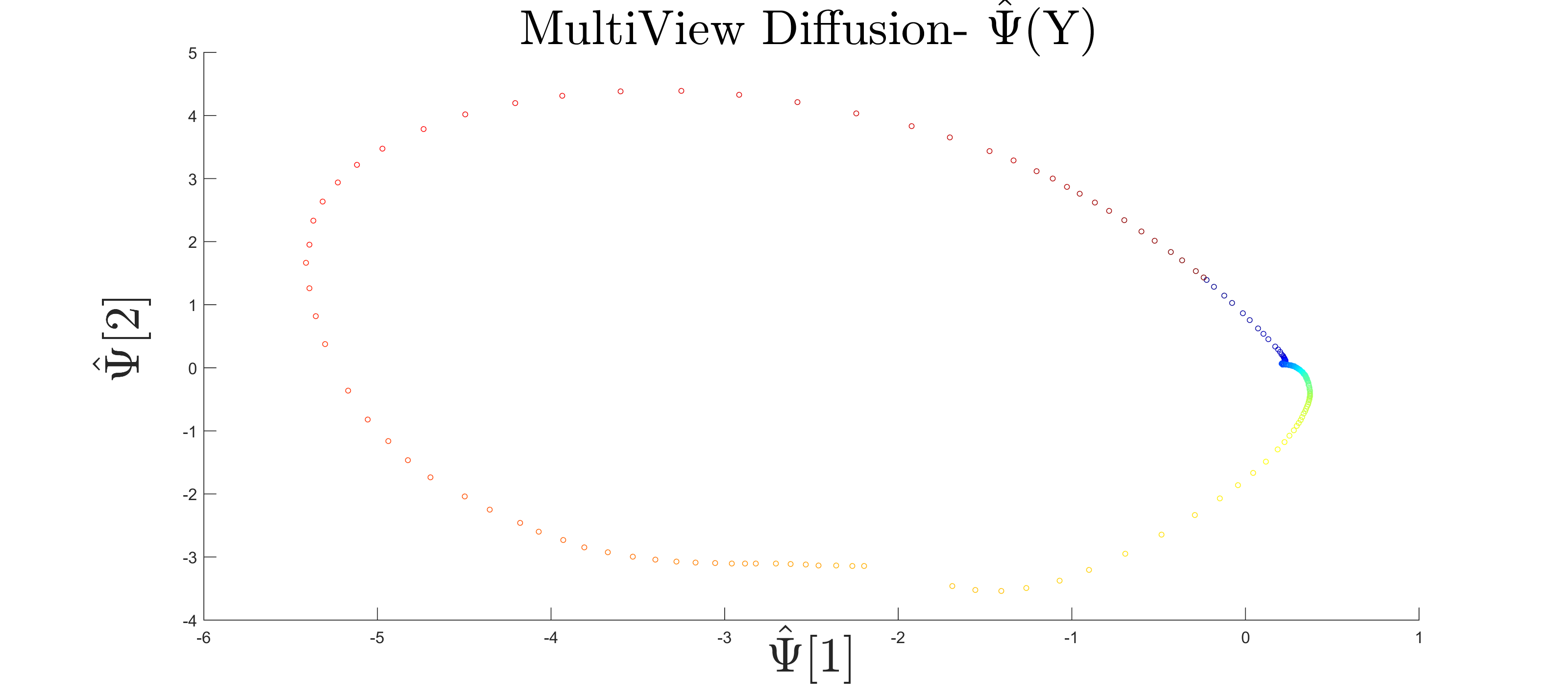}
	\caption{Left: Mapping $\hat{\Psi}(\mymat{X})$. Right: Mapping $\hat{\Psi}(\mymat{Y})$ as extracted by the multi-view based framework.
		Two small gaps, which correspond to the removed images, are visible.}
	\label{FigTrainMV}
\end{figure}

\begin{figure}
	\centering
	\includegraphics[width=6cm]{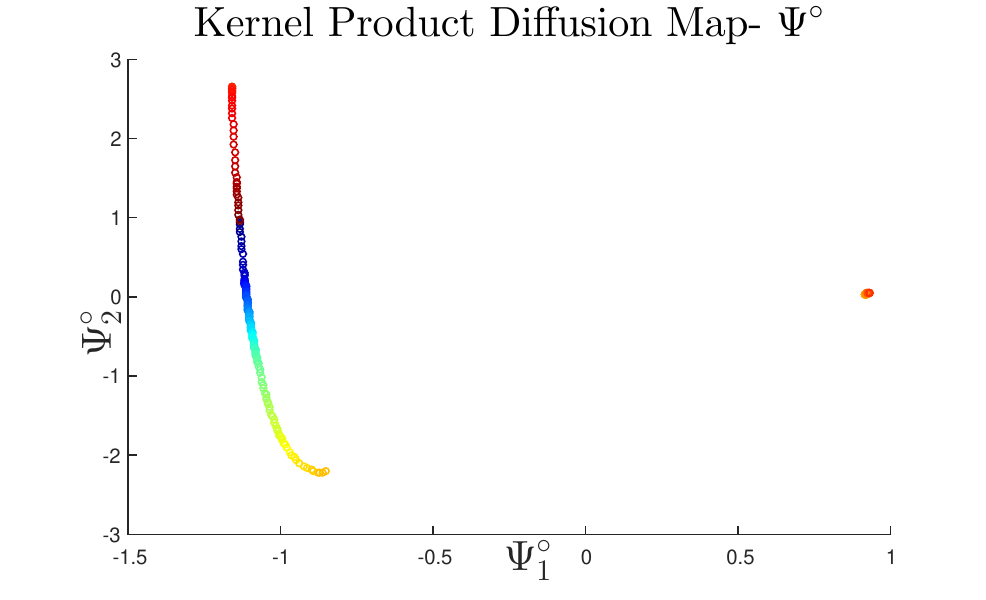}
	\caption{A standard diffusion mapping (Kernel Product-based) that was computed by using the concatenated
		vector from both views that correspond to kernel $\mymat{K}^{\circ}$.}
	\label{FigTrainKP}
\end{figure}

The setting of the experiment is as follows: each camera records a
set of images from a different angle. A sample frame from each view
is shown in Fig.\ \ref{FigTrain1}. The video is sampled at $30$
frames per second  with a resolution of $640\times 480$ pixels per
frame. $M=220$ images were collected from each view (camera). Then,
the R,G,B values were averaged and downsampled to $160\times 120$ pixels
resolution. The matrices were reshaped into column vectors. The
resulted set of vectors are denoted by $\mymat{X}$ and $\mymat{Y}$
where $\myvec{x}_i,\myvec{y}_i\in \Rset^{19,\!200},~1\leq i\leq 220$. The sequential order
of the images is not important for the algorithm. In a normal
setting, one view is sufficient to extract the parameters that
govern the movement of the train and thus extract the natural order
of the images. However, we use two types of interferences to create
a scenario in which each view by itself is insufficient for the
extraction of the underlying parameters. The first interference is a
gap in the recording of each camera. We remove $20$ consecutive frames
from each view at different time intervals. By removing frames, the
bijective correspondence of some of the images in the sequence is
broken. However, even an approximated correspondence is sufficient
for our proposed manifold extraction. A standard $2$-dimensional DM
based mapping of  each view was extracted. The results are bow-shaped
manifolds as presented in Fig.\ \ref{FigTrainSV}. Applying DM
separately to each view extracts the correct order of the data
points (images) along the path. However, the ``missing'' data points
broke the circular structure of the expected manifold and resulted
in a bow-shaped embedding. We use the multi-view based methodology to
overcome this interference by application of the multi-view framework to extract two coupled mappings (Eq.\ (\ref{Map1})). The results are shown in Fig.\ \ref{FigTrainMV}.
The proposed approach overcomes the interferences by smoothing out the
gap inherited in each view through the use of connectivities from
the ``unonbstructed'' view. Finally, we concatenate the vectors from
both views and compute the Kernel Product embedding The results are
presented in Fig.\ \ref{FigTrainKP}. Again, the structure of the
manifold is distorted and incomplete due to the missing images.


\begin{figure} 
	\centering
	\includegraphics[width=6cm]{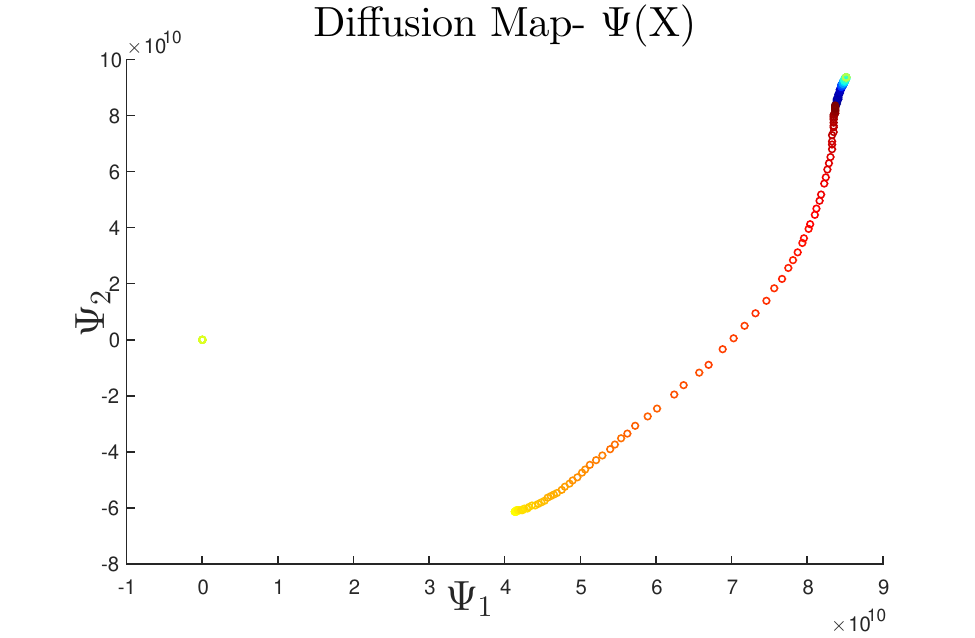}
	\includegraphics[width=6cm]{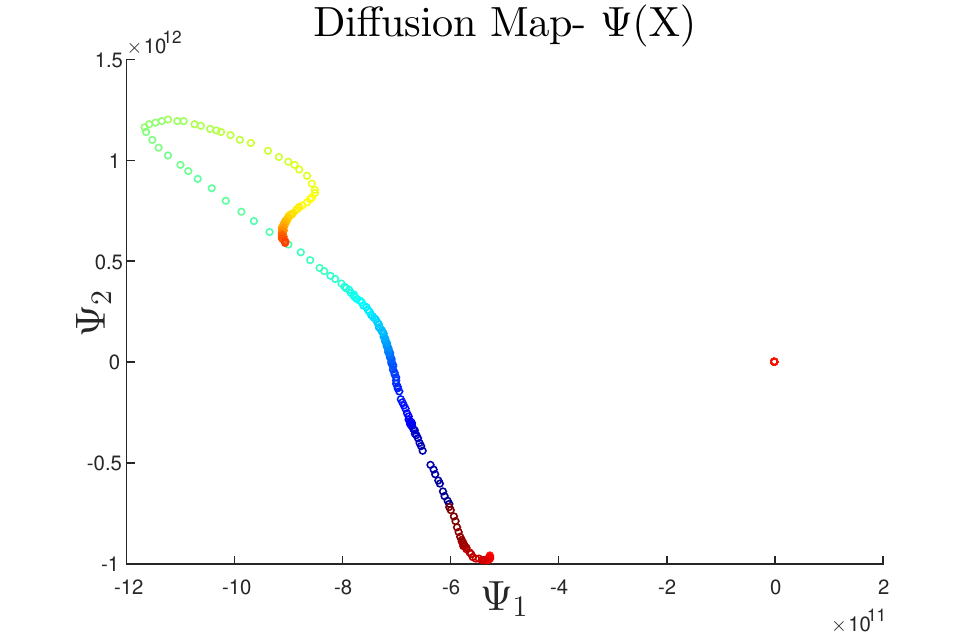}
	\caption{Left: DM-based single-view mapping $\Psi(\mymat{X})$. Right: DM-based single-view mapping $\Psi(\mymat{Y})$). The Gaussian noise deformed the circular structure}
	\label{FigTrainSV2}
\end{figure}

\begin{figure} 
	\centering
	\includegraphics[width=6cm]{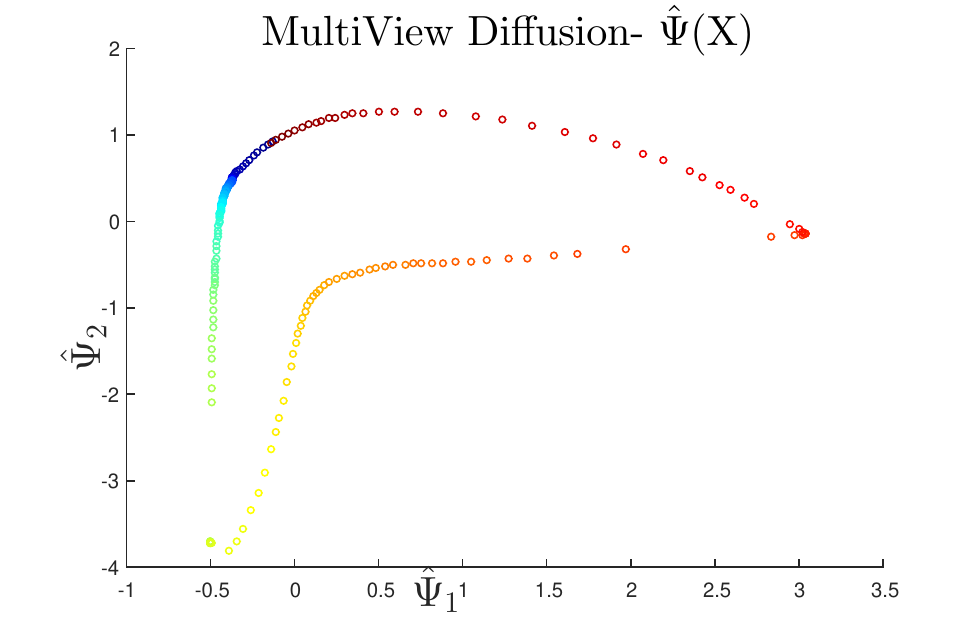}
	\includegraphics[width=6cm]{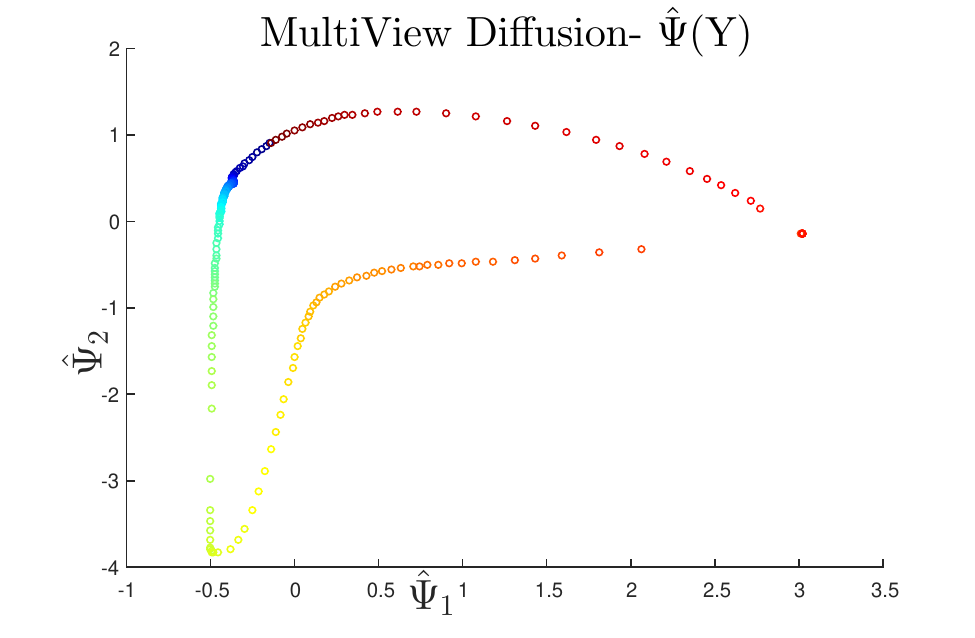}
	\caption{Left: Mapping $\hat{\Psi}(\mymat{X})$. Right: Mapping $\hat{\Psi}(\mymat{Y})$ as extracted by the multi-view framework. Two gaps are visible
		that correspond to Gaussian noise.}
	\label{FigTrainMV2}
\end{figure}

\begin{figure}
	\centering
	\includegraphics[width=6cm]{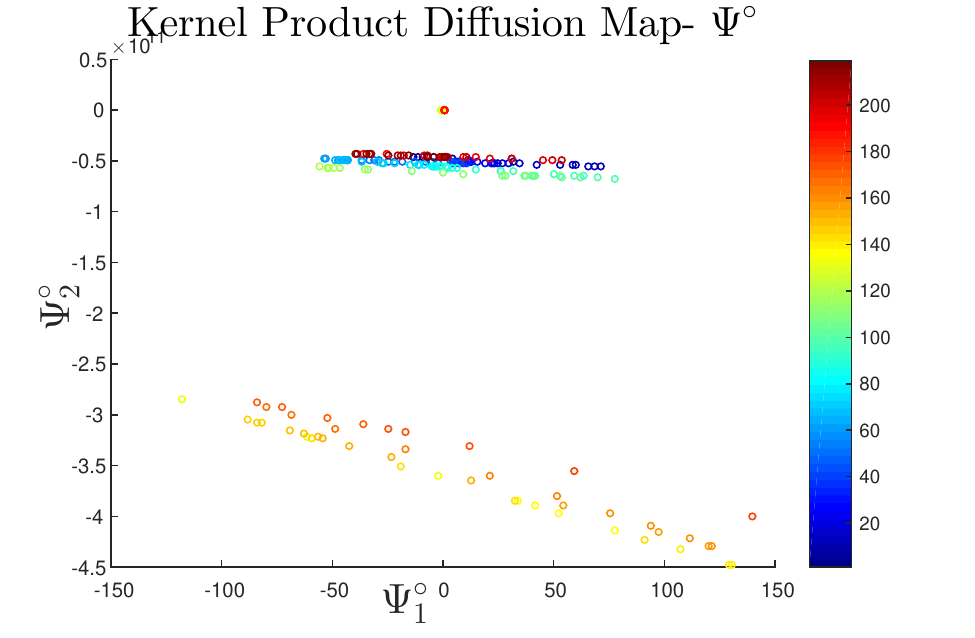}
	\caption{Computation of a standard diffusion mapping (Kernel Product) by using the concatenation vector from both views (corresponding
		to kernel $\mymat{P}^{\circ}$ Eq.\ (\ref{EquationPDH})).}
	\label{FigTrainKP2}
\end{figure}

This experiment was then repeated,  replacing $10$ frames from each
view with ``noise frames" consisting of pixel-wise i.i.d.\ zero-mean Gaussian noise with variance $10$. A single-view DM-based mapping was computed. The Kernel Product-based DM and
the multi-view based DM mappings were computed as well. As presented
in Fig.\ \ref{FigTrainSV2}, the Gaussian noise distorted the
manifolds extracted in each view. The multi-view approach  extracted
two circular structures presented in Fig.\ \ref{FigTrainMV2}. Again,
the data points are ordered according to the position along the
path. This time, the circular structure is unfolded and the gaps are
visible in both embeddings. Applying the Kernel Product
approach (Eq.\ \ref{EquationPDH}) has yielded a distorted manifold
as presented in Fig.\ \ref{FigTrainKP2}.



\section{Applications}
\label{sec:application}

\subsection{Multi-view clustering}
\label{sec:Clust}
The task of clustering has been in the core of machine learning for many years. The goal is to divide a given dataset into subsets based on the inherited structure of the data. We use the multi-view construction to extract low-dimensional mappings from multiple sets of high-dimensional data points. In the following experiments we expand the examples presented in \cite{MVDMC} to cluster artificial and real data sets. For the real data sets applying the multi-view approach requires an eigen decomposition of large matrices. To reduce the runtime of experiments we use an approximate matrix decomposition based on sparse random projections \cite{aizenbud2016matrix}.
\subsubsection{Two circles clustering}
\label{exp1}
Spectral properties of data sets are useful for clustering since they reveal information about the unknown number of clusters.
The characteristic of the eigenvalues of $\mymat{\widehat{P}}$ (Eq.\ (\ref{phat})) can provide insight into the number of clusters within the data set.
The study in \cite{Jordan} relates the number of clusters to the multiplicity of the eigenvalue 1. A different approach in \cite{Perona}
provides an analysis about the relation between the eigenvalue drop to the number of clusters. In this section,
we evaluate how our proposed method captures the clusters' structure when two views are available.

We generate two circles that represent the original clusters using the function

\begin{equation}
\label{Eqz}
\mymat{z}_i=
\begin{bmatrix}
{z_i}[1]\\
{z_i}[2]\\

\end{bmatrix}
=
\begin{bmatrix}
{r \cdot \cos(\theta_i))}\\
{r \cdot \sin(\theta_i)}\\

\end{bmatrix},
\end{equation}
where $M=1,\!600$ points  $\theta_i, 1\leq i\leq M$, are evenly spread in $[0,4\pi]$. The clusters are created by changing the
radius as follows: \\ $r=2,  1\leq i\leq 800$\text{ (first cluster)
}, $r=4,\!801\leq i\leq 1,\!600$ (second cluster). The views
$\mymat{X}$ (Eq.\ (\ref{EC1})) and $\mymat{Y}$ (Eq.\ (\ref{EC2})) are generated by the application of the
following non-linear functions that produce the distorted views
\begin{equation}
\label{EC1}
x_i[1]=
\begin{Bmatrix}
{z_1}[i]+1+n_i[2]|{z_i[2]}\geq0\\
{z_1}[i]+n_i[3]|{z_i[2]} <0 \\

\end{Bmatrix}
,x_{i}[2]= {z_i}[2]+n_i[1]
\end{equation}
and
\begin{equation}
\label{EC2}
y_{i}[1]= {z_i}[1]+n_i[4],
y_{i}[2]=
\begin{Bmatrix}
{z_i}[2]+1+n_i[6]|{z_i}[1]\geq0\\
{z_i}[2]+n_i[6]|{z_i}[1] <0 \\

\end{Bmatrix}
,
\end{equation}
where $n_i[\ell],1\leq \ell\leq6$, are i.i.d.\ random variables drawn from a
Gaussian distribution with $\mu=0$ and $\sigma^2_n\in [0.03,0.6]$. This data is
referred to as the Coupled Circles dataset.
\begin{figure}
	
	\centering
	
	{\includegraphics[width=6cm]{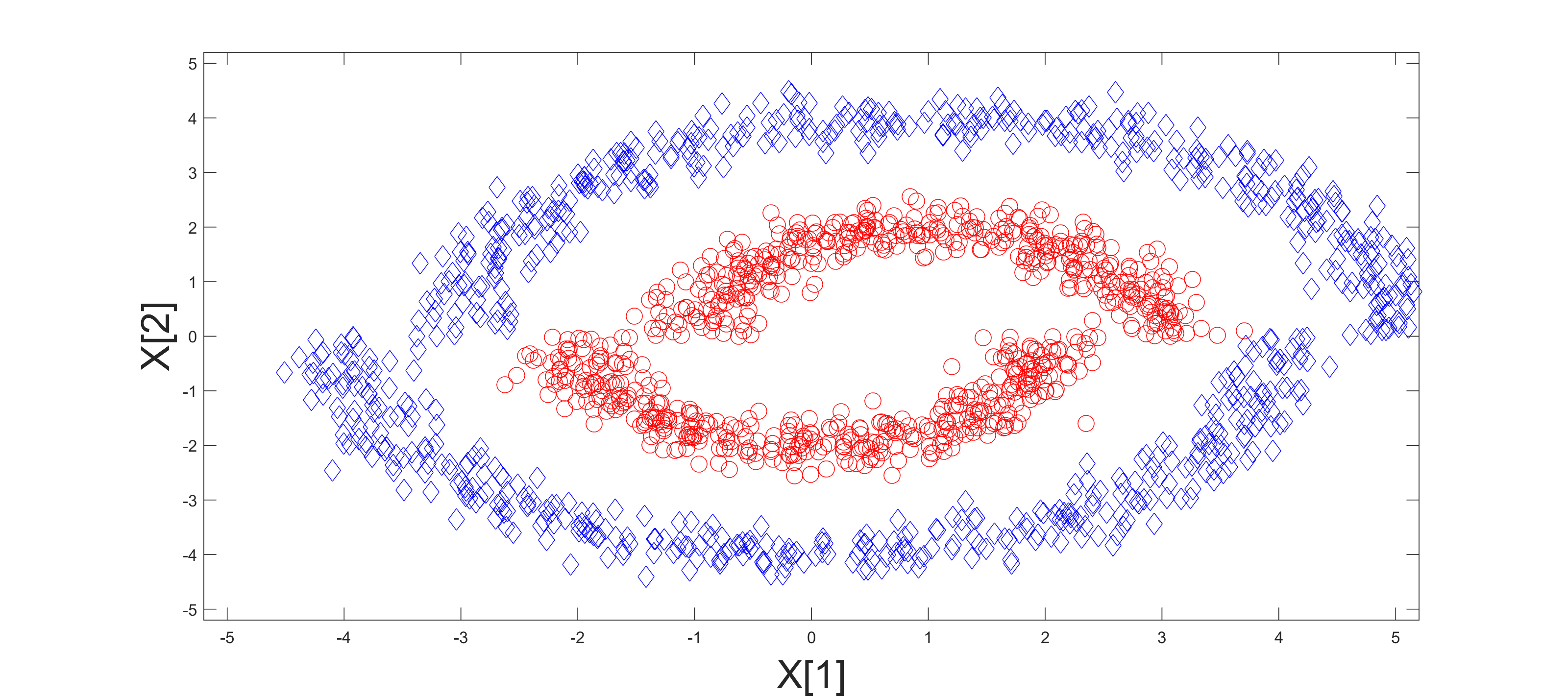}}
	{\includegraphics[width=6cm]{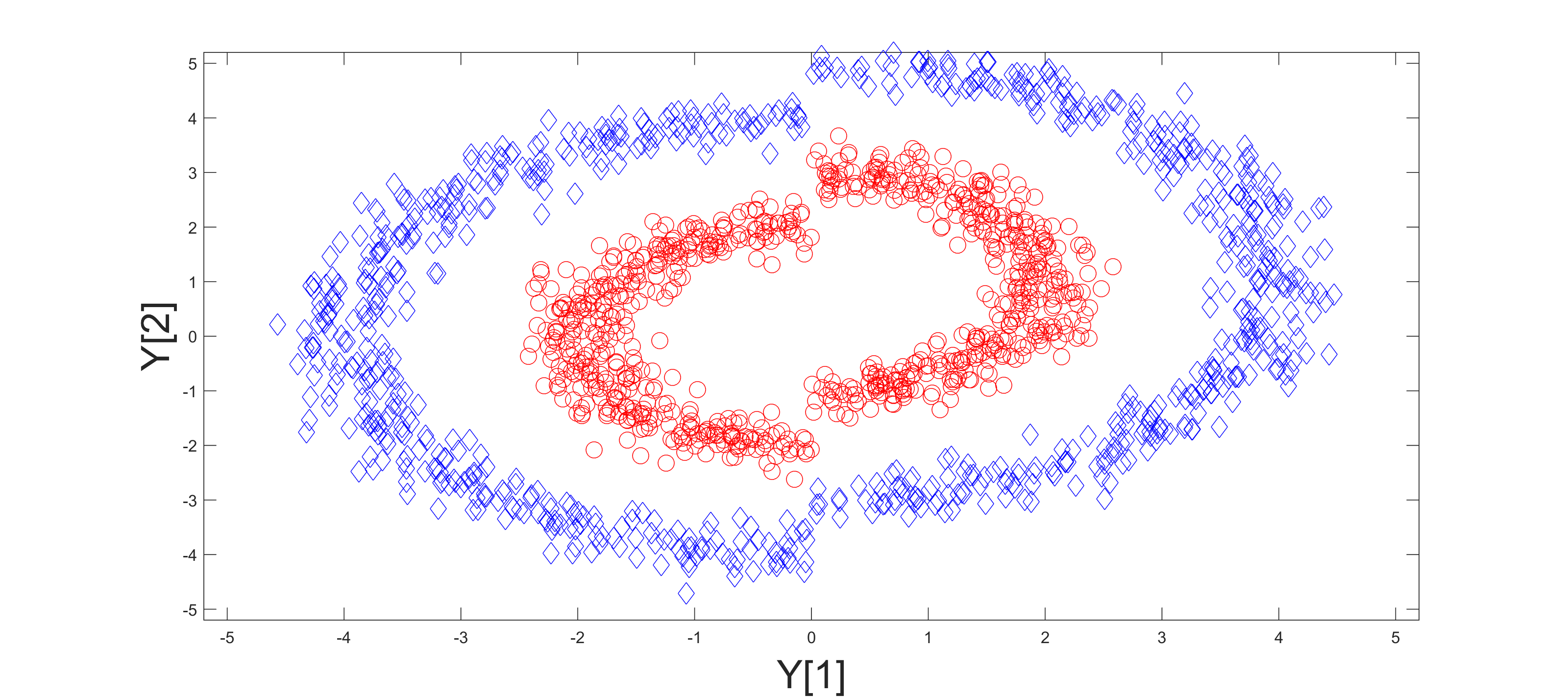}}
	
	\caption{Left: first view $\mymat{X}$. Right: second view
		$\mymat{Y}$. The ground truth clusters are represented by the
		marker's shape and color.} \label{FC1}
\end{figure}

In Fig.\ \ref{FC1}, the views $\mymat{X}$ and $\mymat{Y}$, which were
generated by Eqs.\ (\ref{EC1}) and (\ref{EC2}), are shown. Color
and shape indicate the ground truth clusters. Initially, DM is
applied to each view and clustering is applied using K-means
($K=2$) within the first diffusion coordinate. The kernel bandwidths $\sigma_x\text{ and } \sigma_y$ for all methods are set using the min-max method described in Eq.\ (\ref{eq:MaxMin}). We use $t=1$ since it is optimal for clustering tasks. For the kernel product method we use $\sigma_{\circ}=\sqrt{\sigma_x^2 +\sigma_y^2}$.
We further extract a 1-dimensional representation
using the proposed multi-view framework (Eq.\ (\ref{Map1})), the Kernel Sum DM (Eq.\ (\ref{EquationPDS})), Kernel Product DM (Eq.\ (\ref{EquationPDH})), de Sa's approach (Eq.\ (\ref{eq:DeSa})) and Kernel CCA (Eq.\ (\ref{eq:KCCA})) described in Section \ref{SecMulti}. The regularization parameter is $\gamma=0.01$ for KCCA and we use $100$ components for the Incomplete Cholesky Decomposition \cite{lai2000kernel,bach2002kernel}. Clustering is
performed in the representation space by the application of K-means
where $K=2$.
\begin{figure}
	
	\centering
	
	{\includegraphics[width=9cm]{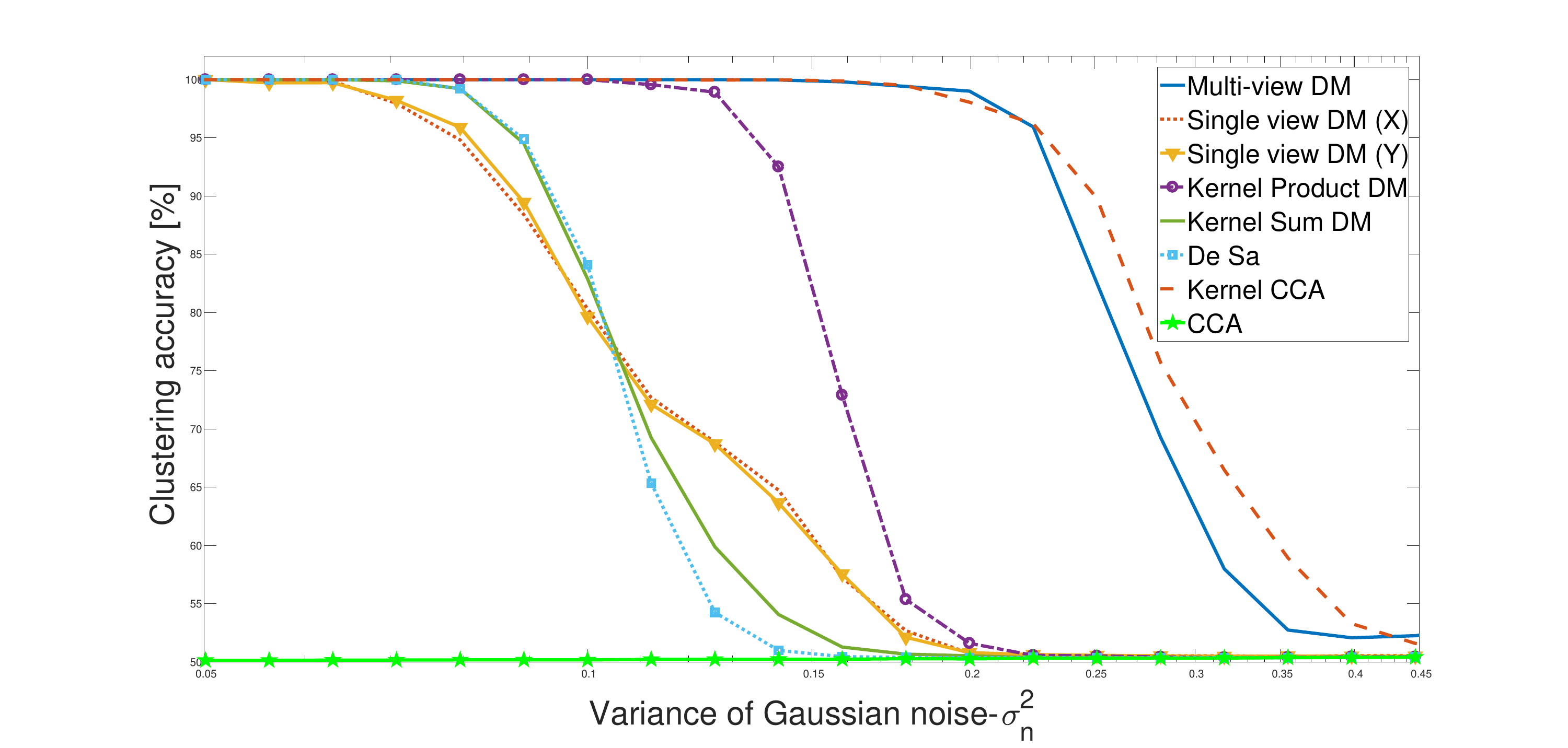}}
	
	\caption{Clustering results from averaging 200 trials vs. the
		variance of the Gaussian noise. The simulation performed on the
		Coupled Circles data (Eqs.\ (\ref{Eqz}), (\ref{EC1}) and
		(\ref{EC2})).} \label{FCcluster}
\end{figure}
To evaluate the performance of our proposed map 100 simulations
with various values of the Gaussian's noise variance (all with zero
mean) were performed. The average clustering success rate is
presented in Fig.\ \ref{FCcluster}. It is
evident that the multi-view based  approach outperforms the DM-based
single-view and the Kernel Product approaches.

\begin{figure}
	
	\centering
	
	{\includegraphics[width=6cm]{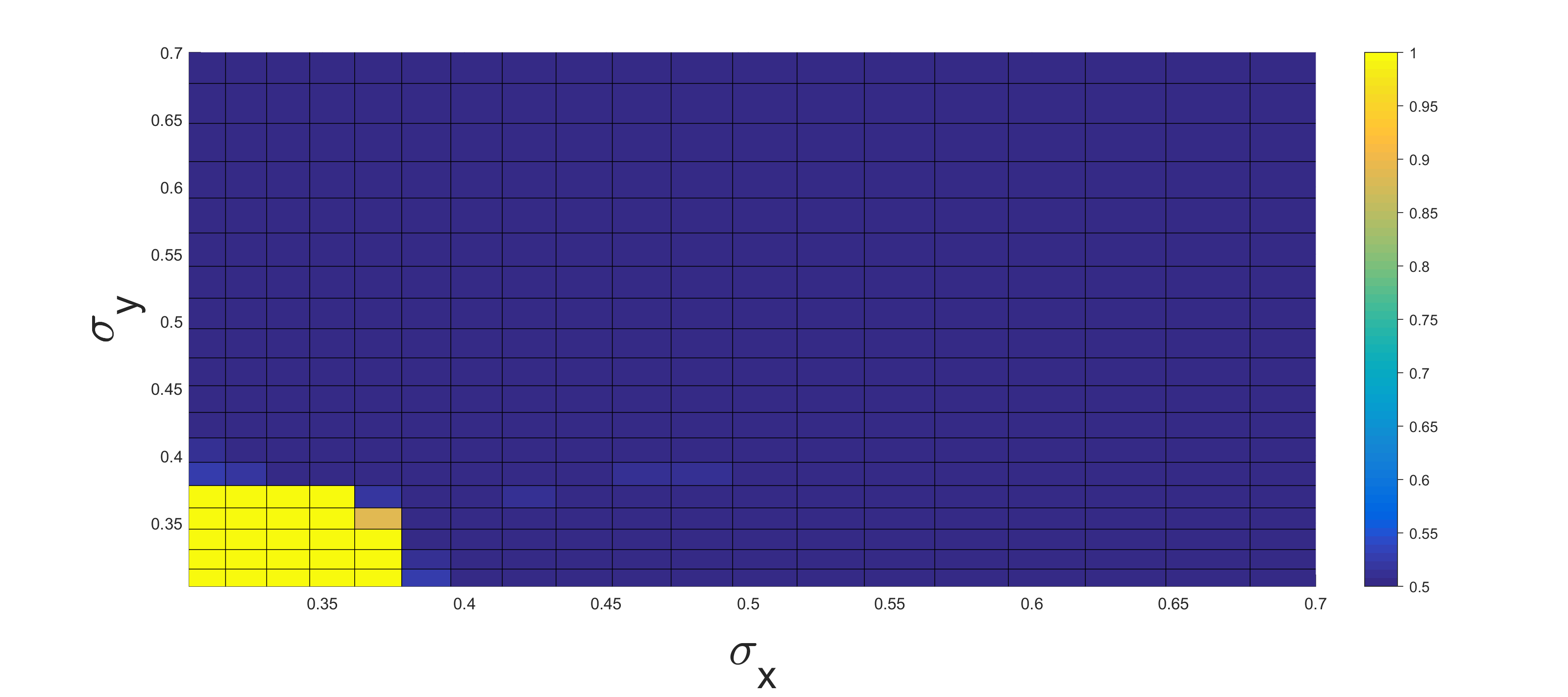}}
	{\includegraphics[width=6cm]{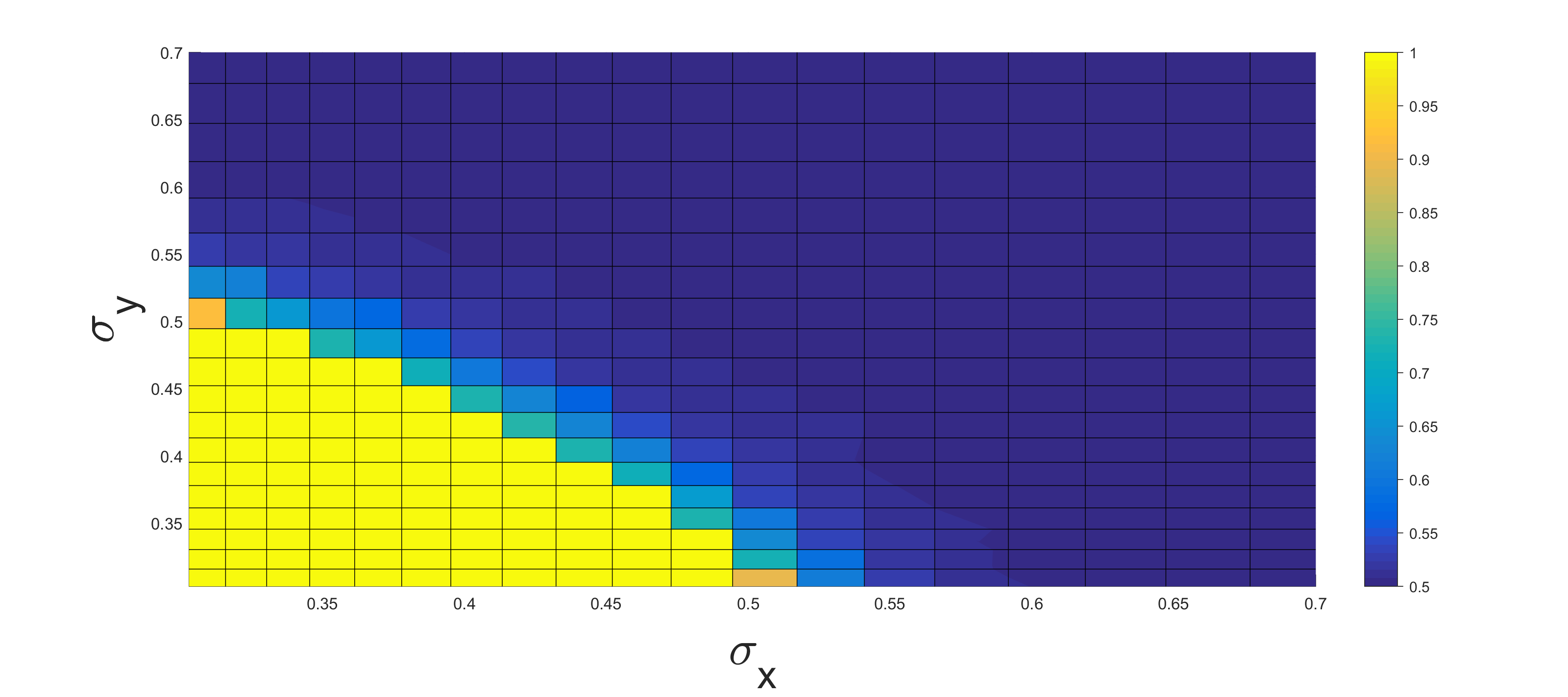}}
	{\includegraphics[width=6cm]{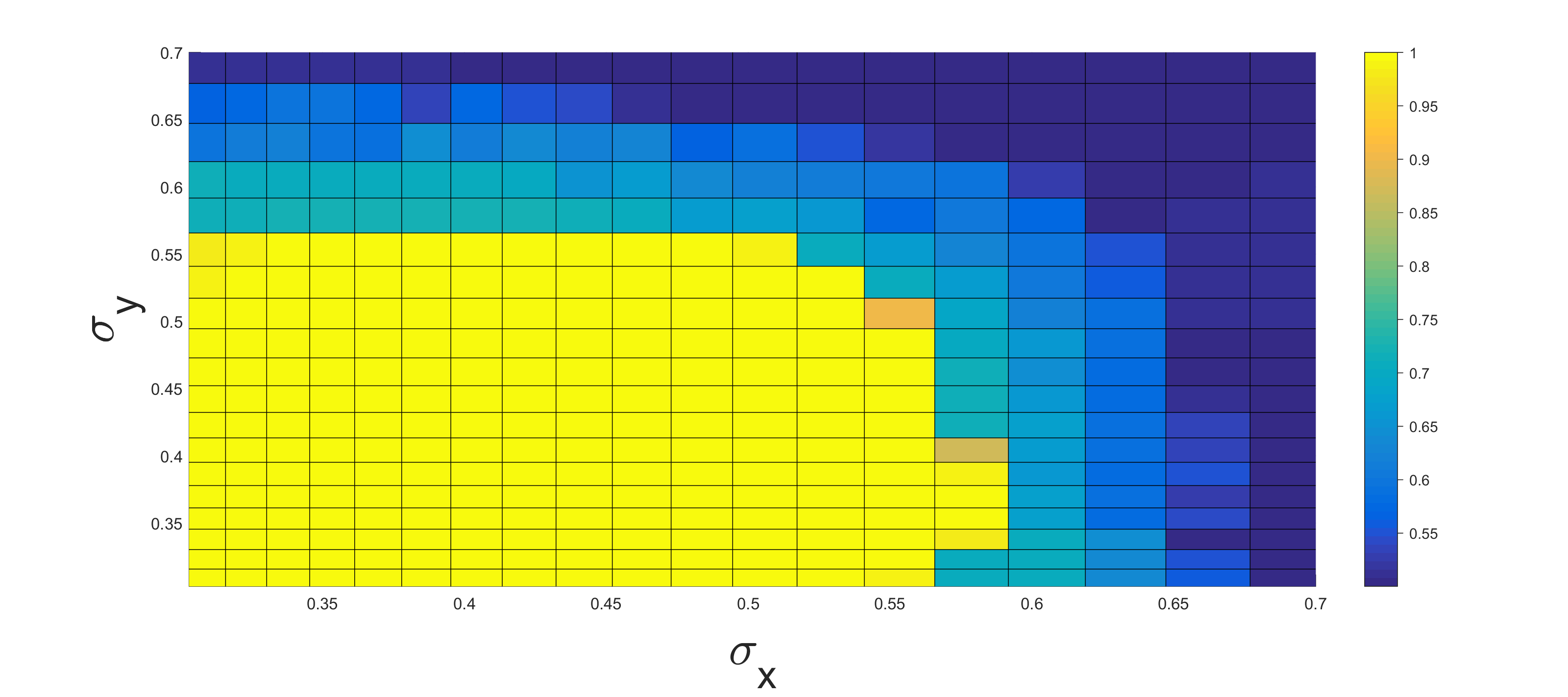}}
	
	\caption{Clustering results from averaging 20 trails using various values of $\sigma_x,\sigma_y$ based on different mappings. The standard deviation of the noise is $\sigma_n=0.16$. Top left- Kernel Sum DM, top right- Kernel Product DM, bottom- Multi View DM.} \label{CirclesParam}
\end{figure}

The performance of kernel methods is highly dependent on setting an appropriate kernel bandwidth $\sigma_x,\sigma_y$, in Algorithm \ref{alg:Singer} we have presented a method for setting such parameters. To evaluate the influence of these parameters on the clustering quality we set $\sigma_n=0.16$ and extract the multi-view, Kernel Sum and Kernel Product diffusion mapping for various values of $\sigma_x,\sigma_y$. The average clustering performance using K-means ($K=2$) are presented in Fig.\ \ref{CirclesParam}.

\subsubsection{{Handwritten digits} }
\begin{figure}

	\centering
	
	{\includegraphics[width=9cm]{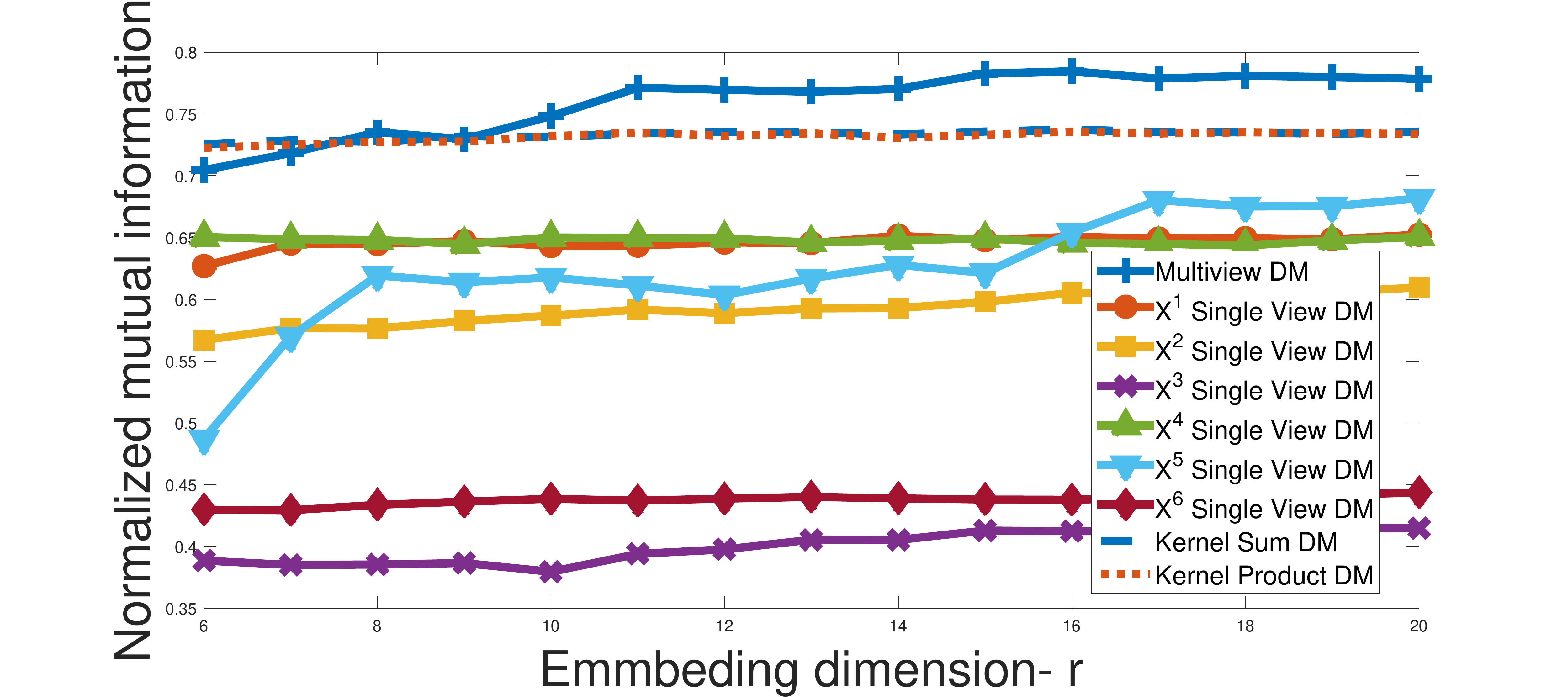}}
	
	\caption{Average clustering accuracy running 100 simulations on the Handwritten data set. Accuracy is measured using the Normalized Mutual Information (NMI).}
	\label{NMI}
\end{figure}

For the following clustering experiment, we use the Multiple Features database~\footnote[2]{http://archive.ics.uci.edu/ml} from the UCI repository.
The data set consists of $2,0\!00$ handwritten digits from $0$ to $9$ that are equally spread. The extracted features from these images are the profile correlations
(FAC), Karhunen-Lo\'{e}ve coefficients (KAR), Zerkine moment (ZER), morphological (MOR), pixel averages in $2 \times 3$ windows and the Fourier coefficients (Fou) as our feature spaces $\mymat{X}^1,\mymat{X}^2$,$\mymat{X}^3,\mymat{X}^4,\mymat{X}^5,\mymat{X}^6$ respectively. We apply dimensionality reduction using a single-view DM, Kernel Product DM, Kernel Sum DM and the proposed Multi-view. We apply K-means to the reduced mapping using 6 to 20 coordinates. The clustering performance is measured using the Normalized Mutual Information \cite{nmi} (NMI). Figure \ref{NMI} presents the average clustering results using K-Means.

\begin{figure}

	\centering
	
	{\includegraphics[width=5cm]{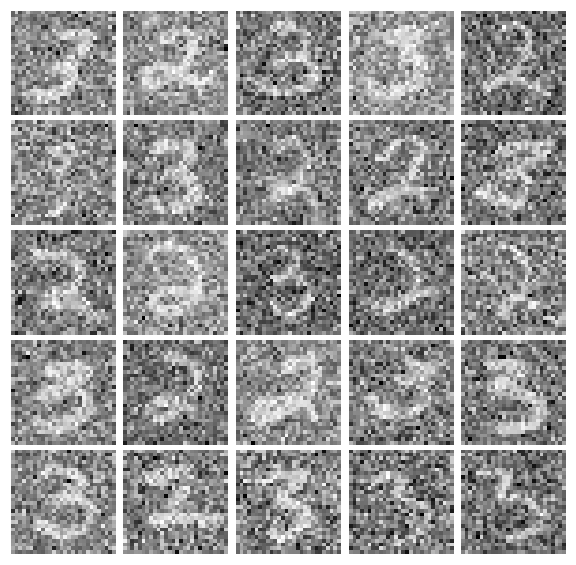}}
	{\includegraphics[width=5cm]{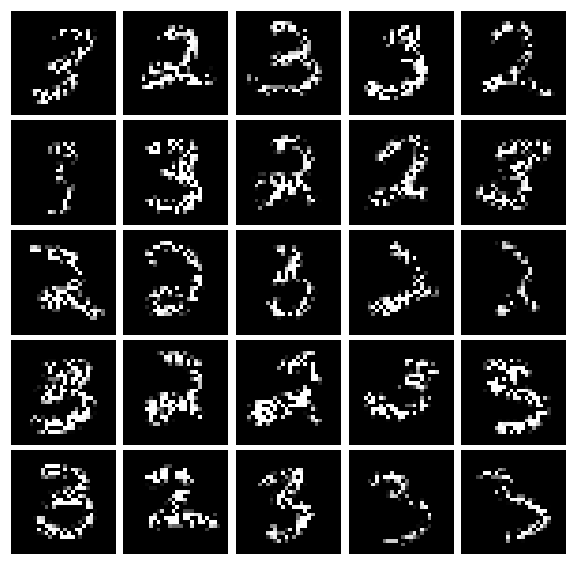}}
	
	\caption{Random samples from both views. Left- $\myvec{X}^1$ generated by adding Gaussian noise. Right- $\myvec{X}^2$ generated by randomly dropping out $50\%$ of the pixels.  }
	\label{MnistL}
\end{figure}

Next, we attempt to cluster $40K$ grey scale images of handwritten $2$'s and $3$'s. The images with dimension $28 \times 28$ were collected from the infinity MNIST dataset \cite{loosli2007training}. We generate two independent noisy versions of the $40K$ samples. By adding pixel Gaussian noise $N(0,0.5)$ to each image we create the first noisy view which is denoted as $\myvec{X}^1$. The second view $\myvec{X}^2$ is created by randomly and independently zeroing each pixel with probability $0.5$. In Fig.\ \ref{MnistL}. we present $25$ examples from each view. To evaluate the clustering performance, we apply K-means $100$ times to the reduced representations with dimensions $5,10,15$ and $20$ and report the top results for each method. In table \ref{MNIST_TAB} we present the top Normalized Mutual Information (NMI) and clustering accuracy's the proposed multi-view and various alternative methods.

\begin{table}[!htb]
 
 	\centering
 	\begin{tabular}{lllll}
 		\toprule
 		Method    &  NMI    & Accuracy  \\
 		\midrule

 		DM $\myvec{X}^1$  & ${0.38}$  & ${84.4\%}$   \\
 		DM $\myvec{X}^2$    & ${0.38}$ & ${84.1\%}$  \\
 		Kernel Prod    & ${0.41}$       & ${85.2\%}$   \\
 		Kernel Sum   & ${0.39}$       & ${84.6\%}$ \\
        Kernel CCA   & ${0.54}$       & ${90.5\%}$ \\
 		de Sa   &  ${0.59}$     & ${91.2\%}$	\\
 		Multiview  & ${\bf{0.70}}$  & ${\bf{94.7\%}} $  \\
 		\bottomrule
 	\end{tabular}
 		\caption{Accuracy of clustering and the normalized mutual information (NMI) on the infinity MNIST dataset. Each view consists a noisy version based on $40K$ images of handwritten $2$'s and $3$'s. The first view $\myvec{X}^1$ is generated by adding Gaussian noise, while $\myvec{X}^2$ is generated by zeroing out random pixels with probability $0.5$.  }
 			\label{MNIST_TAB}
 \end{table}

	\subsubsection{{Isolet data set} }
\begin{figure}

	\centering
	
	{\includegraphics[width=9cm]{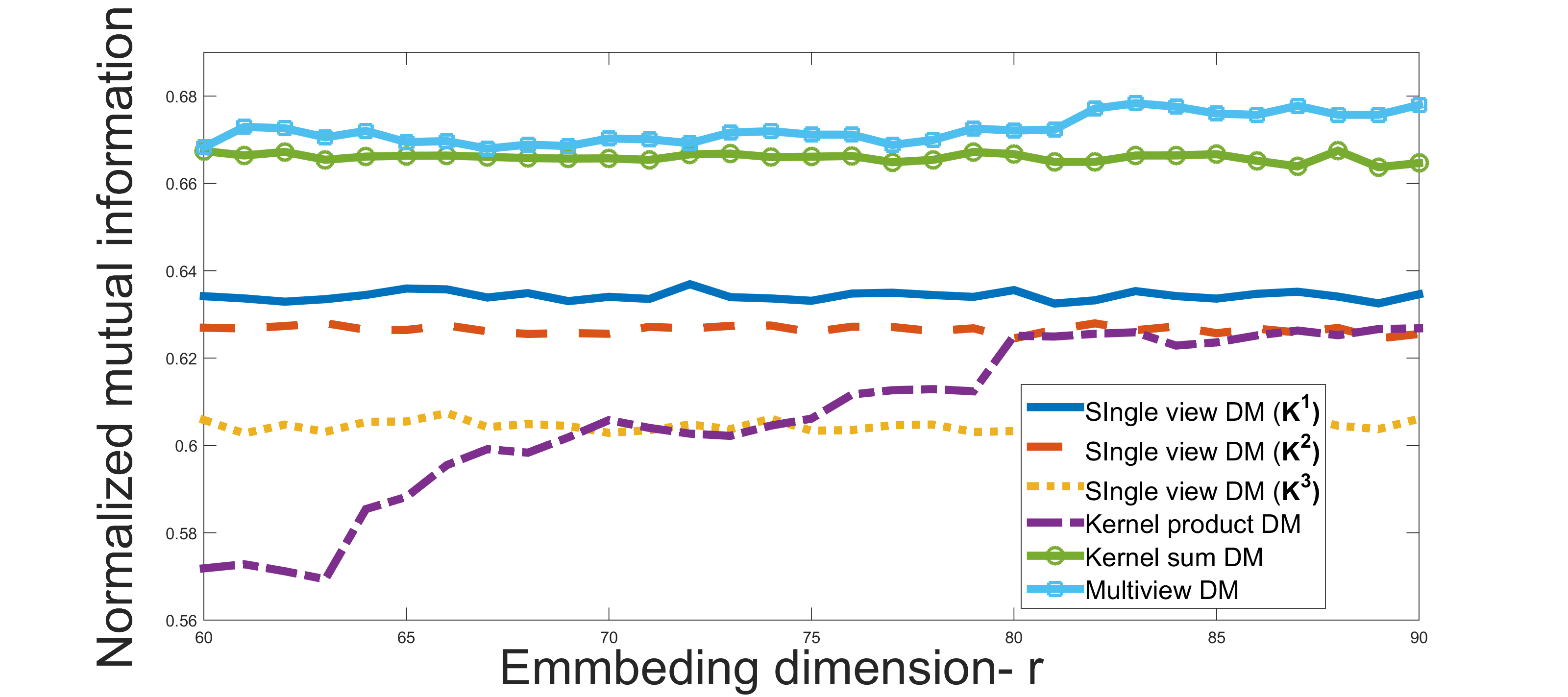}}

	\caption{Clustering accuracy measured with Normalized Mutual Information (NMI) on the Isolet data set by using 3 different kernel matrices. Clustering was performed in the $r$ dimensional embedding space.}
	\label{IsoletNMI}
\end{figure}

The Isolet data set was constructed by recording $150$ people pronouncing each letter twice for all $26$ letters. The feature vector available is a concatenation of the following features: spectral coefficients, contour, sonorant, pre-sonorant and post-sonorant. The authors do not provide the feature's separation, therefore, the dimension of the feature vector is $617$. We use a subset of the data with $1,\!599$ instances, thus the features space is $\mymat{X} \in \mathbb{R}^{1,\!559 \times 617}$. To apply the multi-view approach we compute $3$ different kernels and fuse them together. The first kernel $\myvec{K}^1$ is the standard Gaussian kernel defined in Eq.\ (\ref{EQK}). $\myvec{K}^2$ is a Laplacian kernel defined by \begin{equation}
K^2_{i,j}\defeq \exp \left(\frac {-|\myvec{x}_i-\myvec{x}_j|} {\sigma_2}  \right).
\end{equation} The third kernel $\myvec{K}^3$ is an exponent with a correlation distance as the affinity measure, given by
\begin{equation}
\label{KernelEq}
K^3_{i,j} \defeq \exp \left( \frac{T_{i,j}-1}{2 \sigma^2}   \right) , i,j=1,...,M,
\end{equation} where $T_{i,j}$ is the correlation coefficient between the $i$-th and $j$-th feature vectors, computed by
\begin{equation}
T_{i,j}\defeq \frac{{\myvec{\tilde{x}}_i}^T \cdot \myvec{\tilde{x}}_j }{\sqrt{({\myvec{\tilde{x}}_i}^T\cdot \myvec{\tilde{x}}_i)({\myvec{\tilde{x}}_j}^T \cdot \myvec{\tilde{x}}_j)}} , i,j=1,...,M.
\end{equation} The average subtracted features are $\tilde{\myvec{x}}_i \defeq \myvec{x}_i - \eta_i \cdot \myvec{1}$, where $\eta_i$ is the average of the features for instance $i$. We fuse the kernels using multi-view, kernel product and kernel sum approach, we then apply K-Means to the extracted space. The average NMI for 26 classes is presented in Fig.\ \ref{IsoletNMI}. 

\subsubsection{Caltech 101}
 For this experiment we use an image dataset which consists of $101$ categories collected in \cite{fei} for an object recognition problem. We use two subsets of the dataset, with $10$ and $15$ instances in each category. The views $\mymat{X}^1,\myvec{X}^2$ are represented by the following features: 
 Bag-of-words SIFT descriptors \cite{Bag} and Pyramid Histogram of Oriented Gradients (PHOG) \cite{PHONG}. We use Kernel matrices computed by \cite{gehler}, therefore, we do no set the scale parameters.
 The quality of the clustering is again measured using Calinski-Harabasz Criterion \cite{Calinsk} and the average silhouette width \cite{silhouettes}, for both metrics higher values indicates better separation. The silhouette is defined within the range of $[-1,1]$. The average clustering results based on 10 to 15 coordinates are presented in Table \ref{CAL}.
 \begin{table}[!htb]
 
 	\centering
 	\begin{tabular}{lllll}
 		\toprule
 		Method    &  Silhouette(10)    & Calinski(10)   &  Silhouette(15)      & Calinski(15)   \\
 		\midrule

 		DM $\myvec{X}^1$ (SIFT)   & ${-0.27}$ & ${183}$  & ${-0.27}$ & ${293}$    \\
 		DM $\myvec{X}^2$ (PHOG)   & ${-0.23}$ & ${169}$  & ${-0.19}$ & ${329}$   \\
 		Kernel Prod    & ${-0.08}$       & ${348}$   & ${-0.02}$       & ${683}$\\
 		Kernel Sum   & ${-0.11}$       & ${322}$ & ${-0.05}$       & ${650}$\\
 		de Sa   &  ${-0.35}$     & ${134}$	 &  ${-0.34}$     & ${227}$  \\
 		 		Multiview  & ${\bf{0.32}}$  & ${\bf{727}} $ & ${\bf{0.32}}$  & ${\bf{1275}} $ \\
 		\bottomrule
 	\end{tabular}
 		\caption{Clustering results on two subsets of Caltech-101 data set. Columns represent measures for the quality of clustering. First two rows are scores based on a single-view DM using different features. The last four rows present scores based on the proposed and alternative schemes for a Multiview DM based mapping.}
 			\label{CAL}
 \end{table}

\subsection{Learning from multi sensor seismic data}
Automatic detection and identification of seismic events is an important task, it is carried out constantly for seismic and nuclear monitoring. The monitoring process results in a seismic event bulletin that contains information about the detected events, their locations, magnitudes and type (natural or man made event). Seismic stations usually consist of multiple sensors recording continuously at a low frequency. The amount of available data is huge and only a fraction of the recordings contains the signal of interest. Thus, automatic tools for monitoring are of great interest. A suspect event is usually identified based on the energy of the signal, then, typical discrimination algorithms extract seismic parameters. A simple seismic parameter is the focal depth. Its drawback is that its estimation is usually inaccurate without the depth phases. Other widely used seismic discrimination methods are Ms:mb (surface wave magnitude versus body wave magnitude) and spectral amplitude ratios of different seismic phases \cite{Blandford} \cite{rodgers1997comparison}. Current automatic seismic bulletins comprise a large number of false alarms, which have to be manually corrected by an analyst.

In this subsection we apply the proposed method to extract essential latent seismic parameters. A suspected event is identified based on a short and long time average ratio (STA/LTA) \cite{kuperkoch2012automated}. Then, a time-frequency representation is computed to which multi-view is applied to fuse the data from multiple seismic sensors. Using the multi-view low-dimensional embedding and simple classifiers, we demonstrate capabilities classification of event type (earthquakes vs. explosions) and quarry source for explosions.


\subsubsection{Description of the data set}
The dataset consists of recordings from two different broad band seismic stations MMLI (Malkishua) and HRFI (Harif). Both stations are operated by the Geophysical Institute of Israel (GII) and they are part of the Israel National Seismic Network \cite{Hofstetter}. MMLI and HRFI stations are located to the north and to the south of the analyzed region, respectively. Each station is equipped with a three component STS-2 seismometer, thus the total number of views is $L=6$. All recording are sampled at $F_s=40$Hz. The HRFI data includes $1654$ explosions and $105$ earthquakes, while MMLI includes a subset of $46$ earthquakes, $62$ explosions. The explosions occurred in the south of Israel between the years $2,\!005$-$2,\!015$. 
\subsubsection{ Feature Extraction by Normalized Sonograms }
A seismic event typically generates two underground traveling waves. A primary wave (P) and secondary wave (S). The two waves (P-S) arrive at the recording station with some time delay. A time frequency representation of the recording captures the spectral properties of the event while maintaining the P-S time gap. Here we use a time frequency representation termed sonogram \cite{joswig1995automated} with some modifications. Each single-trace seismic waveform, denoted by $y[n]\in \mathbb{R}^{{N}}$ is a time series signal sampled at the rate of $F_s=40$Hz. The waveform $y[n]$ is decomposed into a set of overlapping windows of length $N_0=256$ using an overlap of $s=0.9$. Thus, the sift between consecutive windows is $N_S=\lfloor 0.1 \cdot 256 \rfloor=25$. A short-time Fourier transform (STFT) is applied to $y[n] $ in each time window. Then, power spectral densities are computed. The resulting spectrogram is denoted by $R(f,t)$, where $f$ is a frequency bin and $t$ is a number of time window (bin). Thus, it contains $T=192$ time bins and $N_0/2=128$ frequency bins.

The sonogram is obtained by summing the spectrogram in equally tempered logarithmically scaled frequency bands, this is done for every time bin. Finally the sonogram is normalized such that the sum of energy in every frequency band is equal to $1$. The result is a normalized sonogram and it is denoted by $S(k,t)$, where $k$ is the frequency band number and $t$ is the time window number. 
The resulting set of sonograms are denoted $\myvec{X}^1,\myvec{X}^2,\myvec{X}^3,\myvec{X}^4,\myvec{X}^5,\myvec{X}^6$. These are the input views for our framework.

\subsubsection{Event classification}
Each sensor records information from the seismic event as well as nuisance noise. We can assume that the noise at each station is independent. Thus, by fusing the measurements from different sensors we may be able to improve detection level.  
To evaluate how well the proposed approach fuses the information, we use a set which includes $46$ earthquakes, $62$ explosions recorded at MMLI (Malkishua) and HRFI (Harif). After we extract the sonogram, the proposed MVDM framework is applied, as well as the single-view DM, kernel product DM and kernel sum DM. Classification is performed by using K-NN (K=1), based on 3 or 4 coordinates from the reduced mapping. The results are presented in table \ref{table1}.
This experiment demonstrates that applying multi-view DM to seismic recordings extracts a meaningful representation.

\begin{figure} 
	\centering
	\includegraphics[width=8cm]{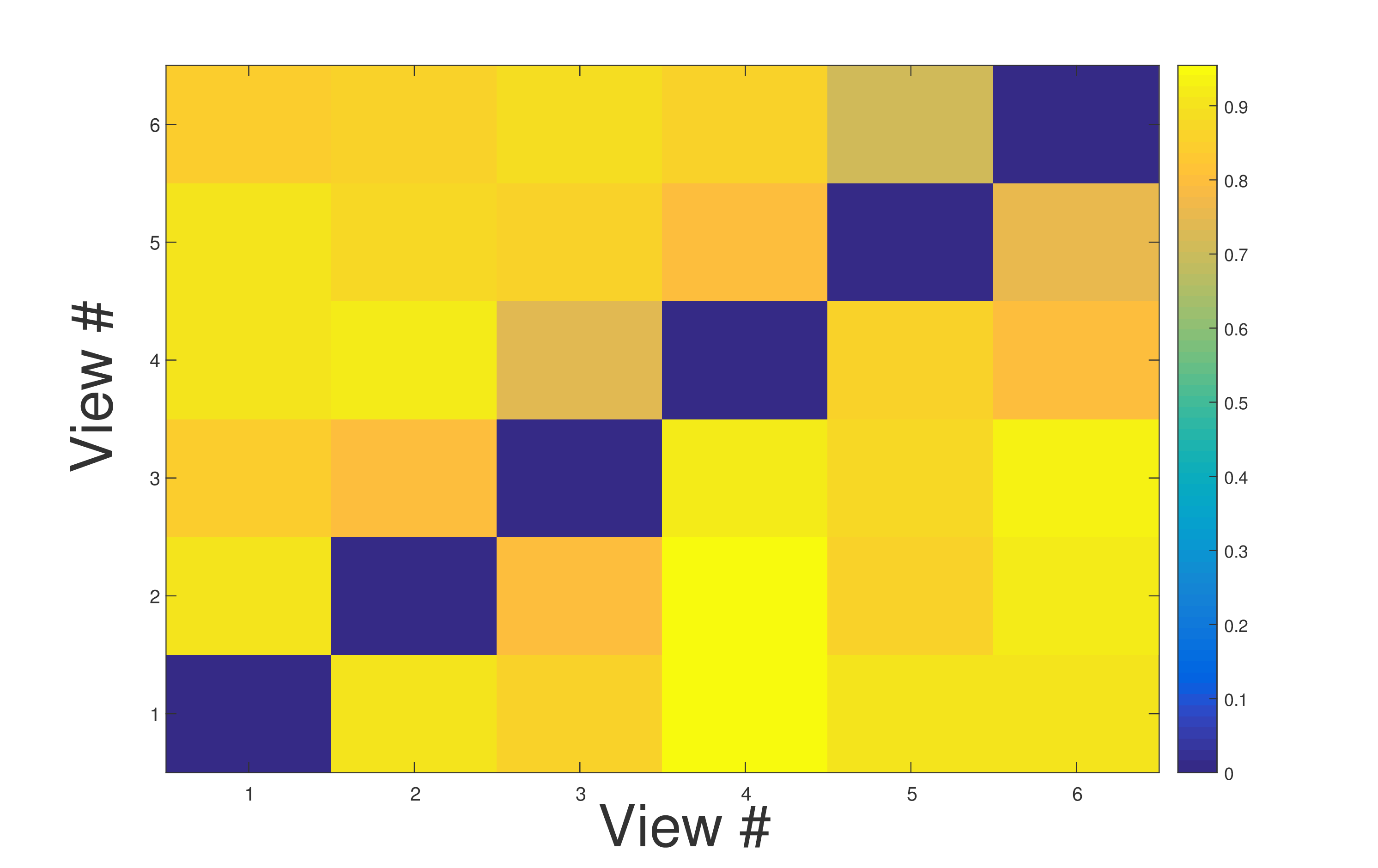}
	
	\caption{Classification accuracy using K-nn (K=1) for all pairs of views $\myvec{X}^l,\myvec{X}^m,l\neq m$. The y-axis is the number of the first view used, while the x-axis is the number of the second view. Classification is performed in the multi-view low-dimensional embedding ($r=4$). The diagonal terms are presented as zero since we did not simulate for $l=m$.}
	\label{FigSubsets}
\end{figure}

In the following test we check how each view affects the detection rate. We do this by applying the MV to subsets of the $L=6$ views. We perform classification using representation computed based on all pairs of view $\myvec{X}^l,\myvec{X}^m,l,m=1,...,6,l\neq m$. The accuracy of classification based on the multi-view representation $\myvec{\widehat{\Psi}}(\myvec{X}^l),l=1,...,6$ given that $\myvec{X}^m,m=1,...,6,m\neq l$ is presented in Fig.\ \ref{FigSubsets}.

\begin{table}[!htb]
	\centering
	\caption{Classification accuracy using 1-fold cross validation. $r$ is the number of coordinates used in the embedding space.}
	
	\label{table1}
	
	\begin{tabular}[t!]{|c|c|c|}
		\hline
		{Method}           & Accuracy  [\%] ($r=3$)  & Accuracy [\%] ($r=4$)    \\
		\hline
		{single-view DM $(\myvec{X}^1)$}     & 89.9	 & 	93.0  \\ 
		{single-view DM $(\myvec{X}^2)$}      & 88.6	& 92.4     \\
		{single-view DM $(\myvec{X}^3)$}  	  	 & 89.3  & 91.1     \\
		{single-view DM $(\myvec{X}^4)$ }     & 89.3   & 89.2  \\
		{single-view DM $(\myvec{X}^5)$}      & 89.3  &90.5 \\
		{single-view DM $(\myvec{X}^6)$}  	   &88.6  & 91.1    \\
		{Kernel Sum DM}          & 93.7  & 94.9    \\
		{Kernel Product DM }          & 94.3	& 93.0  	   \\
		
		{Multi-view DM}          & 97.5	  & 98.1    \\
		\hline

	\end{tabular}
\end{table}

Identification and separation of quarries by attributing the explosions to the known sources is a challenging task \cite{Harris1,Harris2}. Quarry blast have a similar spectral properties, they are usually classified by a triangulation process. Such a process requires to compare the arrival time between distinct seismic stations. Here we attempt to classify the source of the explosions, using $3$-channels from the same station.

\begin{figure}
	\centering
    \includegraphics[scale = 0.1]{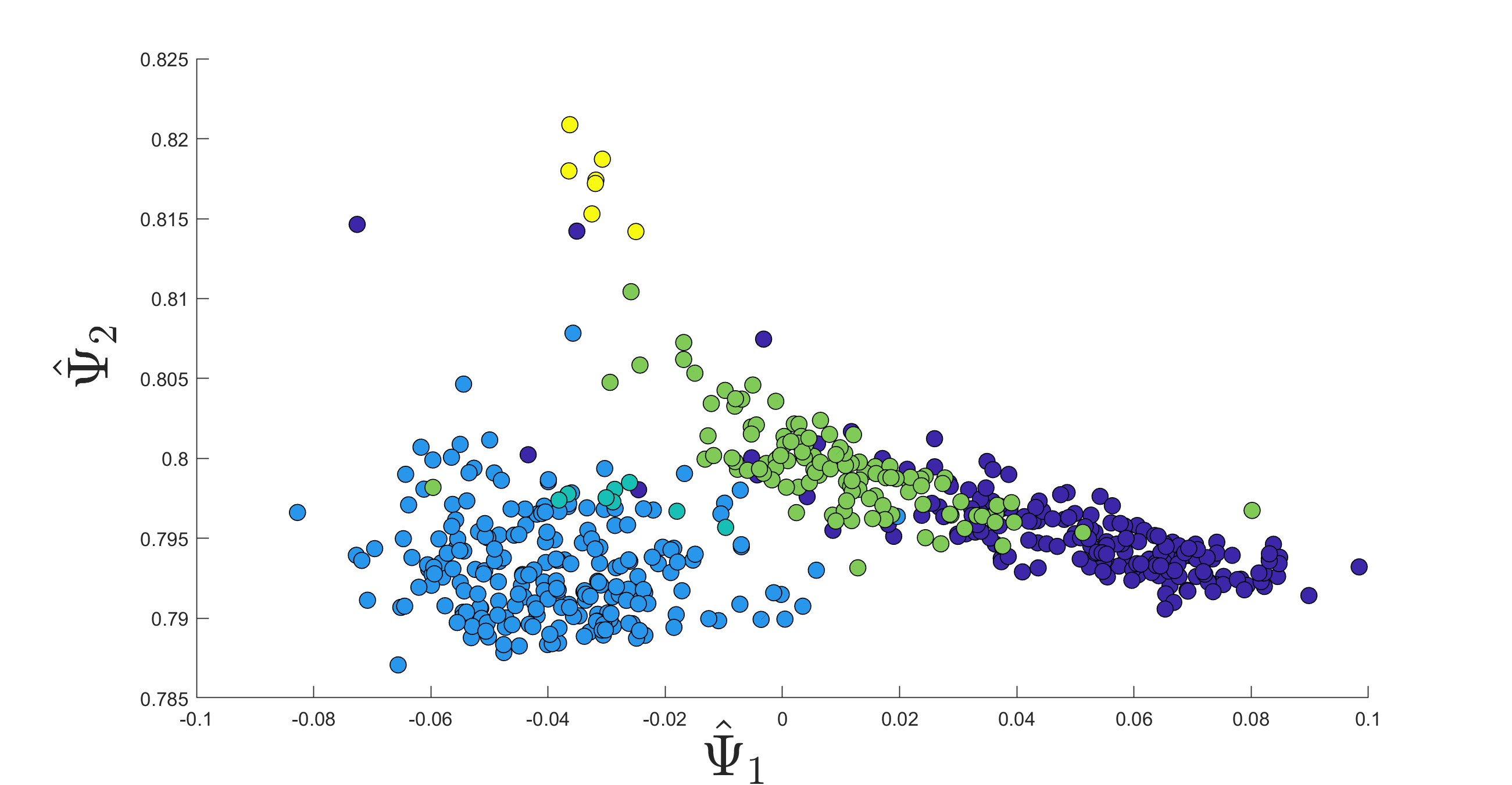}
	\includegraphics[scale = 0.25]{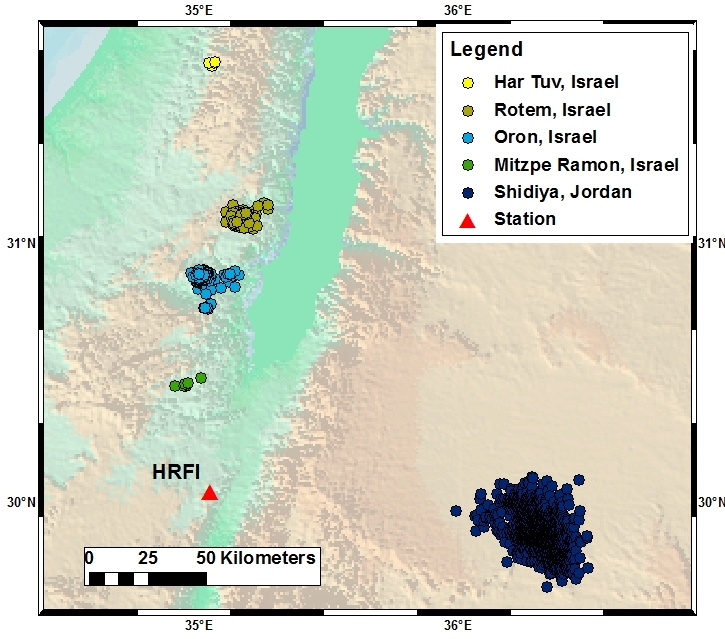}
	\caption{Left- 2 dimensional multi-view DM of the 602 quarry blasts. Points are colored by quarry label. Right- a map with the approximated source location.}\label{fig:QuarryLocations}
\end{figure}

For this experiment $602$ seismograms of explosions are used. The explosions occurred in 4 quarry clusters in Israel and 1 quarry in Jordan. All events were recorded in HRFI station, the distances from the event to the station vary between $50$-$130$Km. The association of each blast to quarry (labeling) is performed manually by an analyst from the GII. After extracting the sonograms, we apply multi-view DM and present the first $2$ coordinates in Fig.\ \ref{fig:QuarryLocations}. In table \ref{table4} we summarize the classification results by applying K-NN $(k=1)$ in a leave one out procedure. Our method is compared to a single-view DM, kernel product and kernel sum.

\begin{table}
\centering
	\caption{Classification accuracy using 1-fold cross validation. $r$ is the number of coordinates used in the embedding space.}
	\label{table4}
	
	\begin{tabular}[t!]{|c|c|c|}
		\hline
		{Method}           & Accuracy  [\%] ($r=3$)  & Accuracy [\%] ($r=4$)    \\
		\hline
		{single-view DM $(\myvec{X}^1)$}     & 80.3	 & 	80.6 \\ 
		{single-view DM $(\myvec{X}^2)$}      & 79.1	& 79.2     \\
		{single-view DM $(\myvec{X}^3)$}  	  	 & 76.4  & 77.8    \\
		{Kernel Sum DM}          & 82.2  & 82.8    \\
		{Kernel Product DM }          & 80.8	& 81.2  	   \\	
		{Multi-view DM}          & 86.2	  & 86.4    \\
		\hline

	\end{tabular}
\end{table}

\section{Discussion}
\label{SecConc} We presented a multi-view based framework for
dimensionality reduction. The framework enables
to extract simultaneous embeddings from coupled embeddings. Our approach is based on imposing an implied cross-domain transition in each single time-step.
The transition probabilities depend on the connectivities in both
views. We reviewed various theoretical aspects of the proposed method
and demonstrated their applicability to both synthetic and real
data. The experimental results demonstrate the strength of the
proposed framework in cases where  data is missing in each view or
each of the manifolds is deformed by an unknown function. The
framework is applicable to various real life machine learning tasks
that consist of multiple views or multiple modalities.

\section{Appendix}
\label{sec:appendix}
\noindent We prove Theorem \ref{T9} (subsection \ref{inf_gen}), repeated here for convenience: 

\setcounter{T1}{6}
\begin{T1}
	The infinitesimal generator induced by the proposed kernel matrix $\widehat{\mymat{K}}$ (Eq.\ (\ref{EQKMAT})) after row-normalization, denoted in here as $\widehat{\mymat{P}}$, converges when
	$M\rightarrow \infty, \epsilon \rightarrow 0$ (with $\epsilon=\sigma_x^2=\sigma_y^2$) to a ``cross domain Laplacian operator''. The functions $f(\myvec{x})$ and $g(\myvec{y})$ converge to eigenfunctions of $\mymat{{\widehat{P}}}$. These functions
	are the solutions of the following diffusion-like equations:
	\begin{align}\label{tophalf} (\widehat{\myvec{P}}f)(\myvec{x}_i)&=g(\myvec{\beta(x_i)})+ {{\epsilon}} \triangle \gamma(\myvec{\beta(\myvec{x}_i)})/\alpha(\myvec{\beta(x_i)}) +\mathcal{O}(\epsilon^{3/2}), \\
	(\widehat{\myvec{P}}g)(\myvec{y}_i)&= f(\myvec{\beta^{-1}}(\myvec{y}_i))+{{\epsilon}} \triangle \eta(\myvec{\beta^{-1}(\myvec{y}_i)})/\alpha(\myvec{\beta(y_i)}+\mathcal{O}(\epsilon^{3/2}),  \end{align} where the functions $\gamma, \eta$ are defined as $\gamma(\myvec{z})\defeq g(\myvec{z})\alpha(\myvec{z}),\eta(\myvec{z})\defeq f(\myvec{z})\alpha(\myvec{z})$.
\end{T1}

\begin{proof}
	
	By extending the single-view construction presented in \cite{Lafon}, the eigenfunction of the limit-operator $\widehat{\myvec{P}}$ is defined using the functions $\myvec{f(x)}$	and $\myvec{g(y)}$ by concatenating the vectors such that
	$$
	\myvec{h}=[{f}(\myvec{x}_1),{f}(\myvec{x}_2),...,{f}(\myvec{x}_M),{g}(\myvec{y}_1),{g}(\myvec{y}_2),...,{g}(\myvec{y}_M)]\in
	\mathbb{R}^{2M}.$$
    The limit of the top-half (Eq.\ (\ref{tophalf})) of the characteristic equation is given by
	\begin{equation} \label{eq:InfGen}\underset{\epsilon\rightarrow 0}{\underset{M\rightarrow \infty}{\text{lim}}} (\widehat{P}h_i)= \underset{\epsilon\rightarrow 0}{\underset{M\rightarrow \infty}{\text{lim}}} {h}_i- \frac{\sum\limits_{j=1}^{2M}{{\widehat{K}_{i,j}}}h_j}{\sum\limits_{j=1}^{2M}{\widehat{K}_{i,j}}}=\underset{\epsilon\rightarrow 0}{\underset{M\rightarrow \infty}{\text{lim}}} f(\myvec{x}_i)- \frac{\sum\limits_{j=1}^{M} \sum\limits_{\ell=1}^{M}{{{K}^x_{i,\ell}}}{{{K}^y_{\ell,j}}}g(\myvec{y}_j)}{\sum\limits_{j=1}^{M} \sum\limits_{\ell=1}^{M}{{{K}^x_{i,\ell}}}{{{K}^y_{\ell,j}}}}, i=1,...,M.\end{equation}
	
	We approximate the summations using a Riemann integral. Beginning with the denominator, we have
	\[\frac{1}{M^2 \epsilon^d}{\sum\limits_{j=1}^{M} \sum\limits_{\ell=1}^{M}{{{K}^x_{i,\ell}}}{{{K}^y_{\ell,j}}}}\underset{\epsilon\rightarrow 0}{\underset{M \rightarrow \infty}{\longrightarrow} }D(\ux)\defeq\frac{1}{\epsilon^d} \int_{\mathbb{R}^d}\int_{\mathbb{R}^d}K\Big(\frac{\myvec{s}-\myvec{x}}{\sqrt{\epsilon}}\Big)K\Big(\frac{\myvec{y}-\myvec{\beta}(s)}{\sqrt{\epsilon}}\Big){\alpha(\myvec{y})}\ds\dy.\] 
	Using a change of variables $\myvec{z}=\frac{\myvec{y-\beta(s)}}{\sqrt{\epsilon}},\myvec{y}=\myvec{\beta}(\myvec{s})+\sqrt{\epsilon}{\myvec{z}}, \dz=\dy\epsilon^{d/2}$ we get
	\[
	D(\ux)=
	\frac{1}{\epsilon^{d/2}} \int_{\mathbb{R}^d}\int_{\mathbb{R}^d}K\Big(\frac{\myvec{s}-\myvec{x}}{\sqrt{\epsilon}}\Big)K(\myvec{z}){\alpha}(\myvec{\beta}(\myvec{s})+\sqrt{\epsilon}{\myvec{z}})\ds\dz.  \]
	Using a first order Taylor expansion of ${\alpha}(\cdot)$ we get 
	\[D(\ux)\approx\frac{1}{\epsilon^{d/2}} \int_{\mathbb{R}^d}\int_{\mathbb{R}^d}K\Big(\frac{\myvec{s}-\myvec{x}}{\sqrt{\epsilon}}\Big)K(\myvec{z})\left[{\alpha}(\myvec{\beta}(\myvec{s}))+\frac{\sqrt{\epsilon}}{2}\myvec{z}^\tps \nabla {\alpha}(\myvec{\beta}(\myvec{s}))+\mathcal{O}(\epsilon)\right]\ds\dz,  \] using the symmetry of the kernel ${K\myvec{(z)}}$ we have
	\[ \int_{\mathbb{R}^d}K(\myvec{z}) \myvec{z}^\tps\dz=\myvec{0}^\tps,\] 	
	therefore, applying another change of variables $\myvec{t}=\frac{\myvec{s}-\myvec{x}}{\sqrt{\epsilon}},\myvec{s}=\sqrt{\epsilon}\myvec{t}+\myvec{x}, \dt=\ds\epsilon^{d/2}$ we get	
	\[D(\ux)\approx \int_{\mathbb{R}^d}K(\myvec{t})[{\alpha}(\myvec{\beta}(\myvec{x}+\sqrt{\epsilon}\myvec{t}))+\mathcal{O}(\epsilon)]\dt\approx {\alpha}(\myvec{\beta}(\myvec{x})) +\mathcal{O}(\epsilon)\] 
	(independent of $\ux$), where the last transition is again based on a Taylor expansion (of $\beta(\cdot)$) and on zeroing out the odd ($1$st) moment of $\myvec{K(t)}$. 
	
	Turning to the numerator (of Eq.\ (\ref{eq:InfGen})),
	\[\frac{1}{M^2 \epsilon^d}{\sum\limits_{j=1}^{M} \sum\limits_{\ell=1}^{M}{{{K}^x_{i,\ell}}}{{{K}^y_{\ell,j}}}g(\myvec{y}_j)}\underset{\epsilon\rightarrow 0}{\underset{M \rightarrow \infty}{\longrightarrow}}N(\ux)\defeq \frac{1}{\epsilon^d} \int_{\mathbb{R}^d}\int_{\mathbb{R}^d}K\Big(\frac{\myvec{s}-\myvec{x}}{\sqrt{\epsilon}}\Big)K\Big(\frac{\myvec{y}-\myvec{\beta}(s)}{\sqrt{\epsilon}}\Big)g(\myvec{y})\alpha(\myvec{y})\ds\dy.\]  
	By applying a change of variables $\myvec{z}=\frac{\myvec{y-\beta(s)}}{\sqrt{\epsilon}}, \myvec{y}=\myvec{\beta}(\myvec{s})+\sqrt{\epsilon}{\myvec{z}}, \dz=\dy\epsilon^{d/2}$ we get
	\[ N(\ux)=\frac{1}{\epsilon^{d/2}} \int_{\mathbb{R}^d}\int_{\mathbb{R}^d}K\Big(\frac{\myvec{s}-\myvec{x}}{\sqrt{\epsilon}}\Big)K(\myvec{z})\gamma(\myvec{\beta}(\myvec{s})+\sqrt{\epsilon}\myvec{z})\ds\dz.\] 
	Using Taylor's expansion of $\gamma\myvec{(\cdot)}$ we get	
	\[ N(\ux)\approx \frac{1}{\epsilon^{d/2}} \int_{\mathbb{R}^d}\int_{\mathbb{R}^d}K\Big(\frac{\myvec{s}-\myvec{x}}{\sqrt{\epsilon}}\Big)K(\myvec{z})\left[ \gamma(\myvec{\beta}(\myvec{s}))+\frac{\sqrt{\epsilon}}{2}\myvec{z}^T \nabla \gamma(\myvec{\beta}(\myvec{s}))+\frac{\epsilon}{2}\myvec{z}^T \mymat{H} \myvec{z}+\mathcal{O}(\epsilon^{3/2}) \right]\ds\dz  \] 
	where ${H}_{i,j}\defeq \frac{\partial^2 \gamma(\myvec{\beta(s)})}{\partial {s_i}\partial s_j}$ is the Hessian. The first term yields the integral over ${K\myvec{(z)}}$, while the second term vanishes due to integration over an odd ($1$st) moment of the symmetric kernel ${K\myvec{(z)}}$.
	The last term yields
	\begin{multline*}
	\int_{\mathbb{R}^d}K(\myvec{z}) \myvec{z}^\tps  \frac{\partial^2 \gamma(\myvec{\beta(s)})}{\partial {s_i}\partial s_j}\myvec{z}\dz
	= \sum_{i,j} \frac{\partial^2 \gamma(\myvec{\beta(s)})}{\partial {s_i}\partial s_j}\int_{\mathbb{R}^d}z_i z_j K(\myvec{z})\dz\\
	= \sum_i \frac{\partial^2 \gamma(\myvec{\beta(s)})}{\partial {s_i}^2}\int_{\mathbb{R}^{d}}z_i^2 K(\myvec{z})\dz = \triangle \gamma(\myvec{\beta(s)}),
	\end{multline*}
	where $\triangle$ denotes the Laplacian operator, and where we assumed that $\int_{\Rset^d}z_iz_jK(\uz)\dz$ vanishes for $i\ne j$ and equals $1$ for $i=j$. Note that this is naturally satisfied, e.g., by the Gaussian kernel function $K(\uz)=c\cdot\exp(-0.5\|\uz\|^2)$ (with $c=(2\pi)^{-d/2}$, as per the scaling requirement). Substituting into $N(\ux)$ we get
	\[N(\ux)\approx\frac{1}{\epsilon^{d/2}} \int_{\mathbb{R}^d}K\Big(\frac{\myvec{s}-\myvec{x}}{\sqrt{\epsilon}}\Big)\left[ \gamma(\myvec{\beta}(\myvec{s}))+\frac{{\epsilon}}{2} \triangle \gamma(\myvec{\beta(s)})+\mathcal{O}(\epsilon^{3/2}) \right]\ds.\]
	Using a change of variables $\myvec{t}=\frac{\myvec{s}-\myvec{x}}{\sqrt{\epsilon}},\myvec{s}=\sqrt{\epsilon}\myvec{t}+\myvec{x}, \dt=\ds\epsilon^{d/2}$ and $\gamma(\myvec{y})=g(\myvec{y})\alpha(\myvec{y})$ we get
	\[N(\ux)\approx \int_{\mathbb{R}^d}K(\myvec{t})\left[ \gamma(\myvec{\beta}(\myvec{x+\sqrt{\epsilon}\myvec{t}}))+\frac{{\epsilon}}{2} \triangle \gamma(\myvec{\beta(\myvec{x+\sqrt{\epsilon}\myvec{t}})})+\mathcal{O}(\epsilon^{3/2}) \right]\dt.\] 
	Using Taylor's expansion (of $\beta(\cdot)$) once again we get
	\[N(\ux) \approx \int_{\mathbb{R}^d}K(\myvec{t})\left[ \gamma(\myvec{\beta}(\myvec{x}))+\frac{{\epsilon}}{2} \triangle \gamma(\myvec{\beta(\myvec{x})})+\frac{{\epsilon}}{2} \myvec{t}^T\mymat{H}\myvec{t}+\mathcal{O}(\epsilon^{3/2}) \right]\dt,\] 
	here we neglected terms involving $\epsilon$ to a power higher than $3/2$ and terms with odd order of $\myvec{t}$ due to the symmetry of the kernel $K$. 
	This 
	leads to
	\[N(\ux)\approx \gamma(\myvec{\beta}(\myvec{x}))+{{\epsilon}} \triangle \gamma(\myvec{\beta(\myvec{x})})+\mathcal{O}(\epsilon^{3/2}) ,\] dividing by the denominator we get 
	\[(\widehat{\myvec{P}}f)(\myvec{x}_i)\approx g(\myvec{\beta(x_i)})+ {{\epsilon}} \triangle \gamma(\myvec{\beta(\myvec{x}_i)})/\alpha(\myvec{\beta(x_i)}) +\mathcal{O}(\epsilon^{3/2})\]

	In the same way, we
	compute the convergence on $g(\myvec{y}_i)$
	\[(\widehat{\myvec{P}}g)(\myvec{y}_i)= f(\myvec{\beta^{-1}}(\myvec{y}_i))+{{\epsilon}} \triangle \eta(\myvec{\beta^{-1}(\myvec{y}_i)})/\alpha(\myvec{\beta(y_i)}+\mathcal{O}(\epsilon^{3/2}) .\]

\end{proof}
The following issues were ignored in the proof:
\begin{itemize}
	\item{Errors due to approximating the sum by an integral; An upper bound on the associated errors in the single-view DM is derived in \cite{Singer}.}
	\item{Deformation due to the fact that the data is sampled from a non uniform density. This changes the result by some constant.}
	\item{The data lies on some manifold. This could be dealt by changing the coordinate system and integrating on the manifold.}
	\item{When assuming that the data lies on some manifold, the Euclidean distance should be replaced by the geodesic distance along the manifold. As in the analysis of \cite{Lafon}, this introduces a factor to the integral.}
\end{itemize}

\section{References}
\bibliographystyle{unsrt}   
\bibliography{listbMul}  

\end{document}